\documentclass[lettersize,journal]{IEEEtran}

\IEEEoverridecommandlockouts 

\usepackage{cite}
\usepackage{tabularx}
\usepackage{amsmath,amssymb,amsfonts}
\usepackage{graphicx}
\usepackage{textcomp}

\DeclareMathOperator{\st}{s.t.}
\usepackage[table,xcdraw]{xcolor}
\usepackage[linesnumbered,ruled,lined]{algorithm2e}

\usepackage{authblk}

\usepackage{enumitem}
\usepackage{hyperref}
\usepackage{breqn}
\usepackage{algpseudocode}
\usepackage{caption}
\usepackage{subcaption}
\usepackage{blindtext}
\usepackage{tcolorbox}
\usepackage{array}
\usepackage{makecell}
\usepackage{balance}

\usepackage[english]{babel}
\usepackage{amsthm}

\allowdisplaybreaks

\newcommand{\scaleboxplotstatic}{0.126}

\newcommand{\scaleboxplotdynamic}{0.13}

\SetCommentSty{mycommfont} 

\title{ MMD-OPT : Maximum Mean Discrepancy Based Sample Efficient Collision Risk Minimization for Autonomous Driving  }
\author{Basant Sharma, Arun Kumar Singh\thanks{Basant and Arun are with the University of Tartu. This research was in part supported by grant PSG753 from Estonian Research Council, collaboration project LLTAT21278 with Bolt Technologies and project TEM-TA101 funded by European Union and Estonian Research Council.
Emails: basant.sharma@ut.ee, arun.singh@ut.ee, Code: \url{https://github.com/Basant1861/MMD-OPT}} 
}

\makeatother

\begin{document} 

\maketitle
\thispagestyle{empty}
\pagestyle{empty}



\begin{abstract}
We propose MMD-OPT: a sample-efficient approach for minimizing the risk of collision under arbitrary prediction distribution of the dynamic obstacles. MMD-OPT is based on embedding distribution in Reproducing Kernel Hilbert Space (RKHS) and the associated Maximum Mean Discrepancy (MMD). We show how these two concepts can be used to define a sample efficient surrogate for collision risk estimate. We perform extensive simulations to validate the effectiveness of MMD-OPT on both synthetic and real-world datasets. Importantly, we show that trajectory optimization with our MMD-based collision risk surrogate leads to safer trajectories at low sample regimes than popular alternatives based on Conditional Value at Risk (CVaR).
\end{abstract}

\def\abstractname{Note to Practitioners}

\begin{abstract}
    Autonomous Driving software stacks have dedicated modules for predicting trajectories of the obstacles(neighboring vehicles). Typically, these predictors provide a set of possible future motions for the obstacles, each of which can have different likelihoods of happening. Thus, a key challenge is to reason about collision risk in a given scene based on predicted trajectories, but without being overly conservative. For example, treating each predicted trajectory as a separate obstacle, without any attention to their likelihood, may not allow any feasible motion to the ego-vehicle. Our work addresses this challenge by proposing a probabilistic approach for modeling and minimizing collision risk. Our core impact lies in improving sample efficiency: that is assessing and minimizing collision risk based on just a handful of predicted trajectories for a given obstacle. Our approach can be easily integrated with any deep neural network based trajectory predictors which have become the de facto standard in autonomous driving industry. Our formulation easily extends to applications like indoor navigation with mobile robots, since human trajectory predictors have structural similarity with those deployed in autonomous driving.  A practical limitation of our approach is that it requires additional computing power (in the form of GPU accelerators) to achieve real-time performance. 
\end{abstract}





\begin{figure}[t!]
    \centering
    \includegraphics[scale=0.5]{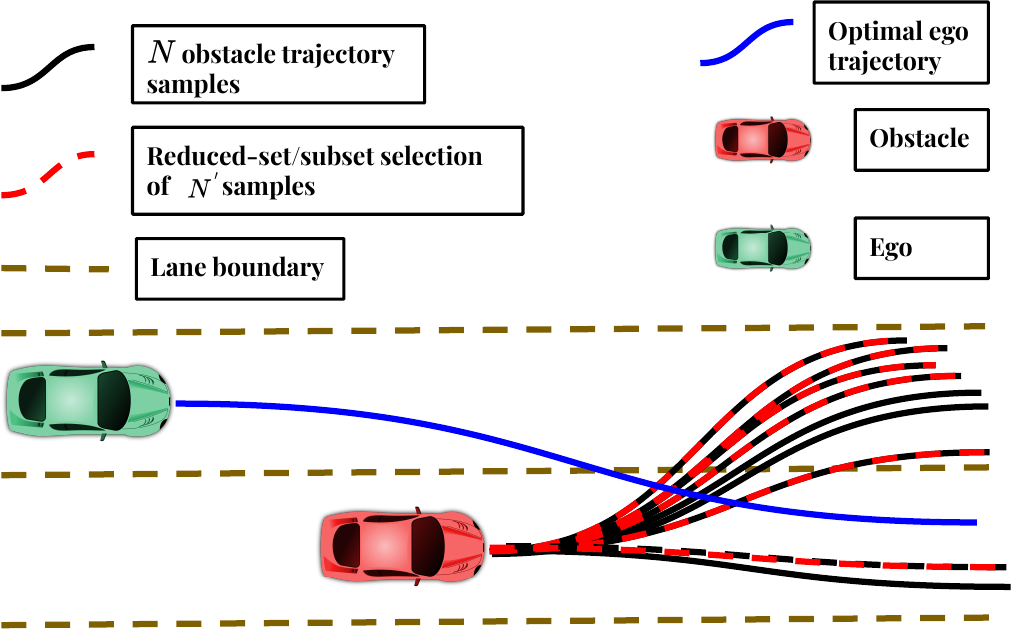}
    \caption{\footnotesize{The figure shows a scenario where an obstacle has multiple intents (lane-change vs lane-following), each associated with a trajectory distribution. However, both intents have wildly different probabilities. In this particular example, the probability of lane-change is higher. For safe navigation, the ego-vehicle needs to consider this multi-modal nature of obstacle trajectories while planning its own motions. Our proposed approach estimates the more likely samples (the reduced-set) from a set of obstacle trajectories sampled from a black-box distribution. This allows us to plan probabilistically safe motions while appropriately discriminating the low and high-probability obstacle manoeuvres.  }}
    \label{fig:iros_teaser}
    \vspace{-0.5cm}
\end{figure}

\section{Introduction}
Ensuring collision avoidance between the ego vehicle and the dynamic (neighbouring vehicles) obstacles is a fundamental requirement in autonomous driving. This in turn, requires predicting the trajectories of the latter over a time horizon. It is often difficult to predict just one deterministic trajectory for the dynamic obstacles because of the uncertainty associated with its intentions and the deployed sensors. Thus, existing works \cite{ivanovic2019trajectron}, \cite{gupta2018social},  \cite{lee2017desire} are focused on predicting a distribution of trajectories that can capture their various likely motions. 

The objective of this paper is to leverage the predictions from off-the-shelf trajectory predictors in the best possible manner for safe motion planning. Specifically, the core emphasis is on developing a risk cost that captures the probability of collision given trajectory predictions of the dynamic obstacles. This collision risk cost can then be minimized with any off-the-shelf optimizer along with other costs associated with control, smoothness and lane adherence. 

\vspace{-0.5cm}
\subsection{Existing Gaps}
\noindent There are two core challenges with regards to developing a collision risk cost. First, the analytical form of prediction distribution may be intractable or unknown. For example, the state-of-the-art trajectory predictors like \cite{ivanovic2019trajectron},\cite{roddenberry2021principledsimplicialneuralnetworks},\cite{ZAMBONI2022108252},\cite{Xu_2022_CVPR} characterize the predicted trajectory distribution in the form of deep generative models that can be arbitrary complex with multiple modes. Thus, existing collision risk derived from chance-constraints under Gaussian uncertainty \cite{zhu2019chance},\cite{5970128},\cite{9006821},\cite{9341193} become unsuitable.

Second, due to the complex nature of the predicted distribution, we often just have access to samples drawn from it. In such a case, the collision risk cost should be amenable to sample efficient estimation. That is, it should capture the true probability of collision with just a few samples of predicted obstacle trajectories. The sample efficiency is also directly related to computational run-time. The use of higher number of predicted trajectory samples translates to an increase in the number of collision checks between the ego and the obstacles.

\vspace{-0.4cm}
\subsection{Contributions}
\noindent In this paper, we propose MMD-OPT that has two novel components: (i) a sample efficient surrogate for collision risk cost that can work with arbitrary prediction distribution of obstacle motions, (ii) a trajectory optimizer for minimizing the developed surrogate cost.

The foundation of our work is built around estimating how different two given distributions are. For example, in the case of Gaussian, this can be calculated using the Kullback-Liebler (KL) divergence. However, computing the dissimilarity between arbitrary distributions is often intractable. One possible workaround is to compute the difference between two distributions in the Reproducing Kernel Hilbert Space (RKHS) \cite{simon2016consistent}, \cite{10.5555/2188385.2188410}. Specifically, given only drawn samples, we can represent the underlying distribution as a point in RKHS. Moreover, the difference between the two embedding can be easily captured by the so-called Maximum Mean Discrepancy (MMD) measure. 

\noindent \textbf{Algorithmic Contribution:} MMD-OPT uses the concept of RKHS embedding and MMD in two novel ways which also forms our core algorithmic contribution. First, given a set of predicted obstacle trajectories, we embed the underlying distribution of collision constraint residuals in RKHS. Subsequently we define a collision risk surrogate as the MMD between the RKHS embedding of collision constraint residuals and Dirac-Delta distribution. The advantage of our MMD-based surrogate is that we can systematically improve its sample efficiency by leveraging the underlying properties of RKHS embedding and MMD. To this end, we propose a bi-level optimization problem that tells us which of the obstacle trajectory samples can be disregarded without compromising the ability of our MMD-based surrogate to capture the true probability of collision (see Fig.\ref{fig:iros_teaser}). This in turn, reduces the number of collision checks required to reliably produce safe trajectories and adds to the overall sample and computational efficiency of MMD-OPT. Finally, we propose a custom sampling-based optimizer to minimize MMD-based collision risk surrogate along with driving discomfort in the form of sharp accelerations and lane boundary violations.


\noindent  \textbf{State-of-the-Art Performance:} We empirically compare planning with our MMD-based collision risk surrogate with that achieved using popular risk-cost alternatives derived from Sample Average Approximation (SAA) \cite{pagnoncelli2009sample} and Conditional Value at Risk (CVaR) \cite{yin2022riskawaremodelpredictivepath}. 
We show that our approach leads to safer trajectories for a given number of samples of obstacle trajectories. The difference is more prominent when the underlying prediction distribution departs significantly from the Gaussian form. We also demonstrate that our approach is computationally fast enough for real-time applications.

\vspace{-0.3cm}
\subsection{Organization of the Paper} 
The paper is structured as follows: \textbf{Section II} defines the motion model, formulates risk-aware trajectory optimization, and introduces exact and sample-based collision risk approximations. \textbf{Section III} provides an overview of RKHS embeddings, kernel functions, and the Maximum Mean Discrepancy (MMD) metric. \textbf{Section IV} presents MMD-OPT, including an MMD-based risk surrogate, reduced-set selection, bi-level optimization, and a sampling-based trajectory optimizer. \textbf{Section V} discusses connections to chance-constrained optimization, risk-aware planning, and RKHS applications. \textbf{Section VI} details implementation, evaluates performance in static and dynamic settings, compares MMD-OPT with CVaR and SAA, and analyzes scalability. \textbf{Section VII} concludes with key findings and future research directions.

\begin{figure}[t!]
    \centering
    \includegraphics[scale=0.49]{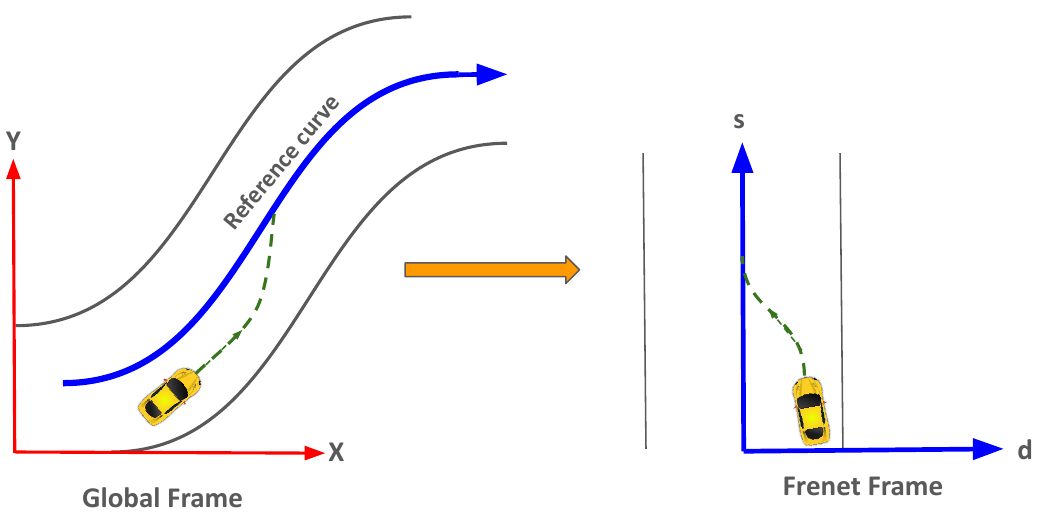}
    \caption{\footnotesize{The image shows that nonlinear reference paths (such as curvy roads) are represented by straight lines on the $s$-axis in Frenet coordinates. However, motions that do not exactly follow the reference path result in nonlinear motions in Frenet coordinates. Instead, such motions result in an offset from the reference path and therefore the $s$-axis, which is described with the $d$ coordinate.} }
    \label{fig:frenet}
    \vspace{-0.3cm}
\end{figure}

\vspace{-0.3cm}

\section{Problem Formulation and Preliminaries} \label{prelim_relatedworks}
\noindent \subsubsection*{ Symbols and Notations} Scalars will be represented by normal-font small-case letters, while bold-faced variants will represent vectors. We will use upper-case bold fonts to represent matrices. Symbol $k$ will represent the discrete-time step while $T$ will represent the transpose operator.
\vspace{-0.3cm}

\subsection{Motion Model in Frenet Frame}
\noindent We formulate trajectory planning of the ego-vehicle in the road-aligned reference frame known as the Frenet frame (Fig \ref{fig:frenet}). In this setting, the curved roads can be treated as ones with straight-line geometry. This is achieved by aligning the longitudinal and lateral motions of the ego-vehicle with the axes of the Frenet frame. Specifically, in Frenet coordinates, we use two variables, namely, $s$ representing longitudinal displacement or the distance along the reference path and $d$ representing the lateral displacement or perpendicular offset from the reference path. The $s$ coordinate starts with $s = 0 $ at the beginning of the reference path. Lateral coordinates on the reference path are represented with $d = 0$. Depending on the convention used for the Frenet frame, $d$ is positive to the left or right of the reference path.

We adopt the  following bicycle model in the Frenet frame for the ego-vehicle \cite{liniger2020safe}, where in the subscript $k$ represents the time-step.
\begin{subequations}
\begin{align}
    s_{k+1} &= s_{k}+ \dot{s}_{k}\Delta t, d_{k+1} = d_{k}+\dot{d}_{k}\Delta t \label{sdot_ddot}\\
    \psi_{k+1} &= \psi_{k}+ \dot{\psi}_{k}\Delta t, v_{k+1} = v_{k}+a_{k}\Delta t \label{psi_v}\\
    \dot{s}_{k+1} &= \frac{v_{k}\cos\psi_{k}}{1-d_{k}\kappa(s_{k})}, \dot{d}_{k+1} = v_{k}\sin\psi_{k}\\
    \dot{\psi}_{k+1} &= \frac{v_{k}\tan(\theta_{k})}{B} - \kappa(s_{k})\frac{v_{k}\cos\psi_{k}}{1-d_{k}\kappa(s_{k})} \label{psidot_k}
\end{align}
\end{subequations}

\noindent where $\psi_{k}$ is the heading of the ego-vehicle in the Frenet Frame and $B$ is the wheelbase of the car. The ego-vehicle is driven by a combination of longitudinal acceleration $a_{k}$ and steering input $\theta_{k} $. The variable $\kappa(s_{k})$ is the curvature of the reference path at the longitudinal coordinate $s_{k}$.

\noindent \textbf{Differential Flatness}\label{section_diff_flat} The motion model \eqref{sdot_ddot}-\eqref{psidot_k} is differentially flat. That is, we can represent the control inputs as a function of position and its derivatives. More precisely, we have
\begin{align}
    v_{k} &= \sqrt{(\dot{s}_{k}(1-d_{k}\kappa(s_{k}))  )^{2} + (\dot{d}_{k})^{2}} ,a_{k} = \frac{v_{k+1}-v_{k}}{\Delta t} \label{diff_flat_1}\\
    {\psi}_{k} & = \tan^{-1}(\dot{d}_{k}, \dot{s}_{k}  ), \dot{\psi}_{k} = \frac{\psi_{k+1}-\psi_{k}}{\Delta t}\label{diff_flat_2}\\
    \theta_{k} &= \tan^{-1}\left(\frac{(\dot\psi_{k} + \kappa(s_{k})\dot{s}_{k})B}{v_{k}}\right)
    \label{diff_flat_3}
\end{align}

\vspace{-0.3cm}

\newtheorem{remark}{Remark}
\begin{remark}\label{diff_flat_remark}
Eqns.\eqref{diff_flat_1}-\eqref{diff_flat_3} is leveraged later in the paper (Eqn. \eqref{planning_cost}) to express control costs over $\theta_k$  and $a_k$ in terms of position-level trajectory ($s_k, d_k$) and their derivatives.  
\end{remark}

\vspace{-0.5cm}
\subsection{Risk-Aware Trajectory Optimization}
\noindent Let $\boldsymbol{\tau}_k$ represent the obstacle position at time-step $k$. Let $\boldsymbol{\tau} \in \mathbb{R}^{2H}$ denote the obstacle trajectory formed by stacking time-stamped way-points over a planning horizon $H$. In the stochastic case, $\boldsymbol{\tau}$ is a random variable with distribution $p_{\boldsymbol{\tau}}$. With these notations in place, we can leverage differential flatness to formulate risk-aware trajectory optimization directly in the trajectory space as shown below:

\small
\begin{align}
    \min_{\mathbf{s}, \mathbf{d}} & \quad c(\mathbf{s}^{(q)}, \mathbf{d}^{(q)}) + r(\mathbf{s}, \mathbf{d}, \boldsymbol{\tau}) \label{cost} \\
    \text{s.t.} & \quad \mathbf{h}(\mathbf{s}^{(q)}, \mathbf{d}^{(q)}) = \mathbf{0} \label{eq_constraints} \\
    & \quad \mathbf{g}(\mathbf{s}^{(q)}, \mathbf{d}^{(q)}) \leq 0, \forall k \label{ineq_constraints}
\end{align}
\begin{align}
    {\mathbf{s}}{^{(q)}} &= (s^{(q)}_1, s^{(q)}_2, \dots, s^{(q)}_k, \dots, s^{(q)}_H) \nonumber \\
    {\mathbf{d}}{^{(q)}} &= (d^{(q)}_1, d^{(q)}_2, \dots, d^{(q)}_k, \dots, d^{(q)}_H), \quad q \in \{0, 1, 2\} \nonumber
\end{align}

\normalsize
\noindent The first term in the cost \eqref{cost} minimizes some function $c(\cdot)$ defined over the $q^{th}$ derivative of the position variable. We can use the differential flatness property to roll any state and control dependent cost into this term (e.g, see \eqref{planning_cost}).
The second term, $r(\cdot)$ models the collision risk of the ego-vehicle colliding with the obstacle following the predicted trajectory $\boldsymbol{\tau}$. Since obstacles move unpredictably, $\tau$ represents a random variable, which can attain many potential instantiations. The risk cost $r(\cdot)$ evaluates the likelihood of the ego-vehicle’s planned motion intersecting with any of the potential instantiations of $\tau$. The equality constraints enforce the initial and final boundary conditions on the planned ego trajectory. The inequality constraints \eqref{ineq_constraints} model the different kinematic and geometric constraints (e.g. lane boundary) that need to be satisfied by the optimal trajectory. We provide the exact mathematical form for these constraints in Section \ref{validation}. 

\begin{remark}\label{rem_1}
    For ease of exposition, the problem formulation \eqref{cost}-\eqref{ineq_constraints} only considers the case of a single neighbouring vehicle that acts as an obstacle to the ego-vehicle. Thus, we only have one random obstacle trajectory $\boldsymbol{\tau}$. The extension to arbitrary traffic density is trivial.
\end{remark}

\begin{remark}\label{basic_premise}
The Optimization \eqref{cost}-\eqref{ineq_constraints} is solved under the assumption that the probability distribution $p_{\boldsymbol{\tau}}$ of $\boldsymbol{\tau}$ can be arbitrary. But we can draw samples from $p_{\boldsymbol{\tau}}$. Moreover, sampling is relatively cheap compared to the computational cost of estimating the risk $r$(defined below) from the predicted samples of obstacle trajectories.
\end{remark}

\begin{remark}\label{opt_mpc}
The  Optimization \eqref{cost}-\eqref{ineq_constraints} represents a single instance planning problem. We can use the optimizer in a receding horizon a.k.a the model predictive control setting by solving \eqref{cost}-\eqref{ineq_constraints} from the current state at each control cycle.
\end{remark}

\vspace{-0.3cm}
\subsection{Exact Collision Risk}\label{exact_collision_risk}
\noindent Let $f_k({s}_{k}, d_{k}, \boldsymbol{\tau}_{k})$ represent the collision constraint function at time step $k$, such that $f_k\leq 0, \forall k$ ensures collision avoidance. Then, we can define the worst-case collision-constraint value ($f$) in the following manner.
\begin{align}
    f(\mathbf{s}, \mathbf{d},\boldsymbol{\tau}) = \max_k(f_k({s}_{k}, d_{k}, \boldsymbol{\tau}_{k}))
    \label{worst_case_distance}
\end{align}
where we recall that $\boldsymbol{\tau}_{k}=(s_{o,k}, d_{o,k})$ is the position of the obstacle at time-step $k$ and is obtained from the trajectory predictor. Stacking $\boldsymbol{\tau}_{k}$ at different time steps gives us $\boldsymbol{\tau}$.
There are several ways to define $f_k({s}_{k}, d_{k}, \boldsymbol{\tau}_{k})$. In the most general setting, this can be a black box function derived from 3D occupancy maps \cite{harithas2022cco}. We can also derive an analytical form for $f_k({s}_{k}, d_{k}, \boldsymbol{\tau}_{k})$ if we approximate the shape of the ego vehicle and the neighboring vehicles in terms of geometric primitives such as ellipses. 

In the stochastic setting, when $\boldsymbol{\tau}$ is a random variable, $f$ provides a distribution of the worst-case collision-constraint value. Thus, using \eqref{worst_case_distance}, we can define collision risk through \eqref{chance_cost}, wherein $P$ denotes the probability.
\begin{align}
    r = P(f(\mathbf{s}, \mathbf{d}, \boldsymbol{\tau}   ) \geq 0).
    \label{chance_cost}
\end{align}

\noindent The r.h.s of \eqref{chance_cost} maps the ego and the obstacle trajectory to the probability that the worst-case collision-constraint value is greater than zero and thus, indicating an imminent collision. Thus, minimizing $r$ within \eqref{cost}-\eqref{ineq_constraints} will produce safer trajectories. However, the main computational challenge stems from the fact that for arbitrary obstacle trajectory distribution, $p_{\boldsymbol{\tau}}$, the r.h.s of \eqref{chance_cost} does not even have an analytical expression. This necessitates the development of approximations/surrogates that can achieve a similar effect as $r$ while being computationally more tractable. This is discussed next.

\begin{figure}[!t]
    \centering
    \includegraphics[scale=0.34]{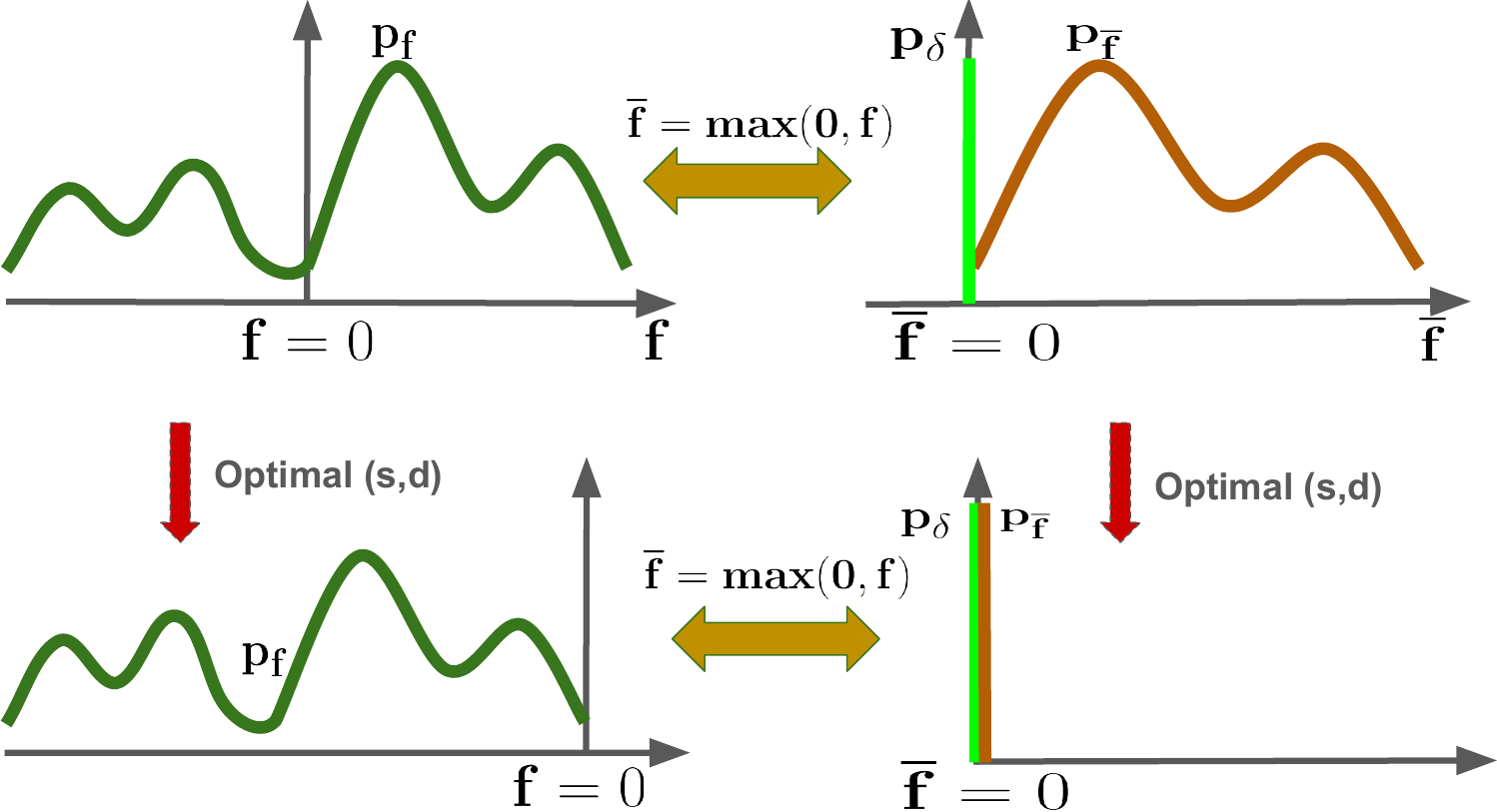}
    \caption{ \footnotesize{Figure shows the distribution ($p_f$) of worst-case collision constraint value ($f$). The distribution ($p_{\overline{f}}$) shows the distribution of collision-constraint residuals. Note that by definition, the mass of $p_{\overline{f}}$ lies to the right of $\overline{f}=0$. We want the mass of $p_f$ to lie completely to the left of $f=0$. Thus, the risk-aware trajectory optimization is essentially finding the ego-vehicle trajectory that reshapes the distribution of $p_f$ in appropriate form. Moreover, as the mass of $p_f$ shifts to the left, the distribution $p_{\overline{f}}$ converges to a Dirac-Delta distribution. } }
    \label{fig:fbar_evolution}
    \vspace{-0.3cm}
\end{figure}
\vspace{-0.5cm}
\subsection{Sample Approximation of Risk Metric}\label{sample_approx} 
\noindent In this sub-section, we will use the samples of obstacle trajectories to approximate $r$. We begin by defining the collision constraint residual function $\overline{f}$ in the following manner
\begin{align}
    \overline{f} = \max(0, f(\mathbf{s}, \mathbf{d}, \boldsymbol{\tau}))
    \label{constraint_viol_exact}
\end{align}

\noindent The $\overline{f}$ as defined in \eqref{constraint_viol_exact} maps the distribution over ego and obstacle trajectories to a distribution over collision constraint residuals. The relationship between the probability distribution of $f$ and $\overline{f}$ is shown in Fig.\ref{fig:fbar_evolution}. Now, assume that we have access to $N$ samples of obstacle trajectories ${^j}\boldsymbol{\tau}, \forall j= 1,2, \dots N $ drawn from $p_{\boldsymbol{\tau}}$. The different approximations of $r$ revolve around using the obstacle trajectories to obtain samples of $\overline{f}$ and computing appropriate empirical statistics on them.

\subsubsection{Sample Average Approximation (SAA) Risk Estimate} A simple yet effective approximation of collision risk $r$ \eqref{chance_cost} can be obtained by simply averaging the collision constraint residuals across different samples of the predicted trajectories \cite{pagnoncelli2009sample}, \cite{Shirai_2023}. That is we have
\begin{align}
    r_{SAA} &= \frac{1}{N} \sum_{j=1}^N \mathbb{I}\left({^j}\overline{f}(\mathbf{s}, \mathbf{d}, {^j}\boldsymbol{\tau} )>0\right)
\label{saa_estimate}
\end{align}

\noindent where $\mathbb{I}(\cdot)$ is an indicator function returning 1 when the conditions in the operand are satisfied and 0 otherwise. In essence, \eqref{saa_estimate} minimizes the frequency of constraint violation over the samples of predicted obstacle trajectories. Although conceptually simple, estimates of the form \eqref{saa_estimate} have proved useful in the existing literature \cite{pagnoncelli2009sample}. 

\subsubsection{Conditional Value at Risk (CVaR) based Risk Estimate} An alternative sample approximation of collision risk can be obtained by computing the empirical/sample estimate of CVaR of $\overline{f}$, using the samples 
$\{^j\overline{f}\}_{j=1}^{j=N}$ \cite{DBLP:journals/ral/LewBP24},\cite{yin2022riskawaremodelpredictivepath},\cite{shapiro2014lectures}. That is we define the risk metric as 
\begin{align}
    r_{CVaR}^{emp} = CVaR_{\alpha}^{emp}(\overline{f})
    \label{cvar_risk}
\end{align}
\noindent where $\alpha$ is some tuneable parameter associated with CVaR. We can follow the approach in \cite{yin2022riskawaremodelpredictivepath} to compute $CVaR_{\alpha}^{emp}(\overline{f})$

\section{Preliminaries on Kernels and Reproducing Kernel Hilbert Space(RKHS)}\label{prelim_rkhs_kernel}
\noindent This section introduces the concept of a kernel, feature map and Reproducing Kernel Hilbert space (RKHS) which forms the backbone of our approach MMD-OPT. For a detailed exposition of these topics, refer \cite{Muandet_2017}.
\vspace{-0.3cm}
\subsection{Feature  Maps and Kernels}\label{kernel_section}
\noindent Let $\mathcal{X}$ be a measurable input space. Consider a vector $\mathbf{z} \in \mathcal{X}$ and a  non-linear transformation $\phi$ called feature-map. We can use $\phi$ to embed $\mathbf{z}$ in the RKHS $\mathcal{H}$(\cite{Muandet_2017} Section 2.2) as $\phi(\mathbf{z})$. A positive definite real-valued kernel function $K: \mathcal{X}\times \mathcal{X}\rightarrow \mathbb{R}$ is related to the feature map through the so-called kernel trick $K(\mathbf{z}, \mathbf{z}^{\prime}) = \langle \phi(\mathbf{z}),\phi(\mathbf{z}^{\prime})\rangle_{\mathcal{H}}$ (\cite{Muandet_2017} Definition 2.1), where $\langle \cdot,\cdot \rangle_{\mathcal{H}}$ is an inner product in $\mathcal{H}$. Intuitively, the inner product provides the distance between $\mathbf{z}, \mathbf{z}^{'}$ in RKHS. The kernel trick has the following implication: to compute the inner product $\langle \phi(\mathbf{z}),\phi(\mathbf{z}^\prime)\rangle_{\mathcal{H}}$, we do not have to explicitly compute $\phi(\mathbf{z})$. Instead, we just need to evaluate the kernel function over the pair $\mathbf{z}, \mathbf{z}^{}$ as  $K(\mathbf{z}, \mathbf{z}^{\prime})$ and this will give us the required inner product. We use the Laplace Kernel throughout our implementation. It is given by:
\begin{align}
     K(\mathbf{z},\mathbf{z}^\prime) = \exp{\left(-\frac{\parallel \mathbf{z}-\mathbf{z}^\prime \parallel_1}{\sigma}\right)}
\end{align}
where $\parallel\cdot\parallel_1$ denotes the $l^1$ norm(also known as the \textit{Manhattan} or \textit{Taxicab} norm). Here $\sigma>0$ is a tuneable bandwidth parameter.
\vspace{-0.3cm}
\subsection{From data points to probability distributions}

\begin{figure}[t!]
    \centering
    \includegraphics[scale=0.385]{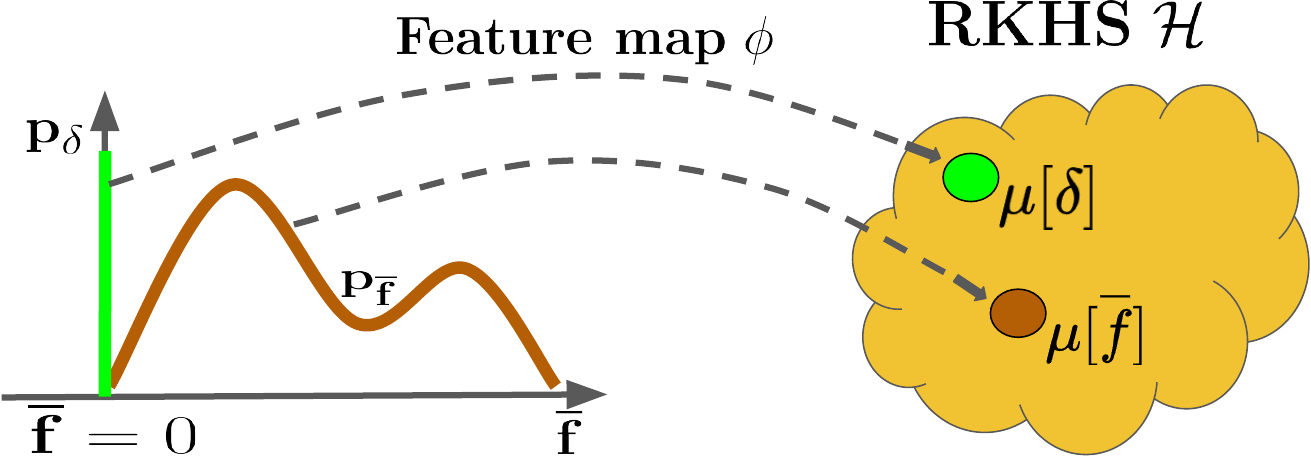}
    \caption{Kernel mean embedding of probability distributions $p_\delta$ and $p_{\overline{f}}$ into RKHS $\mathcal{H}$}
    \label{fig:rkhs}
    \vspace{-0.3cm}
\end{figure}

\noindent We can generalize the concept of a feature map, kernel function and RKHS embedding to work on probability distributions as well (see Figure \ref{fig:rkhs}). Let $u$ be a random variable with probability distribution $p_u$. The RKHS embedding of $u$ ( or $p_u$) is given by the following.
\begin{subequations}
\begin{align}
    \mu[u] = E_{u\sim p_{u}}[\phi(u)] = E_{u\sim p_{u}}[K(u,\cdot)] \label{rkhs_embedding_expectation} \\
    = \int K(u, \cdot)dp_{u}(u) \label{rkhs_embedding_integral}
\end{align}
\end{subequations}
\vspace{-0.3cm}

\noindent Essentially, \eqref{rkhs_embedding_expectation}-\eqref{rkhs_embedding_integral} map probability distribution to a function in RKHS. For arbitrary distribution $p_u$, it is often intractable to compute the r.h.s of \eqref{rkhs_embedding_integral}. Thus, instead, we can resort to the empirical estimate of the RKHS embedding. Let ${^j}u$ be samples of $u$ drawn from $p_u$. The empirical estimate of $\mu[u]$ is given by the following.
    \begin{align}
        \widehat{\mu}[u]:= \sum_{j=1}^N \frac{1}{N}\phi(^{j}{u}) = \sum_{j=1}^N \frac{1}{N}K(^{j}{u},\cdot)
        \label{mu_u}
    \end{align}
    
\subsubsection*{\textbf{Functions of Random Variables}} We can extend the concept of RKHS embedding to functions of random variables as well. Let $g(u)  \in \mathcal{X}\rightarrow \mathcal{Z}$ be an arbitrary function of a random variable $u$ with $p_{g}$ denoting the probability distribution of $g(u)$. Its RKHS embedding $\mu[g]$ and its empirical estimate $\hat{\mu}[g]$ is given by 
\begin{subequations}
\begin{align}
    \mu[g] = E_{u\sim p_{u}}[\phi(g(u))] = E_{u\sim p_{u}}[K(g(u),\cdot)] \label{rkhs_embedding_expectation_g}\\
    = \int K(g(u), \cdot)dp_{u}(u), \label{rkhs_embedding_integral_g}\\
    \widehat{\mu}[g(u)]:= \sum_{j=1}^N \frac{1}{N}\phi({g(^{j}u)}) = \sum_{j=1}^N \frac{1}{N}K({g(^{j}u)},\cdot)
    \label{mu_g}
\end{align}
\end{subequations}
\noindent Note that $K$ in the above expressions \eqref{rkhs_embedding_expectation_g}-\eqref{mu_g} is the same as that used in \eqref{mu_u}.

One of the attractive features of RKHS embedding is that when constructed with characteristic Kernels (Gaussian, Laplacian, etc.), it can capture information about mean and moments of $u$ and $g(u)$ up to infinite order. Furthemore, $\widehat{\mu}[u], \widehat{\mu}[g]$ are consistent estimators. That is they converge to the true RKHS embedding as we increase the sample size $N$. The RKHS embeddings $\widehat{\mu}[u], \widehat{\mu}[g]$ are also tightly coupled through the following theorem from \cite{simon2016consistent}.

\theoremstyle{definition}
\newtheorem{theorem}{Theorem}
\begin{theorem} \label{consistent_theorem}
    Suppose $g$ is a continuous function and the kernel functions are continuous and bounded. Then, the following holds
    \begin{align}
        \text{if} \hspace{0.2cm} \hat{\mu}[u] \rightarrow \mu[u], \hspace{0.2cm}\text{then} \hspace{0.2cm} \hat{\mu}[g] \rightarrow \mu[g]
    \end{align}
\end{theorem}
\noindent The proof can be found in the Appendix of \cite{simon2016consistent}.


\vspace{-0.3cm}

\subsection{Maximum Mean Discrepancy and Improving Sample Efficiency} \label{mmd_prelim}
\noindent One of the core applications of RKHS embedding lies in quantifying similarity/difference between two probability distributions through a concept called Maximum Mean Discrepancy (MMD) defined in the following manner.

\theoremstyle{definition}
\newtheorem{definition}{Definition}
\begin{definition}
    Let $u$ and $\Tilde{u}$ be two random variables with probability distributions $p_u$ and $p_{\Tilde{u}}$ respectively. The MMD between the two distributions is given by 
    \begin{dmath}
        MMD(p_u, p_{\Tilde{u}}) = \left\Vert \mu[u]- {\mu}[\Tilde{u}\right\Vert_{\mathcal{H}}^2 = \left\langle \mu[u]- {\mu}[\Tilde{u}], \mu[u]- {\mu}[\Tilde{u}]\right\rangle 
    \end{dmath}
    \label{MMD}
\end{definition}
\noindent where $\left\Vert(.)\right\Vert_{\mathcal{H}}$ is the RKHS norm and can be expressed through inner products, which in turn can be evaluated using the Kernel trick. The distribution $p_u$ and $p_{\Tilde{u}}$ are same if and only if  $\left\Vert \mu[u]- {\mu}[\Tilde{u}\right\Vert_{\mathcal{H}}^2 =0$ \cite{simon2016consistent}, \cite{scholkopf2007kernel}. 
Typically, it is intractable to obtain the exact MMD and thus, we resort to its empirical estimate given by the following
  \begin{align}
        MMD^{emp}(p_u, p_{\Tilde{u}})  =  \left\Vert \hat{\mu}[u]- \hat{\mu}[\Tilde{u}\right\Vert_{\mathcal{H}}^2
    \end{align}
    \label{MMD_emp}
\vspace{-0.7cm}
\begin{figure*}[t!]
    \centering
    \includegraphics[scale=0.7]{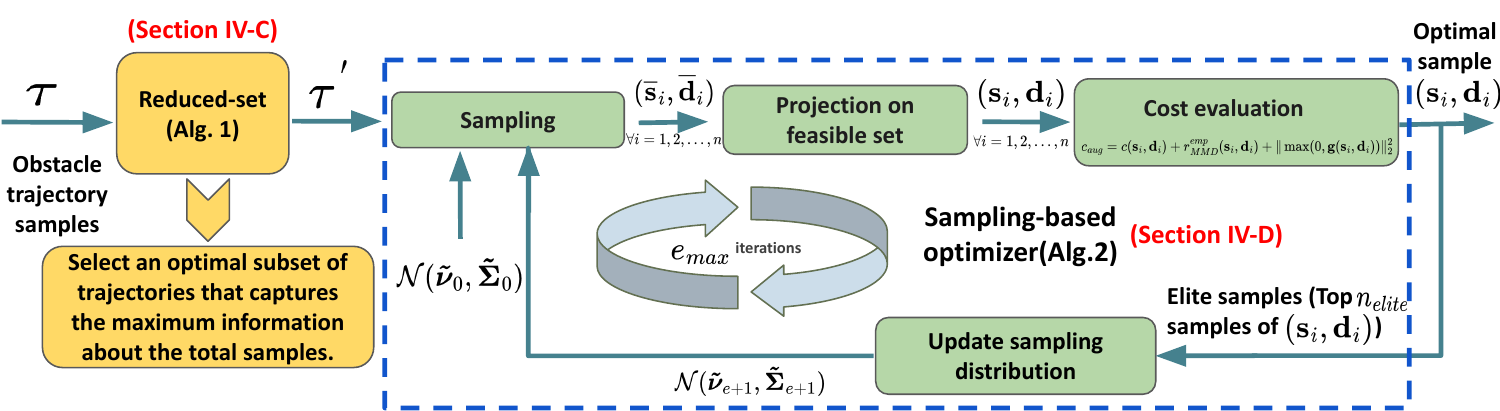}
    \caption{ \footnotesize{Overview of MMD-OPT. It consists of two main parts. The first part is the so-called \textit{reduced-set} algorithm that tells us which obstacle trajectories can be discarded without compromising the ability of our MMD-based surrogate to capture the collision risk in a given scene. The second part is a sampling-based optimizer the minimizes the MMD-based collision risk cost. Our sampling based optimizer also consists of novel components that exploits the problem structure for computational efficiency. } }
    \label{fig:pipeline}
    \vspace{-0.3cm}
\end{figure*}
\section{Main Algorithmic Results}
\noindent This section presents our main algorithmic contributions. An overview of our approach is shown in Fig. \ref{fig:pipeline}. The cornerstone of our work is the \textit{reduced-set selection} (Alg.\ref{algo_1}) which allows us to use only a subset of predicted obstacle trajectories to compute our MMD-based collision risk surrogate. This in turn, drastically reduces the number of collision-checks that we need to perform. The optimal \textit{reduced-set} is then passed to the downstream sampling-based optimizer (Alg.\ref{algo_2}) wherein we sample ego-vehicle trajectories in the Frenet-frame from a Gaussian distribution and evaluate their associated costs (including collision risk). The sampling distribution is updated in an iterative fashion to drive the costs down at each iteration.  

In the next few sub-sections, we present the different components from Fig.\ref{fig:pipeline}. We begin by presenting our surrogate for collision risk.
\vspace{-0.5cm}

\subsection{Collision Risk as Difference of Two Distributions}
\noindent The distribution of ${f}$ and $\overline{f}$, denoted by $p_f$ and $p_{\overline{f}}$ respectively, cannot be characterized analytically, even if we assume that the predicted trajectory of the neighbouring vehicle is drawn from a Gaussian distribution. In general, $p_{\boldsymbol{\tau}}$ can be arbitrarily complex, making it even more challenging to make any assumption on the nature of $p_f$ and $p_{\overline{f}}$. However, we can intuitively reason about some desired characteristics that the latter two distribution should have. 

\begin{itemize}
    \item As shown in Fig.\ref{fig:fbar_evolution}, we want  to shift the mass of distribution  $p_f$ to the left of $f = 0$. The higher the mass is to the left of $f = 0$, the lower the risk $r= P(f\geq 0)$. 
    \item As the mass of $p_f$ gets shifted to the left of $f = 0$, $p_{\overline{f}}$ converges to a Dirac-Delta distribution $p_{\delta}$. This follows directly from the definition of $\overline{f}$  in \eqref{constraint_viol_exact}. 
    \item We can re-imagine risk-aware trajectory optimization as the problem of finding the optimal ego vehicle trajectory $(\mathbf{s}, \mathbf{d})$ that re-shapes $p_{\overline{f}}$ to look similar to a Dirac-Delta distribution. This, in turn, would ensure that the mass of $p_{f}$ gets shifted to the left of $f=0$.
\end{itemize}

\noindent Based on the above insights, we propose the following novel surrogate for collision risk
\begin{align}
r_{MMD} = MMD(p_{\overline{f}}, p_{\delta}) = \left\Vert \mu[\overline{f}]-\mu[\delta]\right\Vert_{\mathcal{H}}^2
\label{r_dist}
\end{align}

\noindent where $\mu[\overline{f}]$ is the RKHS embedding of $\overline{f}$ (or $p_{\overline{f}}$) defined in the following manner
\begin{align}
     \mu[\overline{f}] = \int K(\overline{f}, .)dp_{\overline{f}}(\overline{f}), \quad \widehat{\mu}[\overline{f}] = \sum_{j=1}^{N} \frac{1}{N}\phi({^j}\overline{f})
      \label{mu_f_bar}
\end{align}

\noindent It is connected to the RKHS embedding of obstacle trajectory distribution $p_{\boldsymbol{\tau}}$ defined as:
\begin{align}
     \mu[\boldsymbol{\tau}] = \int K(\boldsymbol{\tau}, .)dp_{\boldsymbol{\tau}}(\boldsymbol{\tau}), \quad \widehat{\mu}[\boldsymbol{\tau}] = \sum_{j=1}^{N} \frac{1}{N}\phi(^j\boldsymbol{\tau})
      \label{mu_tau}
\end{align}

\noindent Similarly, $\mu[\delta]$ is the RKHS embedding of a random variable $\delta \sim p_{\delta}$. As $r_{MMD}$ decreases, $p_{\overline{f}}$ becomes similar to $p_{\delta}$. In the limiting case, as $r_{MMD} \rightarrow 0$, the distribution $p_{\overline{f}}\rightarrow p_{\delta}$ and consequently $P(f\geq 0)\rightarrow 0$.

Since the true $r_{MMD}$ is intractable to compute for arbitrary $p_{\boldsymbol{\tau}}$, we will be working with the following empirical estimate computed through sample-level information
\begin{align}
    r_{MMD}^{emp}  = \left\Vert \hat{\mu}[\overline{f}]-\hat{\mu}[\delta]\right\Vert_{\mathcal{H}}^2
    \label{risk_mmd}
\end{align}



\begin{remark} \label{mmd_consistency}
     By Theorem \ref{consistent_theorem}, $\hat{\mu}[\boldsymbol{\tau}]\rightarrow{\mu}[\boldsymbol{\tau}]$ implies $\hat{\mu}[\overline{f}]\rightarrow{\mu}[\overline{f}]$ and consequently $r_{MMD}^{emp}\rightarrow r_{MMD}$
\end{remark}

\begin{remark} \label{gausss_approx_delta}
    We can approximate $p_{\delta}$ through a Gaussian $\mathcal{N}(0, \epsilon)$, with an extremely small covariance $\epsilon$ ($\approx 10^{-5}$).
\end{remark}

\begin{algorithm*}[!t]
\caption{Reduced-Set Selection and Weight and Kernel Parameter Optimization using Cross-Entropy Method (CEM)}
\label{algo_1}
\SetAlgoLined
$e_{cem}$ = Maximum number of iterations, ${N}$= size of ${\mathbf{O}}$, $N^'$ = size of \textit{reduced-set}  \\
Initialize mean and covariance $\boldsymbol{\nu}_e, \boldsymbol{\Sigma}_e$, respectively, at $e=0$ for sampling and sorting $\boldsymbol{\lambda}$ and the kernel $\sigma$ parameter respectively .\\
\For{$e=1, e \leq e_{cem}, e++$}
{
Draw ${n}_{cem}$ samples $( {^{1}}\boldsymbol{\lambda}, {^{2}}\boldsymbol{\lambda}, {^{m}}\boldsymbol{\lambda}, \dots {^{n_{cem}}}\boldsymbol{\lambda})$ and  $( {^{1}}\sigma, {^{2}}\sigma, {^{m}}\sigma, \dots {^{n_{cem}}}\sigma)$ from $\mathcal{N}(\boldsymbol{\nu}_e, \boldsymbol{\Sigma}_e)$. \\
${^{m}}\boldsymbol{\lambda} = ({^{m, 1}}\lambda, {^{m, 2}}\lambda, \dots {^{m, {N}}}\lambda) $ \tcp*[f]{\textcolor{blue}{The dimension of ${^{m}}\boldsymbol{\lambda}$ is equal to the rows of ${\mathbf{O}}$} } \\
 \vspace{0.1cm}
Initialize $CostList$ = [] \\
 \vspace{0.1cm}
 
\For{$m=1, m \leq {n}_{cem}, m++$} 
{
$ ({^1}\boldsymbol{\tau}^', {^2}\boldsymbol{\tau}^', {^l}\boldsymbol{\tau}^', \dots, {^{N^'}}\boldsymbol{\tau}^'  )  = F_{{^m}\boldsymbol{\lambda}}({\mathbf{O}})$  \tcp*[f]{\textcolor{blue}{Reduced-Set selection formulated as sorting of the elements of ${^m}\boldsymbol{\lambda}$}} \\
$ {^l}\beta^* \gets $ Solve optimization \eqref{bi_level_inner_red_cost}  with the \textit{reduced-set} formed above and with kernel parameter ${^m}\sigma$ \\
$cost_{mmd} \gets$ Compute MMD cost in \eqref{bi_level_inner_red_cost} using $ {^l}\beta^*$ \\
\vspace{0.1cm}
append $cost_{mmd}$ to $CostList$ \\
\vspace{0.1cm}       
}
$EliteSet  \gets$ Select top $n_{elite}$ samples of $({^{m}}\boldsymbol{\lambda})_{m=1}^{m=n_{cem}}$ and $({^{m}}\sigma)_{m=1}^{m=n_{cem}}$  with lowest cost from $CostList$.\\
$\boldsymbol{\nu}_{e+1}, \boldsymbol{\Sigma}_{e+1} \gets $ Update distribution based on $EliteSet$ \tcp*[f]{\textcolor{blue}{Empirical mean and covariance estimate from the ElliteSet samples }}
}
\Return{ $({^1}\boldsymbol{\tau}^', {^2}\boldsymbol{\tau}^', \dots, {^{N^'}}\boldsymbol{\tau}^'  )$, weights ${^l}\beta^*$ and optimal kernel hyperparameter $\sigma$ corresponding to lowest cost in the $EliteSet$}
\end{algorithm*}
\subsubsection*{Finite Sample Guarantees}
\noindent A key advantage of $r_{MMD}^{emp}$ is that it inherits the excellent convergence and finite sample guarantees associated with RKHS embedding and MMD. Specifically, how far $r_{MMD}^{emp}$ is from the true $r_{MMD}$ can be quantified by the following Theorem that is a direct application of the result presented in \cite{10.5555/2188385.2188410}.

\begin{theorem} \label{finite_sample_guarantee}
    Let $\overline{f}, \delta$ be the collision constraint violation and Dirac-Delta random variables respectively. Let $p_{\overline{f}}, p_{\delta}$ be the respective probability distributions of $\overline{f}$ and $\delta$. Given samples $X:=\{^{j}\overline{f}\}_{j=1}^{j=N}$ and $Y:=\{\delta_{j}\}_{j=1}^{j=N}$ i.i.d from $p_{\overline{f}}$ and $ p_{\delta}$ , respectively, and assuming $0\leq K(\xi_1, \xi_2)\leq c_{o}, \forall \xi_1, \xi_2$ and some constant $c_{o}$, then:
    \small    
    \begin{align}
        P_{X,Y} \biggl\{ |r_{MMD}^{emp} - r_{MMD} | \leq 2\Bigl(   2(c_{o}/N)^{\frac{1}{2}} + \varepsilon \Bigr)   \biggr\}\nonumber \\ > 1-\exp\Bigl(\frac{-\varepsilon^2N}{4c_{o}}\Bigr) 
    \end{align}
    \normalsize
    where $P_{X,Y}$ denotes the probability over the $N$-sample $X$ and $Y$, and $\varepsilon>0$ is an arbitrary parameter controlling the probability bound. $K(x,y)$ denotes the kernel function.
\end{theorem}
\noindent Theorem \ref{finite_sample_guarantee} ensures that as the sample size $N$ increases, the probability that the absolute difference between $r_{MMD}^{emp}$ and $r_{MMD}$ tends to a small value approaches $1$. In other words, with increase in samples, not only the estimate $r_{MMD}^{emp}$ improves, the stochasticity or the variance associated with it also reduces. In Section \ref{validation}, we show Theorem \ref{finite_sample_guarantee} directly translates to lower worst-case collision-rate with $r_{MMD}^{emp}$ as compared to $r_{CVaR}^{emp}$ and $r_{SAA}$.

\vspace{-0.3cm}
\subsection{Advantages of $r_{MMD}^{emp}$ as a Collision Risk Surrogate}
\noindent Just like $r_{SAA}$ (Eqn. \eqref{saa_estimate}) and $r_{CVaR}$ (Eqn. \eqref{cvar_risk}), our $r_{MMD}^{emp}$ also requires evaluating the collision constraint residual function $\overline{f}$ over predicted obstacle trajectories. This can be computationally intensive if the collision checks are expensive. The key feature that differentiates $r_{MMD}^{emp}$ from $r_{SAA}$ and $r_{CVaR}^{emp}$ is that we can use the tools from RKHS embedding and MMD to systematically improve its sample complexity. In other words, we can use only a subset $N^{'}<<N$ of the predicted obstacle trajectories to evaluate $\overline{f}$ and yet incur only minimum degradation in estimation of $r_{MMD}^{emp}$. We call this subset as \textit{reduced-set} following the notation from \cite{simon2016consistent}, \cite{10.5555/2188385.2188410}.

Let $^{l}\boldsymbol{\tau}^{'}, l = 1,2, 3, \dots N^{'}$ be the \textit{reduced-set} formed by sub-selecting $N^{'}$ samples out of $N$ samples of $\boldsymbol{\tau}$. Moreover, let us attach weights ${^l}\beta$ with every $^{l}\boldsymbol{\tau}^{'}$ such that $\sum_l {^l}\beta = 1$. Let ${^l}\overline{f^'}$ be the samples of $\overline{f}$ evaluated on $^{l}\boldsymbol{\tau}^{'}$. The empirical RKHS emebeddings of $p_{\boldsymbol{\tau}}$ and $p_{\overline{f}}$ computed using  $^{l}\boldsymbol{\tau}^{'}$ and  ${^l}\overline{f^'}$ are given by :
\begin{align}
    \hat{\mu}[\boldsymbol{\tau}^{'}] = \sum_{l}{^l}\beta\phi({^l}\boldsymbol{\tau}^{'}), \qquad  \hat{\mu}[\overline{f^'}] = \sum_{l}{^l}\beta\phi({^l}\overline{f^'})\label{mu_fbar_red_set}
\end{align}

Now, \cite{simon2016consistent} showed that Theorem \ref{consistent_theorem} can also be leveraged to ensure that if $\hat{\mu}[\boldsymbol{\tau}^{'}]$ is close to $\hat{\mu}[\boldsymbol{\tau}]$ in MMD metric, then $\hat{\mu}[\overline{f'}]$ will be close to $\hat{\mu}[\overline{f}]$ in the same metric. Consequently, $r_{MMD}^{emp}$ computed using $\hat{\mu}[\overline{f'}]$ which requires $N^{'}$ predicted obstacle trajectories will be closer to that computed using  $\hat{\mu}[\overline{f}]$ on the larger set of  $N$ samples. Alternately, if we reduce the sample size in a way that we minimize $\left\Vert \hat{\mu}[\boldsymbol{\tau}^{'}]- \hat{\mu}[\boldsymbol{\tau}]\right\Vert_{\mathcal{H}}^2 $,  the loss of accuracy in estimation of  $r_{MMD}^{emp}$ will be minimal. In the next sub-section, we present several ways by which this can be achieved.
\vspace{-0.4cm}

\subsection{Reduced-Set}\label{reduced_set}
\noindent There are three ways to minimize $\left\Vert \hat{\mu}[\boldsymbol{\tau}^{'}]- \hat{\mu}[\boldsymbol{\tau}]\right\Vert_{\mathcal{H}}^2$. First, we can choose the \textit{reduced-set} in a careful manner. Second, we optimize the values of ${^l}\beta$. Finally, we can also the choose the right set of kernel parameters $\sigma$. In the subsequent subsections, we first present a simple method that adopts only the second approach and is a straightforward application of \cite{simon2016consistent}. Subsequently, we present our key result that adopts all three ways to minimize $\left\Vert \hat{\mu}[\boldsymbol{\tau}^{'}]- \hat{\mu}[\boldsymbol{\tau}]\right\Vert_{\mathcal{H}}^2$.

\subsubsection{Fixed Reduced-Set, Optimal Weights} \label{fixed_red_set} In this approach, we assume that the $N^'$ \textit{reduced-set} samples ${^l}\boldsymbol{\tau}^{'}$ are simply a random sub-selection from $N$ samples of  $\boldsymbol{\tau}$. We then optimize the weights of the ${^l}\boldsymbol{\tau}^{'}$ through the following optimization problem.
\begin{subequations}
   \begin{align}
    \min_{{^l}\beta} \left\Vert \frac{1}{N}\sum_{i = 1}^{i = N } \phi(^{i}{\boldsymbol{\tau}})-\sum_{l = 1}^{l = {N}^' } {^l}\beta\phi({^{l}\boldsymbol{\tau}^'}) \right\Vert_{\mathcal{H}}^2  \label{red_set_tau_cost}\\
    \sum_l {^l}\beta = 1 \label{red_set_tau_const}
\end{align} 
\end{subequations}

Using the kernel-trick, optimization \eqref{red_set_tau_cost}-\eqref{red_set_tau_const} can be rephrased as an equality-constrained QP. Essentially, \eqref{red_set_tau_cost}-\eqref{red_set_tau_const} can be seen as distilling the information from a significantly larger sample set of predicted obstacle trajectories to the \textit{reduced-set} by re-weighting the importance of the samples of the latter. 

\subsubsection{Optimal Reduced-Sets, Weights and Kernel Parameters}
In this approach, we simultaneously optimize the selection of the \textit{reduced-set}, the weights of its individual samples and kernel parameters. Our approach takes the form of a bi-level optimization \eqref{bi_level_red_outer}-\eqref{bi_level_inner_red_cost}, wherein $\mathbf{O}$ is a matrix with $N$ rows, each containing a sample of $\boldsymbol{\tau}$ drawn from $p_{\boldsymbol{\tau}}$. 
\begin{subequations}
    \begin{align}\label{bi_level_red_outer}
    \min_{\boldsymbol{\lambda}, \sigma} & \left\Vert  \frac{1}{{N}}\sum_{i = 1}^{i = {N} } \phi(^{i}{\boldsymbol{\tau}})-\sum_{l = 1}^{l = {N}^' } {^l}\beta^*\phi(^l\boldsymbol{\tau}^') \right\Vert_{\mathcal{H}}^2  \\
    & F_{\boldsymbol{\lambda}}({\mathbf{O}}) = (^{1}\boldsymbol{\tau}^', ^{2}\boldsymbol{\tau}^', \dots, ^{N'}\boldsymbol{\tau}^'  ) \\
        & {^l}\beta^*= 
        \begin{aligned}[t]\label{bi_level_inner_red_cost}
        &&& \arg\min_{ {^l}\beta  }  \left\Vert  \frac{1}{{N}}\sum_{i = 1}^{i = {N} } \phi(^{i}{\boldsymbol{\tau}})-\sum_{l = 1}^{l = {N}^' } {^l}\beta\phi(^l\boldsymbol{\tau}^') \right\Vert_{\mathcal{H}}^2  \\
            &&&\st  \sum{{^l}\beta} = 1
     \end{aligned}
\end{align}
\end{subequations}

\noindent In \eqref{bi_level_red_outer}-\eqref{bi_level_inner_red_cost}, $F_{\boldsymbol{\lambda}}$ is a selection function that chooses $N^'$ rows out of $\mathbf{O}$ to form the \textit{reduced-set}. It is parameterized by $\boldsymbol{\lambda}$. That is, different $\boldsymbol{\lambda}$ leads to different \textit{reduced-set}. We discuss the algebraic form for  $F_{\boldsymbol{\lambda}}$ later in this section. But, it is worth pointing out that if we  fix $\boldsymbol{\lambda}$ and consequently $F_{\boldsymbol{\lambda}}$ (e.g just a random $\boldsymbol{\lambda}$ ) and the kernel parameter $\sigma$, then we recover the earlier approach described in  \eqref{red_set_tau_cost}-\eqref{red_set_tau_const}.

As can be seen, the inner optimization \eqref{bi_level_inner_red_cost} is defined over just the weights ${^l}\beta$ for a fixed $\boldsymbol{\lambda}$ and kernel parameter $\sigma$. The outer optimization in turn optimizes in the space of $\boldsymbol{\lambda}$ and kernel parameter in order to reduce the MMD cost associated with optimal  ${^l}\beta^*$. Our bi-level approach provides two pronged benefits. First, it gets rid of the variance associated with randomly selecting the \textit{reduced-set}. Second, our approach leads to $\hat{\mu}[\boldsymbol{\tau}^']$ that has lower MMD difference from $\hat{\mu}[\boldsymbol{\tau}]$ as compared to that achieved from the previous method of Section \ref{fixed_red_set}. This in turn, improves the embedding of $\hat{\mu}[\overline{f^'}]$ and consequently the estimate of $r_{MMD}^{emp}$.

\noindent\textbf{Selection Function $F_{\boldsymbol{\lambda}}$:} Let $\boldsymbol{\lambda}\in \mathbb{R}^{{N}}$ be an arbitrary vector and $\lambda_t$ be its $t^{th}$ element. Assume that $\vert \lambda_t\vert $ encodes the value of choosing the $t^{th}$ row of $\mathbf{O}$ to form the \textit{reduced-set}. That is, the larger the $\vert \lambda_t\vert $, the higher the impact of choosing the $t^{th}$ row of ${\mathbf{O}}$ in minimizing \eqref{bi_level_red_outer}. With these constructions in place, we can now define $F_{\boldsymbol{\lambda}}$ through the following sequence of mathematical operations.
\begin{subequations}
    \begin{align}
        \mathcal{S} = Argsort\{\vert\lambda_1\vert, \vert\lambda_2\vert, \dots \vert\lambda_{N}\vert)\} \label{argsort}\\
        \mathbf{O}^{'} = [ {\mathbf{O}}_{t_1}, {\mathbf{O}}_{t_2}, \dots, {\mathbf{O}}_{t_{N}}  ], \forall t_e \in \mathcal{S} \label{sort}  \\
        F_{\boldsymbol{\lambda}}({\mathbf{O}}) = \mathbf{O}^{'}_{t_{N-N^'}: t_{N}} =  ({^1}\boldsymbol{\tau}^', {^2}\boldsymbol{\tau}^', \dots, {^{N^'}}\boldsymbol{\tau}^'  )\label{selection}
    \end{align}
\end{subequations}

\noindent As can be seen, we first sort in increasing value of $\vert\lambda_t\vert$ and compute the required indices $t_e$. We then use the same indices to shuffle the rows of $\mathbf{O}$ in \eqref{sort} to form an intermediate variable $\mathbf{O}^{'}$. Finally, we choose the last $N^'$ rows of $\mathbf{O}^{'}$ as our \textit{reduced-set} in \eqref{selection}. Intuitively, \eqref{argsort}-\eqref{selection} parameterizes the sub-selection of the rows of ${\mathbf{O}}$ to form the \textit{reduced-set}. That is, different $\boldsymbol{\lambda}$ gives us different \textit{reduced-sets}. Thus, a large part of solving \eqref{bi_level_red_outer}-\eqref{bi_level_inner_red_cost} boils down to arriving at the right $\boldsymbol{\lambda}$. We achieve this through a combination of gradient-free search and quadratic programming (QP). We present the details of our approach next.

\begin{figure*}[h!]
    \centering
    \includegraphics[scale=0.3865]{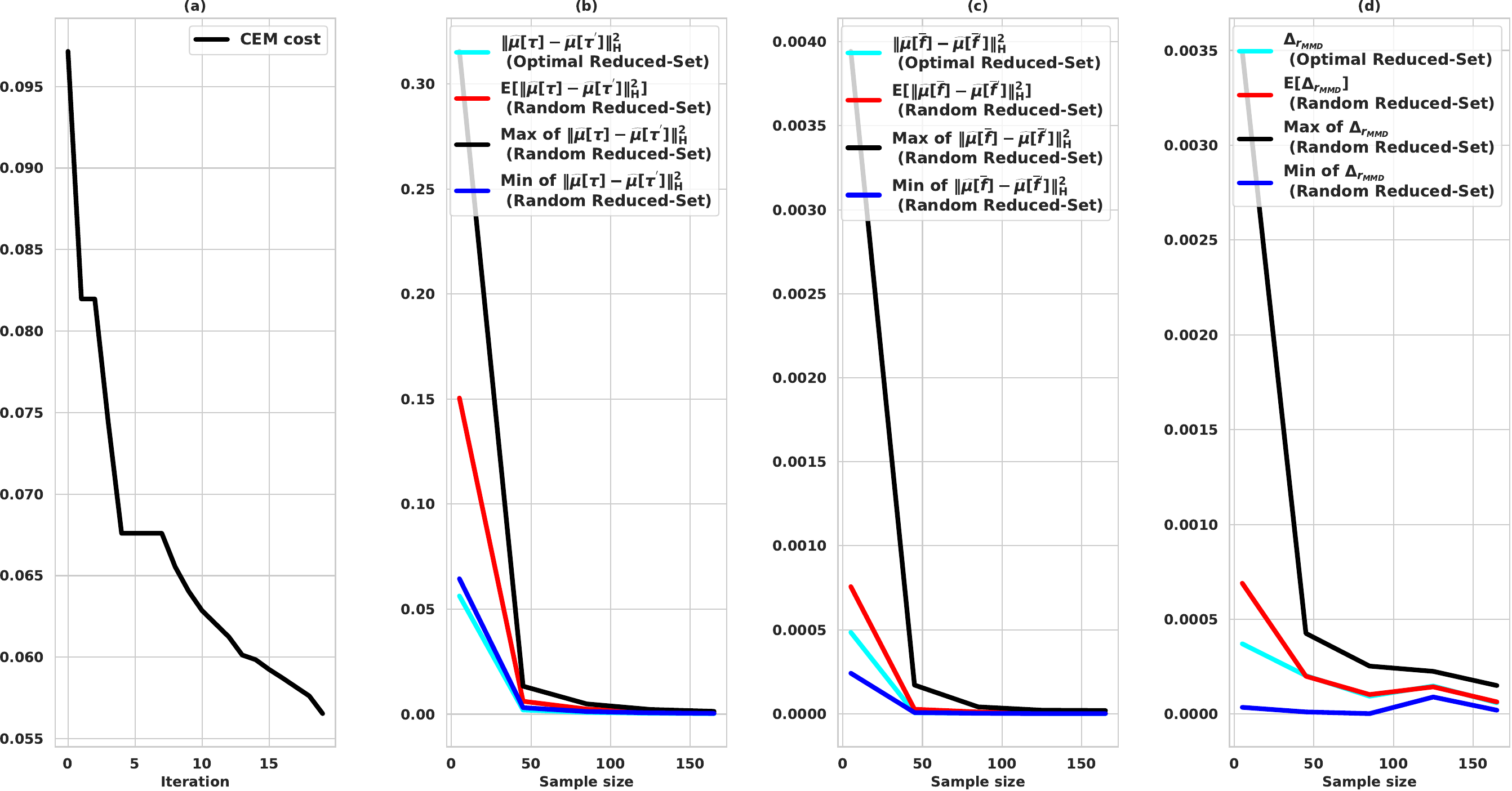}
    \caption{ \footnotesize{Validation of performance of Alg.\ref{algo_1}. Fig. (a) shows the progression of outer-level cost \eqref{bi_level_red_outer} across iterations. Fig. (b) validates the efficacy of our optimal \textit{reduced-set} selection. We sampled different random \textit{reduced-sets} and computed the average, minimum and maximum of MMD $\Vert \hat{\mu}[\boldsymbol{\tau}^{'}]- \hat{\mu}[\boldsymbol{\tau}]\Vert_{\mathcal{H}}^2 $. As shown in Fig. (b), our optimal \textit{reduced-set} leads to an $\Vert \hat{\mu}[\boldsymbol{\tau}^{'}]- \hat{\mu}[\boldsymbol{\tau}]\Vert_{\mathcal{H}}^2 $ that is very close to what can be achieved by exhaustively trying out different randomly selected \textit{reduced-set}.  Fig. (b) validates our hypothesis based on Theorem \ref{consistent_theorem} and insights from \cite{simon2016consistent}. As  $\hat{\mu}[\boldsymbol{\tau}^']\rightarrow \hat{\mu}[\boldsymbol{\tau}] $, the MMD $\Vert \hat{\mu}[\overline{f}]- \hat{\mu}[\overline{f^'}]\Vert_{\mathcal{H}}^2 \rightarrow 0$. We can once again observe that our optimal \textit{reduced-set} performs close to the best-case performance than can be hoped to be achieved by trying-out different random \textit{reduced-sets}. Fig. (c) verifies that our optimal \textit{reduced-set} selection ensures minimal loss in estimation of $r_{MMD}^{emp}$ while reducing the sample size. To this end, we computed $\Delta_{r_{MMD}}$ through \eqref{e_mmd} which captures the estimation error between $r_{MMD}^{emp}$ computed over a large set and the smaller \textit{reduced-set}. As can be seen, $r_{MMD}^{emp}$ computed over the optimal \textit{reduced-set} samples perform close to the average performance that can be achieved by exhaustively trying out different random choices of the \textit{reduced-set}. It should be noted that in real-world setting, exhaustive sampling is not possible and our bi-level optimization \eqref{bi_level_red_outer}-\eqref{bi_level_inner_red_cost} provides a one-shot solution that is as good as exhaustive sampling in expectation.    }   }
    \label{fig:red_set_analysis}
    \vspace{-0.3cm}
\end{figure*}

\noindent{\textbf{Solution Process}:} Our approach for solving \eqref{bi_level_red_outer}-\eqref{bi_level_inner_red_cost} is  presented in Alg.\ref{algo_1} that combines gradient-free Cross-Entropy Method (CEM) \cite{botev2013cross} and QP. The input to the algorithm is the set ${\mathbf{O}}$ and the required size of the reduced-set $N^'$. The output is the \textit{reduced-set} samples, their weights and the optimal kernel parameter $\sigma$. Alg.\ref{algo_1} begins by initializing a Gaussian distribution (Line 2) with mean $\boldsymbol{\nu}_e$ and covariance $\boldsymbol{\Sigma}_e$ at iteration $e = 0$. The loop defined between lines 3-15 is the outer CEM optimizer. At each iteration of CEM, we sample $n_{cem}$ batch of $\boldsymbol{\lambda}$ and $\sigma$. Note that the dimension of $\boldsymbol{\lambda}$ is equal to the number of obstacle trajectory samples contained in ${\mathbf{O}}$. Subsequently, between Lines 7-12, we loop through each sampled ${^m}\boldsymbol{\lambda}$ and ${^m}\sigma$ and use them to formulate and solve the inner optimization problem. First, we use $F_{\boldsymbol{\lambda}}(\mathbf{O})$ as defined in \eqref{argsort}-\eqref{selection} to obtain  the \textit{reduced-set} samples. These are then used to solve optimization \eqref{bi_level_inner_red_cost} along with $^{m}\sigma$ as the kernel parameter to obtain ${^l}\beta^*$. At line 10, we compute the optimal MMD resulting from our choice of ${^l}\boldsymbol{\tau}^'$ samples, ${^l}\beta^*$ and $^{m}\sigma$ and append it to our cost-list. Lines 13-14 refines the sampling distribution of the outer loop. We select the top $n_{elite}$ samples of ${^m}\boldsymbol{\lambda}$ and ${^m}\sigma$ that led to the lowest MMD cost and call it the $EliteSet$. Line 14 estimates the empirical mean and covariance over the  ${^m}\boldsymbol{\lambda}$ and ${^m}\sigma$ contained in the $EliteSet$.
Alg.\ref{algo_1} uses GPU parallelization to achieve computational efficiency. Specifically, the inner QP loop in lines 7-12 can be performed in parallel over different $m$. Moreover, we also leverage the fact that the equality constrained QP \eqref{bi_level_inner_red_cost} have a closed-form solution. Fig.\ref{fig:red_set_analysis} validates the efficacy  of Alg.\ref{algo_1} in solving \eqref{bi_level_red_outer}-\eqref{bi_level_inner_red_cost}: the outer-level cost gradually decreases with iterations. \\
\noindent \textbf{Validating the Efficacy of the optimal \textit{reduced-set}:} To empirically showcase the importance of our optimal \textit{reduced-set}, we sampled $500$ obstacle trajectories to form ${\mathbf{O}}$. We then sub-selected $N^'$ rows from it, either randomly or by solving the bi-level optimization \eqref{bi_level_red_outer}-\eqref{bi_level_inner_red_cost}. We perform a large number of random sub-selections to compute the average, minimum and maximum of $\left\Vert \hat{\mu}[\boldsymbol{\tau}^{'}]- \hat{\mu}[\boldsymbol{\tau}]\right\Vert_{\mathcal{H}}^2 $. Moreover, we compute the same MMD over the optimal \textit{reduced-set} as well. \\
As can be seen from Fig.\ref{fig:red_set_analysis} (b), $\left\Vert \hat{\mu}[\boldsymbol{\tau}^{'}]- \hat{\mu}[\boldsymbol{\tau}]\right\Vert_{\mathcal{H}}^2 $ over the optimal \textit{reduced-set} performs close to the best-case performance that we can obtain by exhaustively trying out different random \textit{reduced-sets}. The performance difference is more stark at lower $N'$. It should be noted that in real-world setting, exhaustive search is prohibitive and thus, \eqref{bi_level_red_outer}-\eqref{bi_level_inner_red_cost} provides a computationally efficient one-shot solution. Fig.\ref{fig:red_set_analysis} (c) shows that the advantages of optimal \textit{reduced-set} translates to empirical embedding $\hat{\mu}[\overline{f^'}]$ as well, with $\left\Vert\hat{\mu}[\overline{f}]- \hat{\mu}[\overline{f^'}]\right\Vert_{\mathcal{H}}^2$  being only slightly inferior to the best-case value obtained with exhaustive sampling of \textit{reduced-set}. Fig.\ref{fig:red_set_analysis} validates our hypothesis that minimizing $\left\Vert \hat{\mu}[\boldsymbol{\tau}^{'}]- \hat{\mu}[\boldsymbol{\tau}]\right\Vert_{\mathcal{H}}^2 $ and $\left\Vert\hat{\mu}[\overline{f}]-\hat{\mu}[\overline{f^'}]\right\Vert_{\mathcal{H}}^2$ ensures minimal loss in the estimation of $r_{MMD}^{emp}$. To this end, we computed

\small
\begin{dmath}
    \Delta_{r_{MMD}} = \left| \left\| \sum_{j=1}^{N} \frac{1}{N} \phi(\overline{f}^j) 
    - \sum_{j=1}^{N} \frac{1}{N} \phi(\delta^j) \right\|_{\mathcal{H}}^2 
    - \left\| \sum_{l=1}^{N'} \beta^l \phi(\overline{f'}^l) 
    - \sum_{l=1}^{N'} \frac{1}{N'} \phi(\delta^l) \right\|_{\mathcal{H}}^2 \right|
    \label{e_mmd}
\end{dmath}
\normalsize

The first term is the $r_{MMD}^{emp}$ computed on the larger set of $N$ samples, while the second term is the same computed on the smaller \textit{reduced-set}. Like before, we constructed the \textit{reduced-set} either randomly or by solving \eqref{bi_level_red_outer}-\eqref{bi_level_inner_red_cost}. As shown in Fig.\ref{fig:red_set_analysis}, the $\Delta_{r_{MMD}}$ obtained with the optimal \textit{reduced-set} is close to the average error that can be achieved by exhaustively trying different choices of the \textit{reduced-set}. 

\begin{figure}[t!]
    \centering
    \includegraphics[width=\columnwidth,height = 0.5\columnwidth]{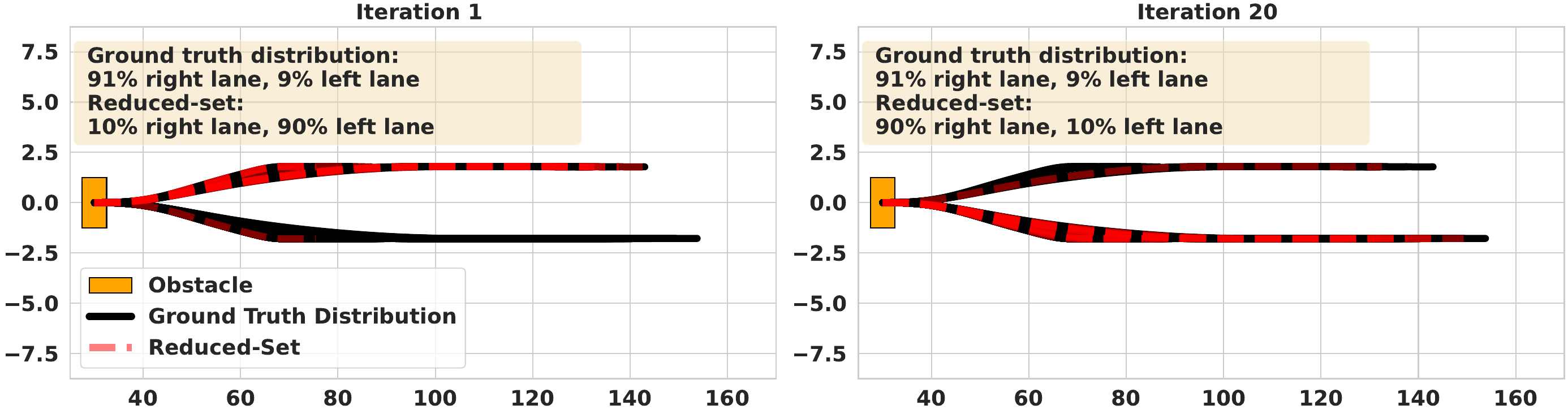}
    \caption{ \footnotesize{The figures demonstrate the physical significance of the trajectory samples used to form the optimal \textit{reduced-set}. We generate a synthetic multi-modal trajectory distribution that captures the different lane-change intents of a dynamic obstacle (orange rectangle). More importantly, due to our specific generation procedure, we know the approximate probability of each lane-change intent. When Alg.\ref{algo_1} starts, the optimal \textit{reduced-set} in the first iteration consists mostly
    of the less likely samples due to the random initialization of the Gaussian distribution from
which $\lambda$ is sampled}. \textbf{For example in the left figure, the \textit{reduced-set} samples are mostly formed by the left lane-change maneuvers which is less likely}. But as Alg.\ref{algo_1} makes progress, the composition significantly changes. For example, in the right figure, the majority of the \textit{reduced-set} samples switch from left-lane to the right-lane maneuvers. Thus, by just minimizing MMD between two sets of trajectory samples, Alg.\ref{algo_1} is able to implicitly infer which trajectory samples are more likely and accordingly split the composition of the optimal \textit{reduced-set} between the less and more likely samples.} 
    \label{fig:red_set_select}
    \vspace{-0.4cm}
\end{figure}


\noindent\textbf{Physical Manifestation of Optimal \textit{Reduced-Set}:} We now present additional empirical analysis which sheds more light into the concept of optimal \textit{reduced-set}. Specifically, we analyze if the obstacle trajectory samples contained in the optimal \textit{reduced-set} have a physical significance. To this end, consider Fig.\ref{fig:red_set_select} which shows a synthetic multi-modal obstacle trajectory distribution that spans two different lane-change intents. This distribution was generated by sampling set-points for lateral offsets (from the lane center-line) and forward velocity from a binomial distribution and a Gaussian Mixture Model respectively and mapping those to trajectory distribution using the planner presented in Appendix \ref{frenet_appendix}. Due to discrete nature of the lateral offset set-point distribution, we can approximately infer which of the lane-change maneuvers of Fig.\ref{fig:red_set_select} are more likely. We now apply Alg.\ref{algo_1} to choose  a subset of trajectory samples to form the optimal \textit{reduced-set}. We visualize the chosen samples in Fig.\ref{fig:red_set_select}. As can be seen, at the first iteration of Alg.\ref{algo_1}, the optimal \textit{reduced-set} is primarily formed by less likely samples from the distribution. But as Alg.\ref{algo_1} progresses and the outer MMD cost \eqref{bi_level_red_outer} goes down, the composition of the optimal \textit{reduced-set} drastically changes. Specifically, more samples of more likely trajectories are found in the optimal \textit{reduced-set} at later iterations. In fact, the split into more and less likely trajectory samples in the optimal \textit{reduced-set} closely follow the ground-truth probability. We observed this trend consistently across all our simulations, and we believe, it points to an interesting emergent behavior in Alg.\ref{algo_1}. To summarize, although, Alg.\ref{algo_1} does not use the actual probability of a trajectory sample in any way, it finds an implicit way of inferring the more likely samples by just minimizing the MMD between two sets of trajectory samples. This in turn, ensures that $r_{MMD}^{emp}$ constructed on the optimal \textit{reduced-set} demonstrates excellent ability to capture collision risk and thus lead to safer motions.
\vspace{-0.3cm}

\noindent 
\begin{algorithm}[!ht]
\caption{Sampling-Based Optimizer to Solve \eqref{cost}-\eqref{ineq_constraints} with $r_{MMD}^{emp}$ as the risk cost}
\small
\label{algo_2}
\SetAlgoLined
$e_{max}$ = Maximum number of iterations, \textit{reduced-set} samples of obstacle trajectories $({^1}\boldsymbol{\tau}^', {^2}\boldsymbol{\tau}^', \dots {^{N^'}}\boldsymbol{\tau}^')$ and kernel parameter $\sigma$ from Alg.\ref{algo_1}\\
Initiate mean $\boldsymbol{\tilde{\nu}}_{e}, \boldsymbol{\tilde{\Sigma}}_{e}$, at $e=0$ for sampling Frenet-Frame behavioural inputs $\mathbf{b}$\\
\For{$e=1, e \leq e_{max}, e++$}
{
Draw ${n}$ samples $(\mathbf{b}_1, \mathbf{b}_2,\dots \mathbf{b}_i,\dots, \mathbf{b}_n)$ from $\mathcal{N}  (\boldsymbol{\tilde{\nu}}_{e}, \boldsymbol{\tilde{\Sigma}}_{e})$ \\
 \vspace{0.1cm}
Initialize $CostList$ = []\\
 \vspace{0.1cm}

Query Frenet-planner for $\forall \mathbf{b}_i$: $(\mathbf{\overline{s}}_i, \mathbf{\overline{d}}_i) = \text{Frenet Planner}(\mathbf{b}_i)$ \\
\vspace{0.1cm}
\text{Project to Constrained Set}
\small
\begin{align*}
    (\mathbf{s}_i, \mathbf{d}_i) = \arg\min_{\mathbf{s}_i, \mathbf{d}_i} \frac{1}{2}\Vert \mathbf{\overline{s}}_i-\mathbf{s}_i\Vert_2^2\nonumber \\+\frac{1}{2}\Vert \mathbf{\overline{d}}_i-\mathbf{d}_i\Vert_2^2\\
     \mathbf{h}(\mathbf{s}_i^{(q)}, \mathbf{d}_i^{(q)}) = \mathbf{0} \\
    \textbf{g}(\mathbf{s}_i^{(q)}, \mathbf{d}_i^{(q)}) \leq 0 
\end{align*}

Define constraint residuals: $\Vert \max(0, \mathbf{g}( \mathbf{s}_i, \mathbf{d}_i)   )\Vert_2^2$\\
\vspace{0.1cm}

$ConstraintEliteSet  \gets$ Select top $n_{c}$   samples of $\mathbf{b}_i, (\mathbf{s}_i, \mathbf{d}_i)$ with lowest constraint residual norm.\\
\vspace{0.1cm}

Define $c_{aug} = c(\mathbf{s}_i, \mathbf{d}_i)+r_{MMD}^{emp}(\mathbf{s}_i, \mathbf{d}_i)+\Vert \max(0, \mathbf{g}( \mathbf{s}_i, \mathbf{d}_i)   )\Vert_2^2$\\
$cost \gets$ $c_{aug}$,  over $ConstraintEliteSet$, $\forall i$ \\
\vspace{0.1cm}

append each computed ${cost}$ to $CostList$ \\
\vspace{0.1cm}
    
$EliteSet  \gets$ Select top $n_{elite}$ samples of \\ ($\mathbf{b}_i , \mathbf{s}_i , \mathbf{d}_i  $)  with lowest cost from $CostList$.\\
 \vspace{0.1cm}
$(\boldsymbol{\tilde{\nu}}_{e+1}, \boldsymbol{\tilde{\Sigma}}_{e+1}) \gets$ Update distribution based on \\ $EliteSet$ 

}

\Return{ Frenet parameter $\mathbf{b}_i $ and  $(\mathbf{s}_i , \mathbf{d}_i)$ corresponding to lowest cost in the $EliteSet$}
\normalsize
\end{algorithm}

\vspace{-0.6cm}
\subsection{Trajectory Optimizer}
\noindent We minimize \eqref{cost} subject to \eqref{eq_constraints}-\eqref{ineq_constraints} with $r_{MMD}^{emp}$ as our risk cost, through sampling-based optimization that combines features from CEM \cite{botev2013cross}, Model Predictive Path Integral (MPPI)\cite{williams2016aggressive} and its modern variants \cite{bhardwaj2022storm}. This class of optimizers operates by drawing at each iteration, a batch of ego-vehicle trajectories from a parametric distribution and evaluating the cost over all of them. Subsequently, the cost values are used to adapt the parameters of the sampling distribution for the next iteration. Typically, these optimizer are used in unconstrained setting. Thus, to account for inequality constraints \eqref{ineq_constraints}, we embed a projection optimizer into the sampling pipeline that is responsible for pushing the sampled trajectories towards feasible region (e.g lane boundary) before evaluating their cost.

Our proposed sampling-based optimizer is presented in Alg.\ref{algo_2} \footnote{We use right subscript to denote ego trajectory samples generated at each iteration of Alg.\ref{algo_2} and left superscript to denote obstacle trajectories sampled from the trajectory predictor.}. The input to our algorithm is the allowable iteration limit and the optimal \textit{reduced-set} samples computed from Alg.\ref{algo_1}. We leverage the problem structure by sampling trajectories in Frenet-Frame. To this end, we first sample ${n}$ behavioural inputs $\textbf{b}_i$ such as desired lateral offsets and longitudinal velocity set-points from a Gaussian distribution (line 4). These are then fed into a Frenet space planner inspired by \cite{wei2014behavioral} in line 6, effectively mapping the distribution over behavioural inputs to that over trajectories. We present the details about the Frenet planner in Appendix \ref{appendix}. The obtained trajectories are then passed in line 7 to a projection optimization that pushes the trajectories towards a feasible region. Our projection optimizer is built on top of \cite{masnavi2022visibility}, \cite{adajania2022multi} and can be easily parallelized over GPUs. In line 8, we evaluate the constraint residuals over the projection outputs $(\mathbf{s}_i, \mathbf{d}_i)$. In line 9, we select top $n_{c}$ $(<n)$ projection outputs with the least constraint residuals. We call this set the $ConstraintEliteSet$. In line 10, we form $c_{aug}$ that consists of the primary cost \eqref{cost} augmented with the constraint residual. In line 11-12, we append $c_{aug}$ to the $CostList$. In line 13, we extract the top $n_{elite}(\leq n_c)$ trajectory samples form $ConstraintEliteSet$ that led to the lowest $c_{aug}$. We call this the $EliteSet$. In line 14, we update the sampling distribution of behavioural inputs based on the cost samples collected in the $EliteSet$. The exact formula for mean and covariance update is presented in \eqref{mean_update}-\eqref{cov_update}. Herein, the constants $\gamma$ and $\eta$ are the temperature and learning rate respectively.
The notion of $EliteSet$ is brought from classic CEM, while the mean and covariance update follow from MPPI \cite{williams2016aggressive}, and \cite{bhardwaj2022storm}. An important feature of Alg.\ref{algo_2} is we have effectively encoded long horizon trajectories with a low dimensional behavioural input vector samples $\mathbf{b}_i$. This, in turn, improves the computational efficiency of our approach.

\small
\begin{subequations}
\begin{align}
    \boldsymbol{\tilde{\nu}}_{e+1} = (1-\eta)\boldsymbol{\tilde{\nu}}_{e}+\eta\frac{\sum_{i=1}^{i=n_{elite}} {\omega_i}\hspace{0.1cm}\mathbf{b}_i   }{\sum_{i=1}^{i=n_{elite}} {\omega_i}}, \label{mean_update}\\
    \boldsymbol{\tilde{\Sigma}}_{e+1} = (1-\eta)\boldsymbol{\tilde{\Sigma}}_{e}+\eta\frac{ \sum_{i=1}^{i=n_{elite}} {\omega_i}(\mathbf{b}_i-\boldsymbol{\tilde{\nu}}_{e+1})(\mathbf{b}_i-\boldsymbol{\tilde{\nu}}_{e+1})^T}   {\sum_{i=1}^{i=n_{elite}} {\omega_i}} \label{cov_update}\\
    {\omega_i} = \exp{\frac{-1}{\gamma}(c_{aug}(\mathbf{s}_i, \mathbf{d}_i )   }) \label{s_formula}
\end{align}
\end{subequations}
\normalsize

\vspace{-0.5cm}
\subsection{Extension to Multiple Obstacles}
\noindent We follow a very basic approach for extending our planning pipeline to multiple obstacles. For each obstacle, we compute individual optimal \textit{reduced-sets} and kernel parameters through Alg.\ref{algo_1}. Since each obstacle computation is decoupled from the others, this step can potentially be parallelized. In Alg.\ref{algo_2}, we compute $r_{MMD}^{emp}$ for each obstacle and add them to estimate the total collision risk.

\vspace{-0.4cm}
\section{Connections to Existing Works}
In this section, we connect our approach to current state-of-the-art (SOTA), while also pointing out how we go beyond them.For the ease of exposition, we have classified the related works into specific categories.

\vspace{-0.3cm}
\subsection{Deterministic Approaches} 
\noindent Motion Planning is one of the most well-studied problems in autonomous driving.  A variety of techniques based on potential-field based methods \cite{rasekhipour2016potential}, spatio-temporal corridor generation \cite{ding2019safe} and trajectory optimization \cite{adajania2022multi} have been proposed. A large section of the existing works including those cited here assume a deterministic environment, where the future trajectories of the obstacles/neighboring vehicles are precisely known. However, such assumptions do not hold in real-world setting as the intents of the other vehicles cannot be known a priori.

\textit{Improvement over SOTA: Our proposed approach provides a stronger safety guarantee in a more realistic setting where the trajectories of the obstacles have a general structure in the form of a probability distribution. }

\vspace{-0.3cm}
\subsection{Chance Constrained Optimization}
\noindent Instead of minimizing risk $r$ (as defined in \eqref{chance_cost}), if we add constraints of the form $r\leq \zeta$ in \eqref{cost}-\eqref{ineq_constraints}, for some constant $\zeta$, we obtain the setting of chance-constrained optimization (CCO). The core effort in CCO lies in computing tractable surrogate of the chance-constraints. To this end, $r_{SAA}\leq \zeta$, based on \eqref{saa_estimate} is one such potential surrogate used in works like \cite{pagnoncelli2009sample}. Similarly, a CVaR estimate defined over the collision constraint function $f$ (recall eqn.\ref{worst_case_distance}) is used in \cite{DBLP:journals/ral/LewBP24} as a substitute for chance-constraints. 


An alternative surrogate for chance-constraints namely the so-called scenario approximations is also often used \cite{calafiore2006scenario},\cite{calafiore2012robust},\cite{9455144},\cite{5399917},\cite{SCHILDBACH20143009}. Here  $r\leq \zeta$  is replaced with deterministic scenario constraints of the form ${^j}f(\mathbf{s},\mathbf{d},{^j}\boldsymbol{\tau})\geq 0$, defined with $j^{th}$ sample of obstacle trajectory prediction. Although conceptually simple, scenario-approximation requires adding a large number of constraints to the optimization problem that can be computationally prohibitive. Moreover, as we accommodate more scenario constraints, the overall solution quality can drastically deteriorate \cite{zhu2020kernel}. 

\noindent \textit{Improvement over SOTA: Our approach can be seen as a relaxation of CCO where we minimize risk instead of constraining it to $\gamma$ level. This alleviates the problem of infeasibility in case $\gamma$ is set too low. Moreover, our approach puts no restriction in the underlying distribution of the uncertainty, which in our case stems from the obstacle trajectory prediction. In Section \ref{det_section}, we also show that our approach outperforms baseline scenario approach for CCO.}
\vspace{-0.3cm}

\subsection{Risk-Aware Planning}
\noindent Our work defines collision risk using the collision constraint residual function $\overline{f}$ (recall \eqref{constraint_viol_exact}). A similar set-up can be found in \cite{yin2022riskawaremodelpredictivepath}, wherein the authors used the empirical CVaR estimate of $\overline{f}$ as a risk estimate. Authors in \cite{mustafa2024racp} define risk in terms of spatio-temporal overlap of ego-vehicle and obstacle footprint and severity of collision. It is simple to observe that $\overline{f}$ actually captures the footprint overlap in space and time. Several other risk metrics like time-to-collision \cite{nguyen2023risk} can also be used as a risk metric. However, estimating it for a given complex distribution of obstacle trajectories could be challenging.



\textit{Improvement over SOTA: One key differentiating feature between our approach and all the above cited works is the focus on sample complexity. In fact, we show in Section \ref{validation} that our collision risk $r_{MMD}^{emp}$ can ensure safety with much fewer number of samples of obstacle trajectories than existing risk metrics based on CVaR and SAA.}

\subsection{RKHS Embedding for Stochastic Trajectory Optimization}
\noindent There have been prior works that have also tried to leverage some concepts from RKHS embedding and MMD for either chance-constrained or risk minimization. The authors in \cite{zhu2020kernel} use an MMD based rejection sampling to make the scenario approximations of chance-constraints less conservative. A recent work \cite{nemmour2022maximum} uses the MMD to define ambiguity sets in distributionally robust chance-constrained problems. 

\textit{Improvement over SOTA: Our work departs from \cite{zhu2020kernel}, \cite{nemmour2022maximum} in a way that it constructs the risk surrogate itself on the top of MMD. This opened the doors to leveraging the power of \textit{reduced-set} in a much more prominent manner than \cite{zhu2020kernel}, \cite{nemmour2022maximum}. Our current work extends \cite{sharma2023hilbert}, \cite{harithas2022cco}, \cite{gopalakrishnan2021solving}, to simultaneously optimize the \textit{reduced-set} selection, weights of the individual sample and the kernel parameters. Especially, in \cite{sharma2023hilbert}, the reduced-set selection and weights optimization were decoupled from each other with the kernel parameter being manually tuned to a fixed value. Our current approach not only leads to a better collision risk estimate but is also superior in terms of the final trajectory quality.}

\begin{figure*}[t!]
    \centering
    \includegraphics[scale=0.38]{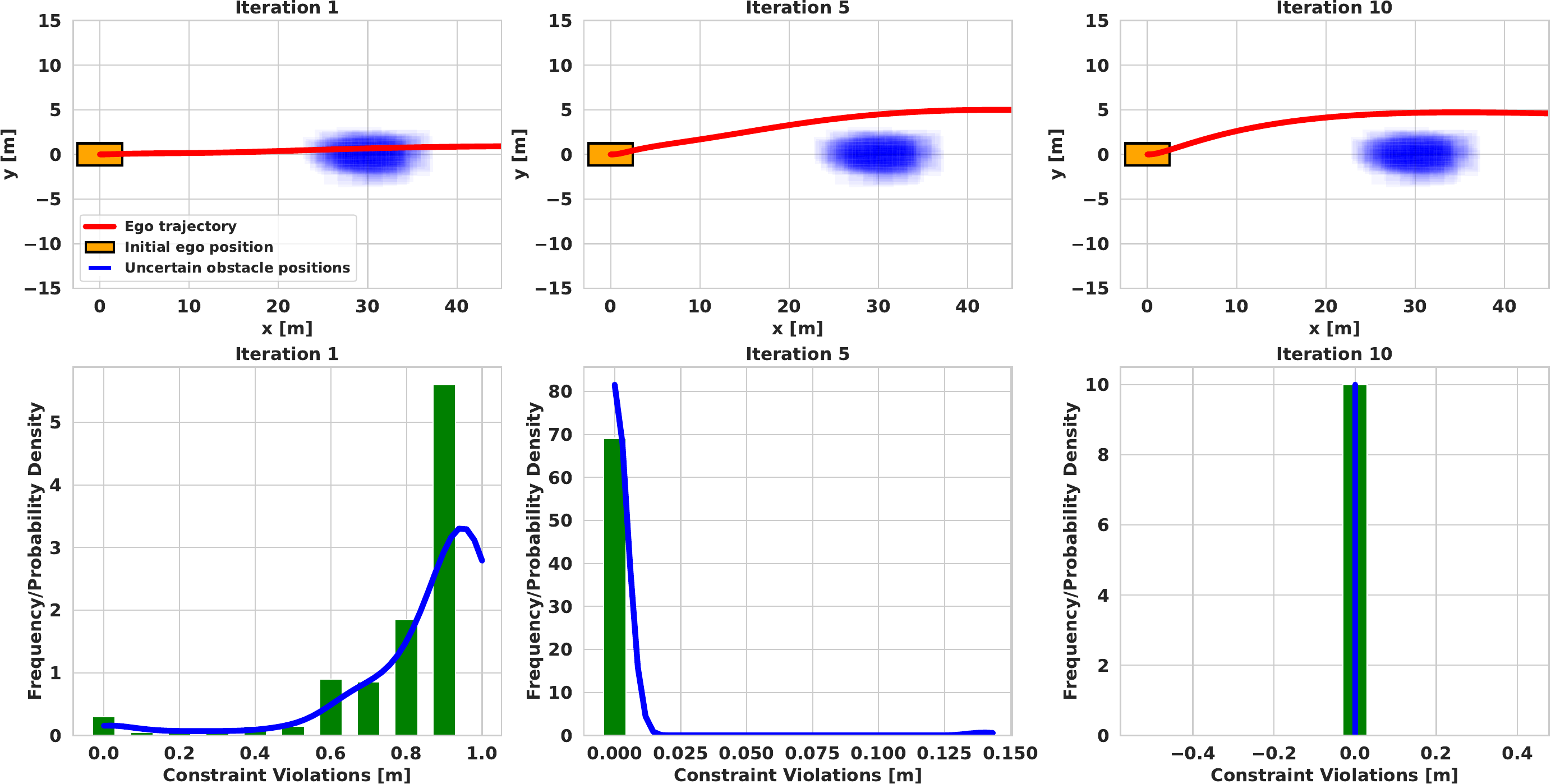}
    \caption{\footnotesize{The top figure shows an ego-vehicle avoiding a uncertain static obstacles. The blue shaded region shows the uncertainty in the location of the obstacle. The trajectory in red depicts the output from Alg.\ref{algo_2} at different iterations. The bottom figure shows the histogram of collision-constraint violation function $\overline{f}$ across iterations. We also fit a KDE to the histogram to approximately represent $p_{\overline{f}}$. As can be seen, $p_{\overline{f}}$ gradually converges to a Dirac-Delta distribution.  }}
    \label{fig:kde_hist_cem_evolve}
    \vspace{-0.5cm}
\end{figure*}
\vspace{-0.2cm}

\section{Validation and Benchmarking}\label{validation}
\noindent This section aims to answer the following research questions

\begin{enumerate}
    \item \textbf{Q1:} How is the sample efficiency of $r_{MMD}^{emp}$ as compared to CVaR and SAA estimate of collision constraint residual distribution? This is covered in Sections \ref{synthetic_static_env}, \ref{synthetic_dynamic_env}, \ref{trajectron}
    \item \textbf{Q2:} What is the improvement in sample efficiency obtained by defining $r_{MMD}^{emp}$ over an optimal \textit{reduced-set} as compared to randomly selected one? Refer to Section \ref{ablation}
    \item \textbf{Q3:} What is the scalability of $r_{MMD}^{emp}$ based risk-aware trajectory optimization with respect to the number of obstacles (Section \ref{computation_time_section}) ?
\end{enumerate}


\begin{figure*}[t!]
     \centering
     \begin{subfigure}{0.325\textwidth}
         \centering
         \caption{\footnotesize{\textbf{Gaussian distribution}}}
         \includegraphics[scale=\scaleboxplotstatic]{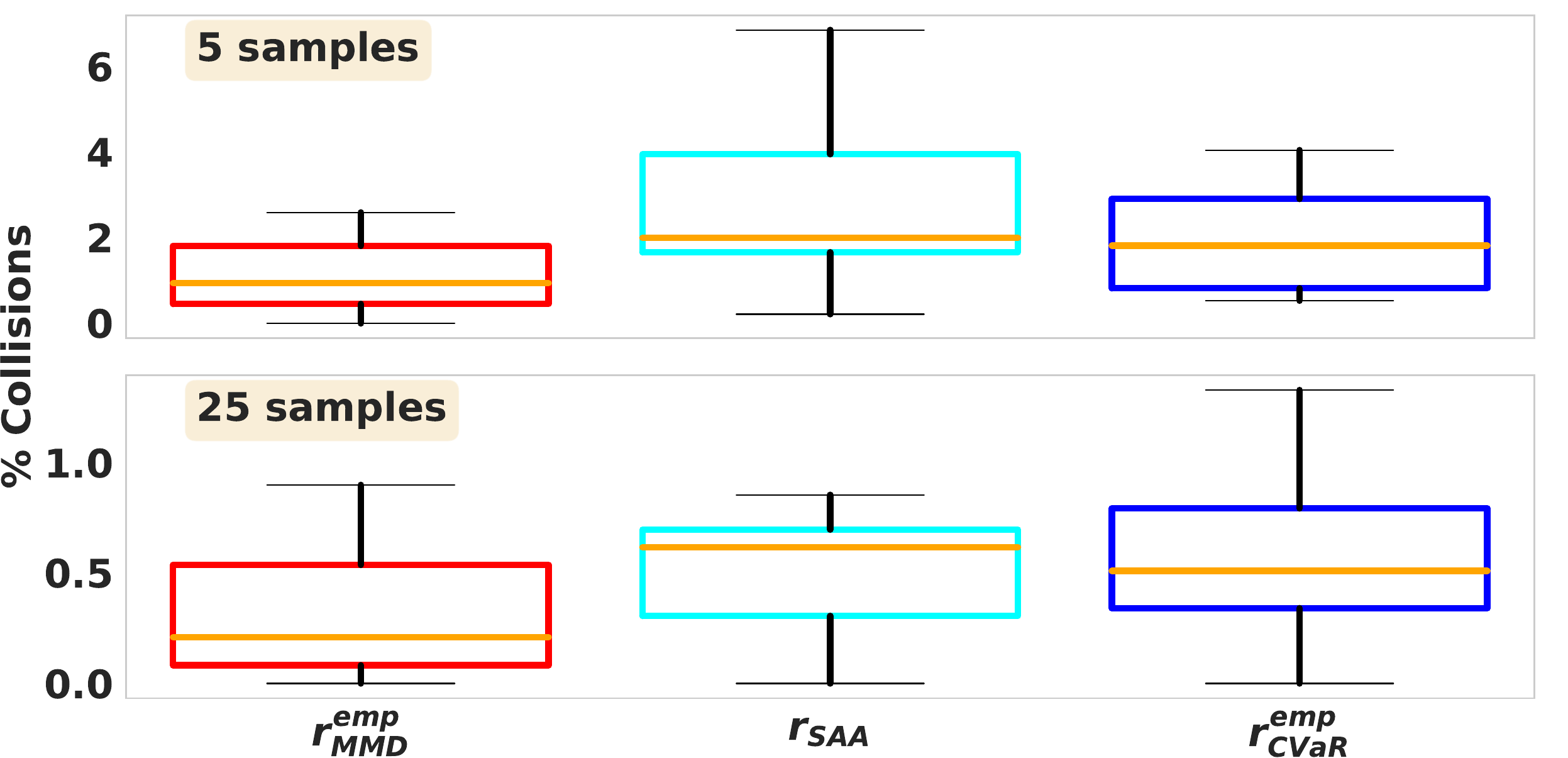}%
         \label{fig:static_plot_gaussian}
     \end{subfigure}
     \begin{subfigure}{0.325\textwidth}
         \centering
         \caption{\footnotesize{\textbf{Bimodal distribution}}}
         \includegraphics[scale=\scaleboxplotstatic]{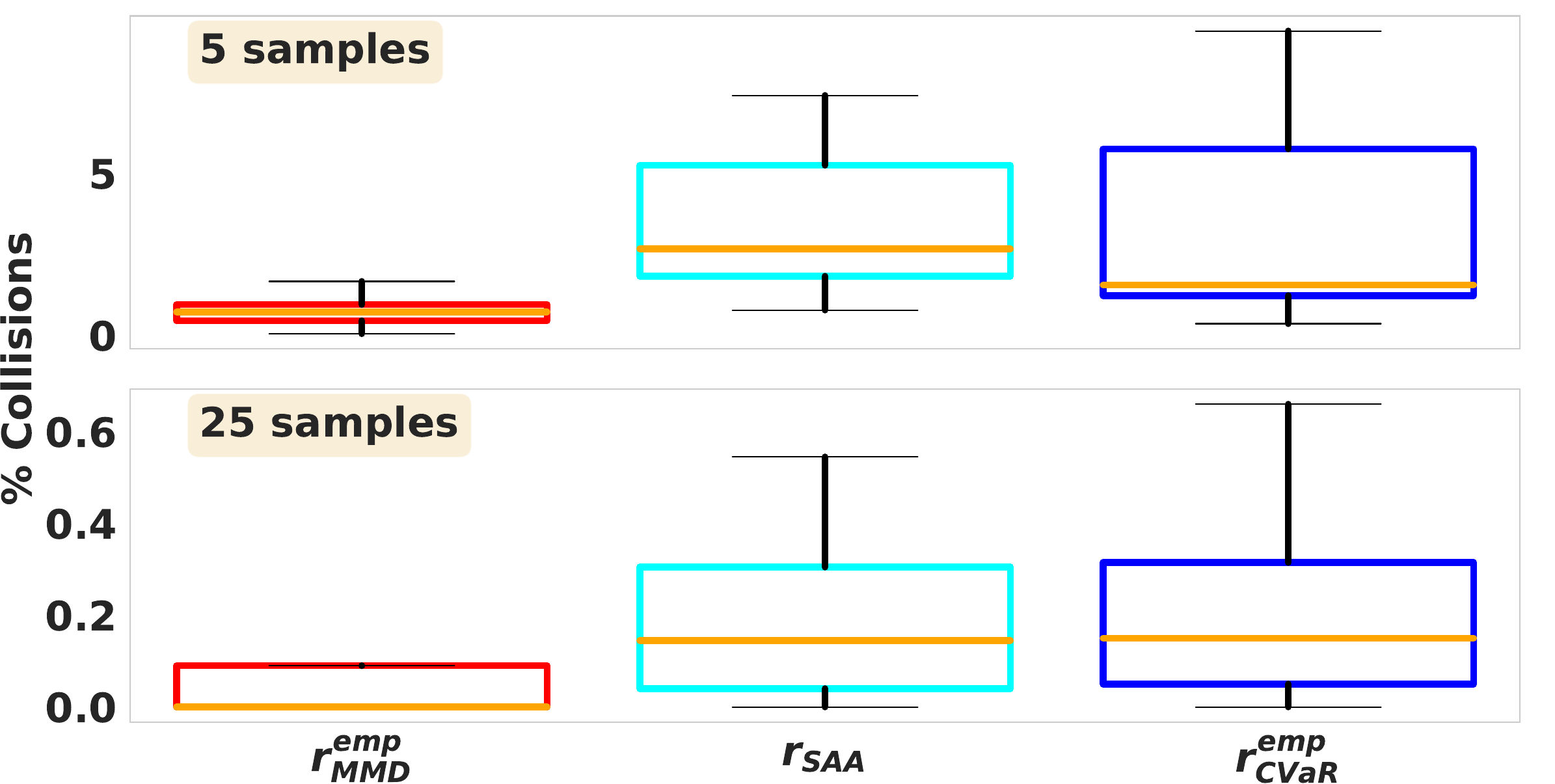}%
         \label{fig:static_plot_bimodal}
     \end{subfigure}
     \begin{subfigure}{0.325\textwidth}
         \centering
         \caption{\footnotesize{\textbf{Trimodal distribution}}}
         \includegraphics[scale=\scaleboxplotstatic]{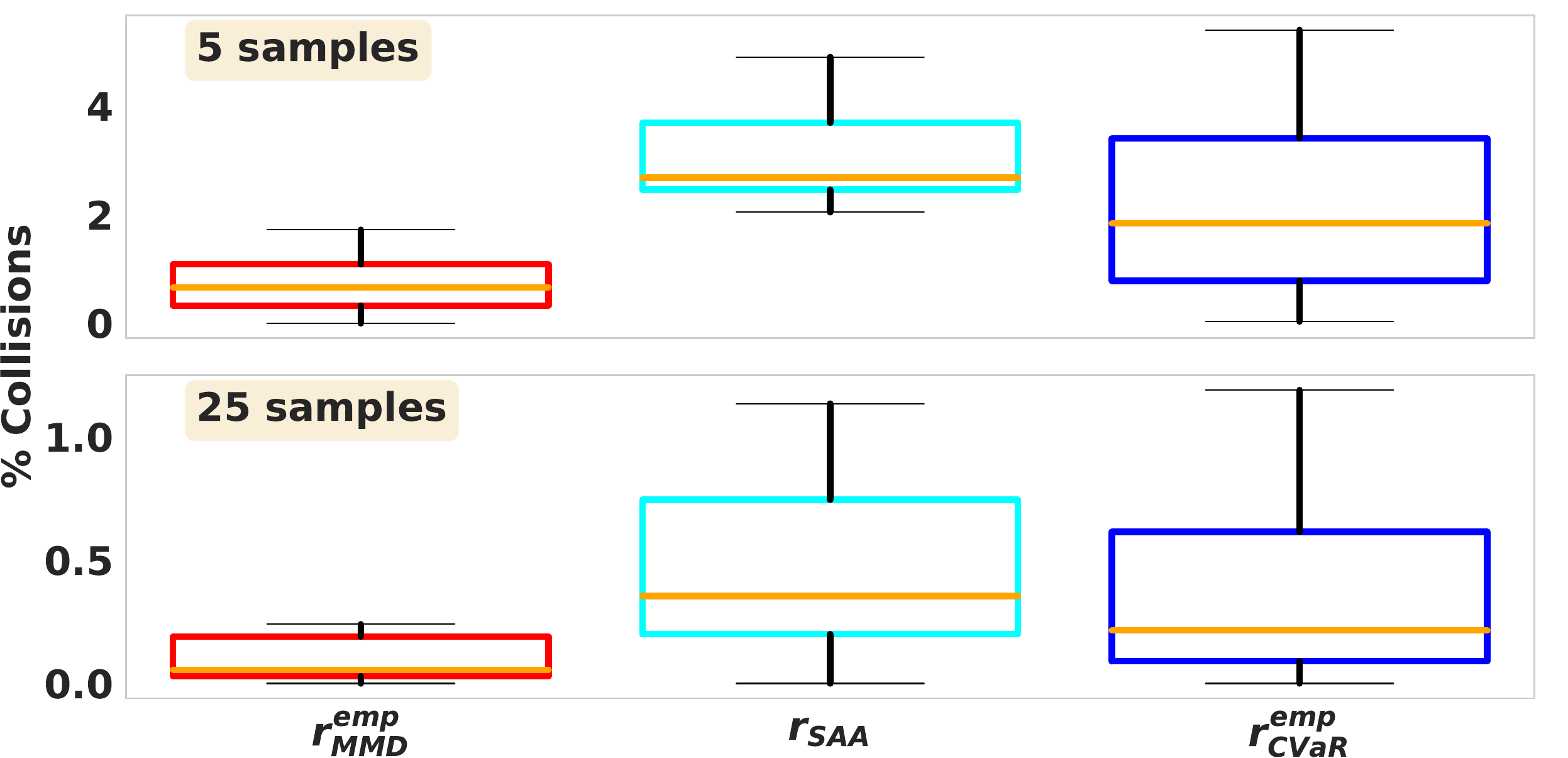}%
         \label{fig:static_plot_trimodal}
     \end{subfigure}
        \caption{\footnotesize{The figure shows the quantitative analysis (in the form of box plots) of collision rates achieved with different risk costs. We experiment with different types of obstacle trajectory distribution and for each type, we perform analysis across different sample sizes. In the case of Gaussian distribution (a), and $N^'=5$, we observe that $r_{MMD}^{emp}$ leads to a worst-case collision-rate that is two times lower than that achieved using  $r_{SAA}$. The performance gap between $r_{MMD}^{emp}$ and the other two baseline is more enhanced when the underlying obstacle trajectory distribution departs significantly from Gaussian. An example of this is show in (b) that is based on bimodal obstacle trajectory distribution. We observe that $r_{MMD}^{emp}$ leads to a worst-case collision-rate that is around five times lower than that achieved using  $r_{CVaR}^{emp}$/$r_{SAA}$. }}
        \label{fig:static_box_plots}
        \vspace{-0.5cm}
\end{figure*}
\vspace{-0.4cm}

\subsection{Implementation Details}
\noindent We implemented our optimal \textit{reduced-set} selection Alg.\ref{algo_1} and Alg. \ref{algo_2} in Python using JAX \cite{jax2018github} library as our GPU-accelerated linear algebra back-end. Our collision constraint function $f_k$ is designed assuming that the ego-vehicle and the obstacle are modeled as axis-aligned ellipses with size $(a_1, a_2)$. Thus, it has the following mathematical form
\begin{align}
    f_k:= -\frac{(s_{k}-s_{o,k})^2}{a_1^2}-\frac{(d_{k}-d_{o,k})^2}{a_2^2}+1
\end{align}

\noindent where $\boldsymbol{\tau}_k=(s_{o,k}, d_{o,k})$ is the position of the obstacle at time-step $k$ and is obtained from the trajectory predictor. Stacking $\boldsymbol{\tau}_k$ at different time steps give us $\boldsymbol{\tau}$.

\noindent The cost term $c$ in \eqref{cost} has the following form
\begin{align}
    c(\mathbf{s}^{(q)}, \mathbf{d}^{(q)}) = c_{\theta}+c_{v}+c_{a}+c_{lane}
    \label{planning_cost}
\end{align}

\noindent with

\begin{dmath}
    c_{\theta} = w_{\theta , 1}\sum_k \max(0, \vert \theta_k\vert-\theta_{max})+w_{\theta , 2}\theta^2_k+w_{\theta , 3}\dot{\theta}^2_k+w_{\theta , 4}\ddot{\theta}^2_k
\end{dmath}

\noindent The variable $\theta_k$ as defined earlier, is the steering angle. Using the motion-model \eqref{diff_flat_1}-\eqref{diff_flat_3}, we can express $\theta_k$ as a function of the the first and second derivatives of the position level trajectory. The different $w_{\theta , i}$ are weights used to trade-off the effect of the cost terms with respect to each other. The derivatives of the $\theta_{k}$ are approximated through finite differencing. 

\noindent The cost, $c_v$, $c_a$, $c_{lane}$ are defined as 
\small
    \begin{align}
        c_v = \sum_k(\dot{s}_k-v_{des})^2, c_a = \sum_k (\ddot{s}^2_k+\ddot{d}^2_k), c_{lane} = \sum_k(d_k-d_{des})^2
    \end{align}
\normalsize
\noindent where $v_{des}$, $y_{des}$ is the desired longitudinal velocity and lane-offset to be tracked by the ego-vehicle.

The equality constraints $\mathbf{h}(\mathbf{s}^{(q)}, \mathbf{d}^{(q)}) = \mathbf{0}$ from \eqref{eq_constraints} have the following components
\begin{subequations}
\begin{align}
    (s_0, \dot{s}_0, \ddot{s}_0) = (0, v_{x, init}, a_{x, init})  \label{boundary_conditions_1}\\
    (d_0, \dot{d}_0, \ddot{d}_0, \dot{d}_H) = (d_{init}, v_{y, init}, a_{y, init}, 0)
    \label{boundary_conditions_2}
\end{align}
\end{subequations}

\noindent where the r.h.s in \eqref{boundary_conditions_1}-\eqref{boundary_conditions_2} consists of the initial and terminal values of position variables and their derivatives along the axes of the Frenet-Frame.

The inequality constraints $\textbf{g}(\mathbf{s}^{(q)}, \mathbf{d}^{(q)}) \leq 0$ have the following components.
\begin{subequations}
    \begin{align}
        d_{min} \leq {d}_k\leq d_{max}, \forall k \label{lane_bound}\\
        \sqrt{\dot{s}_k^2+\dot{d}_k^2}\leq v_{max}, \forall k \label{v_bounds}\\
        \sqrt{\ddot{s}_k^2+\ddot{d}_k^2}\leq a_{max}, \forall k \label{a_bounds}
    \end{align}
\end{subequations}
The first inequality \eqref{lane_bound} is the lane-boundary constraint. The parameters $(d_{min}, d_{max})$ are defined with respect to the lane center-line or a reference path. The inequalities \eqref{v_bounds}-\eqref{a_bounds} bounds the magnitude of the velocity and acceleration respectively.

\subsubsection{Benchmarking Environments} We consider the following collision avoidance scenarios for validating our approach as well as comparing against existing baselines.\\
\noindent \textbf{Static Obstacles:} We consider a two-lane driving scenario with static obstacles. We consider uncertainty in the location of the obstacles and three different choices for its underlying distribution: Gaussian distribution and Gaussian Mixture Model with two and three modes .

\noindent \textbf{Dynamic Environments with Synthetic Trajectory Generator:} In this benchmark, we consider a single dynamic obstacle whose future trajectory has two modes representing the different intents in its motions (e.g lane following or change). Moreover, we also consider uncertainty within the execution of each intent. We generate the trajectory distribution by sampling lane and desired velocity set-points from a discrete and a tri-modal GMM respectively, and passing them through the Frenet Space planner described in Appendix \ref{frenet_appendix}. The scenarios considered cover the typical conflicts observed in autonomous driving.

\noindent \textbf{nuScenes Dataset with Trajectron++ Predictor:} In this benchmark, we consider the driving scenarios contained in the  nuScenes \cite{caesar2020nuscenes} dataset. The scenarios cover a diverse set of conflicts, e.g those arising in un-signaled intersections (see Fig.\ref{fig:teaser}). We use an off-the-shelf pre-trained trajectory predictor Trajectron++ \cite{ivanovic2019trajectron} to generate obstacle trajectory distribution. The goal is to generate a safe trajectory for the ego-vehicle that avoids the obstacle trajectory distribution with high probability. Although originally meant for trajectory forecasting, the NuScenes dataset has also been extensively used to validate safe motion planning, including approaches specifically based on risk-aware trajectory optimization \cite{karkus2023diffstack}, \cite{ren2022chance}.

\subsubsection{Baselines and Benchmarking Protocol}\label{baselines} We compare our approach with two baselines obtained by just replacing our $r_{MMD}^{emp}$ with $r_{SAA}$ and $r_{CVaR}^{emp}$ in the trajectory optimizer of Alg.\ref{algo_2}. 
Our aim is to primarily compare the effectiveness of each risk cost at given sample size $N^'$. Thus, to ensure that the planner biases has no influence on the result, we ensure that optimal trajectories resulting from the baselines and our approach indeed achieve zero risk cost. For example, optimal trajectory resulting from minimizing $r_{CVaR}^{emp}$ would achieve $r_{CVaR}^{emp}=0$ in all benchmarking examples. By adopting such a protocol, we facilitate explicitly comparing the finite-sample estimation error associated with each risk cost.

We split the sampled obstacle trajectories into two sets namely: optimization and validation set. The former set contains the trajectories from which we construct the $r_{MMD}^{emp}$ (including reduced-set computation), $r_{SAA}$ and $r_{CVaR}^{emp}$. The validation set trajectories are never seen by the planner. Moreover, the number of trajectories in validation set are at least two-orders of magnitude larger than the optimization set.

For our approach, we sample $100$ obstacle trajectories and then select $N^'$ out of those as  our reduced-set through Alg.\ref{algo_1}. We perform collision-checks along only  $N^'$ to estimate $r_{MMD}^{emp}$. Since $r_{SAA}$ and $r_{CVaR}^{emp}$ do not have any notion of \textit{reduced-set}, we directly sample $N^'$ obstacle trajectories to estimate their values.

\subsubsection{Metrics}\label{metrics} We primarily focus on the collision percentage on the validation set which is computed as follows. We check the computed optimal ego-trajectory for collision with each obstacle trajectory sample in the validation-set and divide the collision-free cases with the size of the validation set. We perform analysis for different sample size ($N^'$) of the optimization set. Thus, implicitly $N^'$ is also one of our comparison metrics.
\vspace{-0.3cm}

\subsection{Qualitative Analysis of risk-aware planning with $r_{MMD}^{emp}$}
\noindent Figure \ref{fig:kde_hist_cem_evolve} shows the iteration-wise evolution of the optimal trajectory resulting from our optimizer in Alg.\ref{algo_2} while using $r_{MMD}^{emp}$ as the collision risk cost. A static obstacle is placed at an arbitrary position and we introduce uncertainty by adding noise to its ground-truth position.  As the iterations progress, the optimal trajectory shifts away from the potential obstacle locations making the trajectory increasingly safe. In the histogram plot of constraint violations, we observe that with each iteration, the probability of constraint violations goes down and by the $10^{th}$ iteration, the distribution $p_{\overline{f}}$ of the collision constraint violation function $\overline{f}$ has already converged to the Dirac delta distribution $p_{\delta}$.
\vspace{-0.3cm}

\subsection{Benchmarking in Synthetic Static Environments} \label{synthetic_static_env}
\noindent In this benchmark, we randomly generated, 100 different obstacle configurations, each with 3 static obstacles. We introduced noise in the positions of all the obstacles. Three different types of noises were used: Gaussian, Bimodal GMM and Trimodal GMM. The initial position of the ego vehicle was fixed at the same value for all experiments. Similarly, the initial velocity of the ego vehicle was chosen as $3m/s$ in every experiment. We report the collision statistics (Number of collisions of the optimal trajectory with the obstacle samples from the validation set (recall Sections \ref{baselines} and \ref{metrics}).

We present the qualitative results in the complimentary video.  The statistical benchmarking is presented in Fig.\ref{fig:static_box_plots}. The following points are worth noting. First, the $r_{MMD}^{emp}$ trajectories show far lower number of collisions on the validation set than that computed using $r_{CVaR}^{emp}$ and $r_{SAA}$ risk costs at all sample sizes. For example, under Gaussian noise, $r_{MMD}^{emp}$ leads to a worst-case collision-rate that is two times lower than that achieved using  $r_{SAA}$. Second, the difference in performance between $r_{MMD}^{emp}$ and $r_{CVaR}^{emp}$/$r_{SAA}$ becomes more stark as the underlying obstacle trajectories become increasingly more multi-modal. This showcases the ability of $r_{MMD}^{emp}$ to leverage RKHS embedding of distributions to capture the underlying probability distribution of obstacle trajectories and collision constraint residual function $p_{\overline{f}}$ with very few samples. Finally, $r_{MMD}^{emp}$ leads to fastest reduction in collision-rate with increase in sample size. 

\begin{figure*}[t!]
     \centering
     \captionsetup{justification=centering}
     \begin{subfigure}{0.325\textwidth}
         \centering
         \caption{\footnotesize{\textbf{In-lane driving scenario with low probability obstacle cut-in }}}
         \includegraphics[scale=\scaleboxplotdynamic]{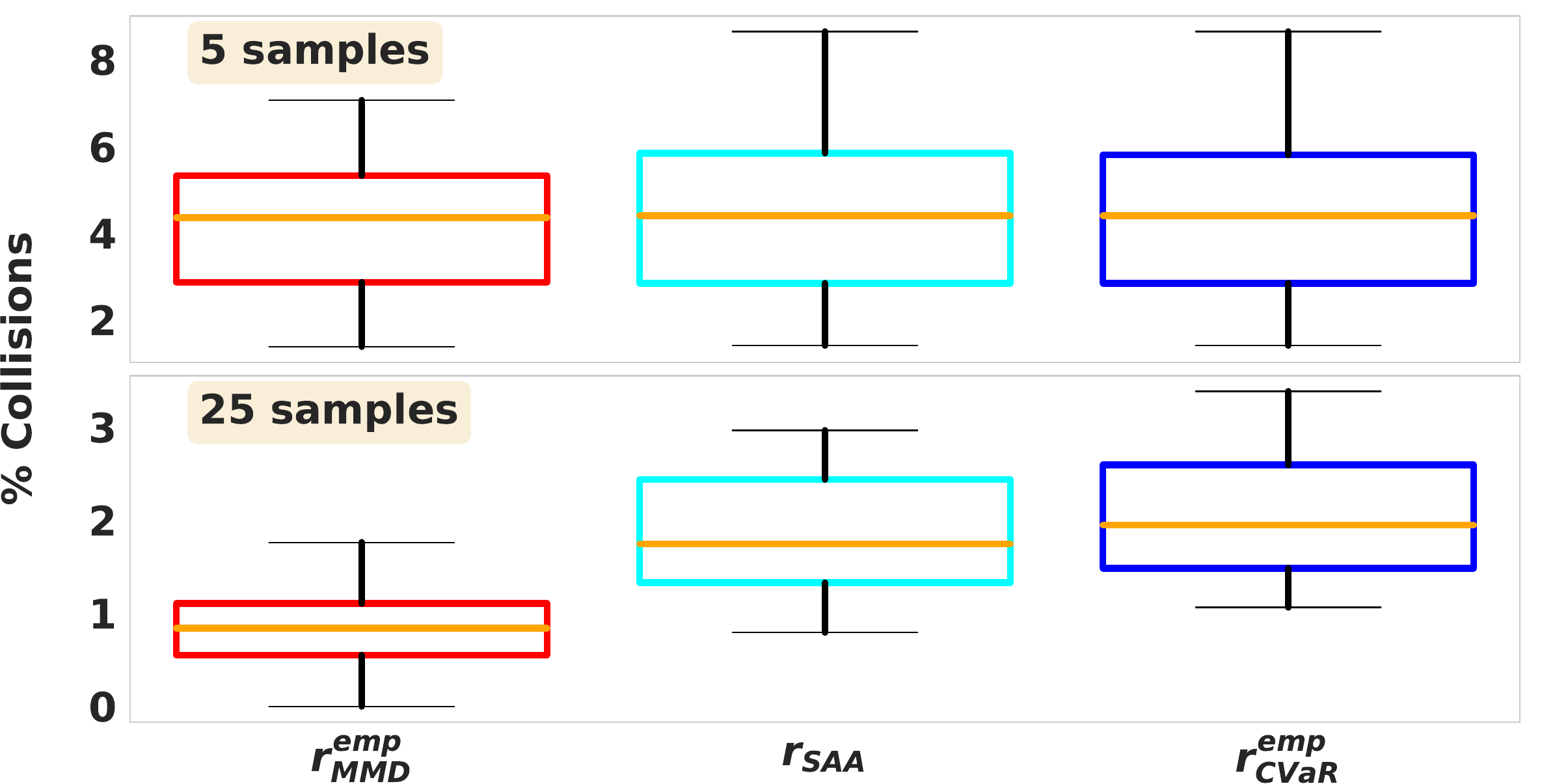}%
         \label{fig:box_plot_dynamic_plot_inlane_cut_in_low}
     \end{subfigure}
     \begin{subfigure}{0.325\textwidth}
         \centering
         \captionsetup{justification=centering}
         \caption{\footnotesize{\textbf{In-lane driving scenario with high probability obstacle cut-in }}}
         \includegraphics[scale=\scaleboxplotdynamic]{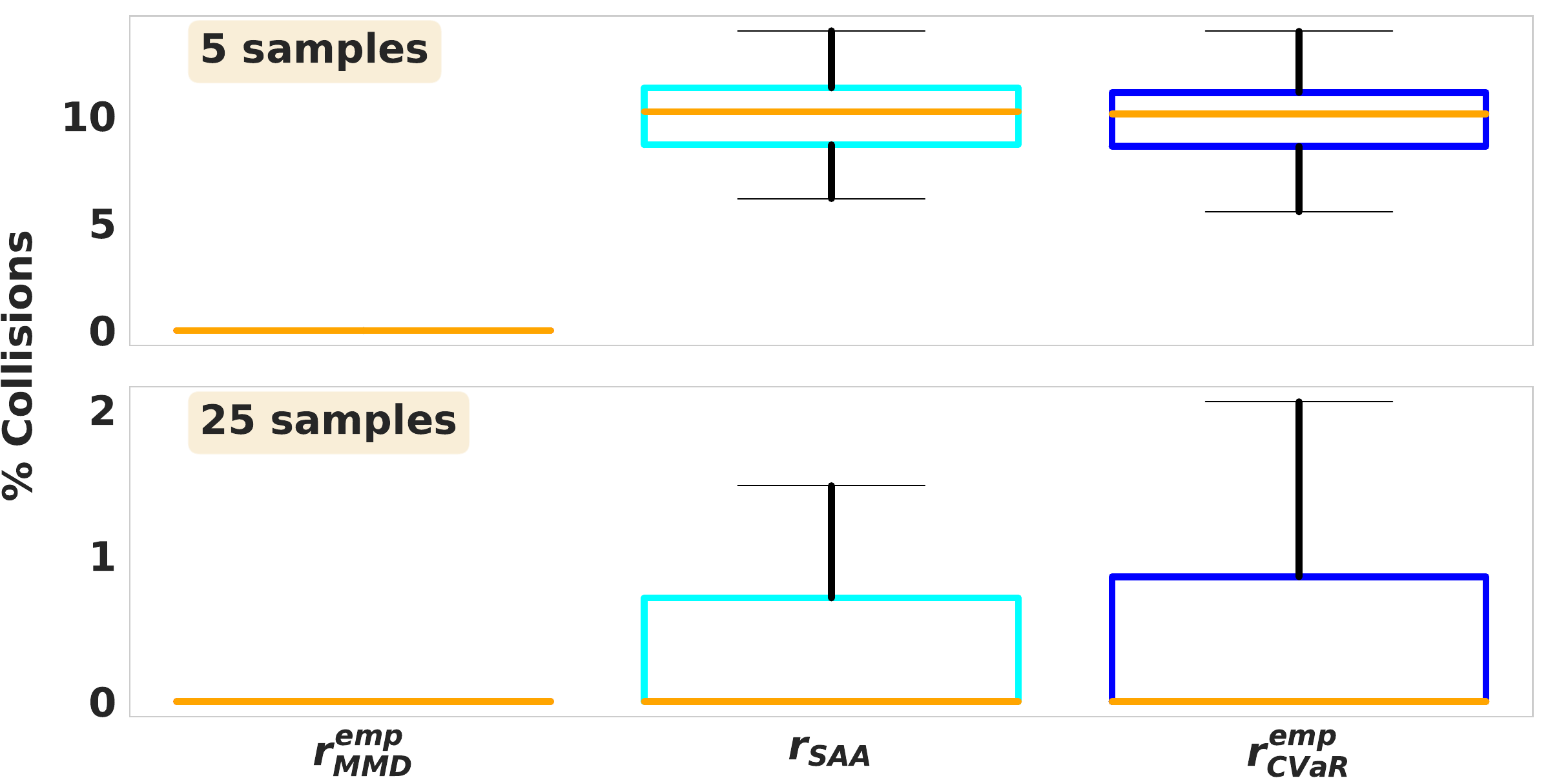}%
         \label{fig:box_plot_dynamic_plot_inlane_cut_in_high}
     \end{subfigure}
     \begin{subfigure}{0.325\textwidth}
         \centering
         \captionsetup{justification=centering}
         \caption{\footnotesize{\textbf{Lane-change driving scenario with high probability obstacle cut-in}}}
         \includegraphics[scale=\scaleboxplotdynamic]{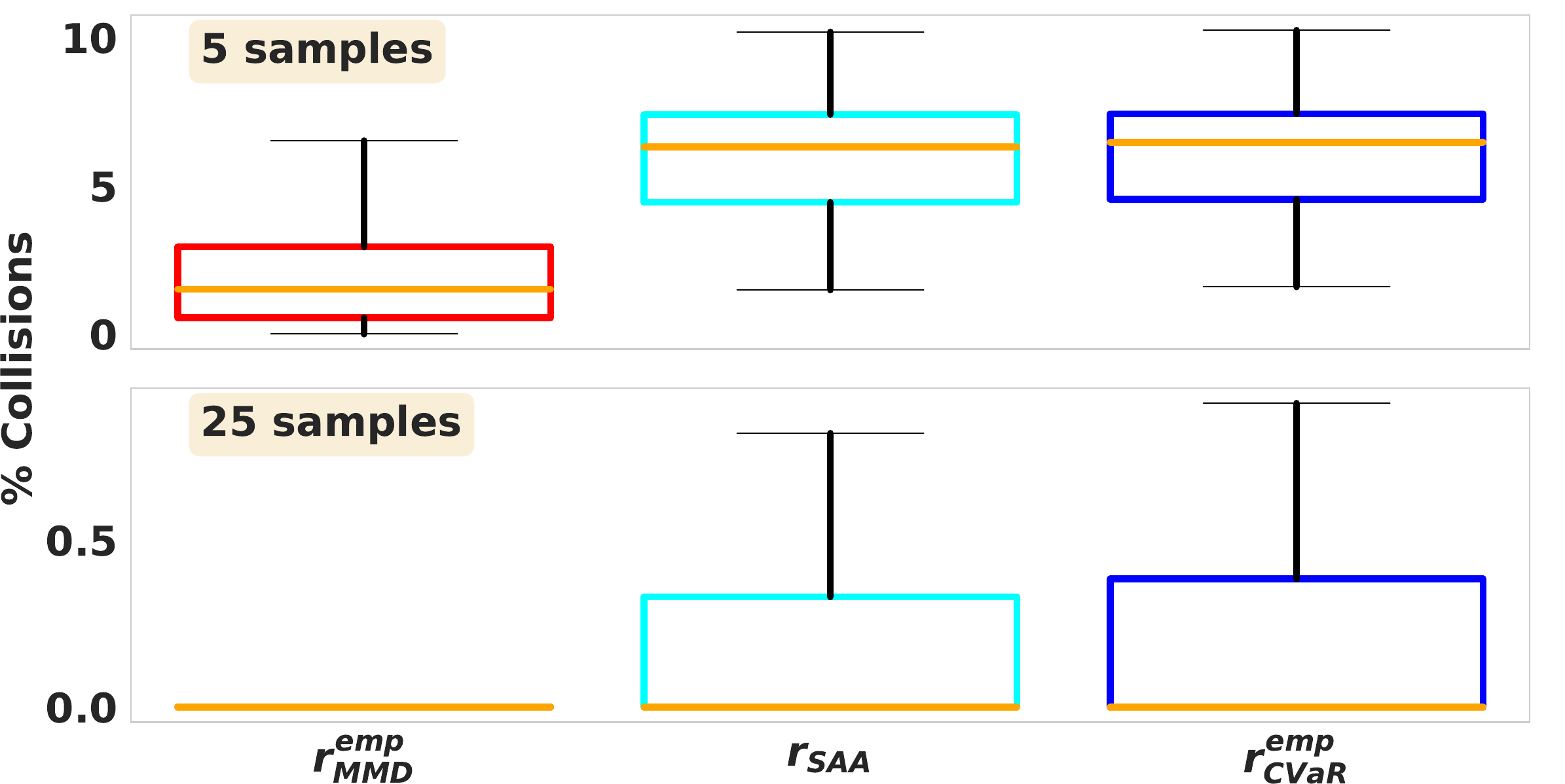}%
         \label{fig:box_plot_dynamic_plot_overtake}
     \end{subfigure}
        \caption{\footnotesize{The box plots show the lower collision-rate achieved by minimizing $r_{MMD}^{emp}$ over the baseline risk cost $r_{SAA}$ and $r_{CVaR}^{emp}$ for the dynamic obstacle avoidance scenario involving multi-modal obstacle trajectories.  }}
        \label{fig:dynamic_box_plots}
\end{figure*}
\vspace{-0.3cm}

\subsection{Benchmarking in Synthetic Dynamic Environments}\label{synthetic_dynamic_env} 
\noindent In this benchmark, we introduced a single dynamic obstacle with uncertain multi-modal future trajectory. The quantitative results are summarized in  Fig.\ref{fig:box_plot_dynamic_plot_inlane_cut_in_low}-\ref{fig:box_plot_dynamic_plot_overtake}. We once again observe that optimal trajectories obtained by minimizing $r_{MMD}^{emp}$ provides the best collision-avoidance performance across all sample-size $N^'$. In particular, while using $N^{'} = 5$ obstacle trajectory samples, the median collision-rate achieved by minimizing $r_{MMD}^{emp}$ is 3 times lower than that resulting from  $r_{CVaR}^{emp}$ and $r_{SAA}$. The performance difference shown in Fig.\ref{fig:box_plot_dynamic_plot_inlane_cut_in_high}, is even more stark with $r_{MMD}^{emp}$ demonstrating zero collisions on the validation set.

\begin{figure}[t!]
\centering
\includegraphics[scale=0.335]{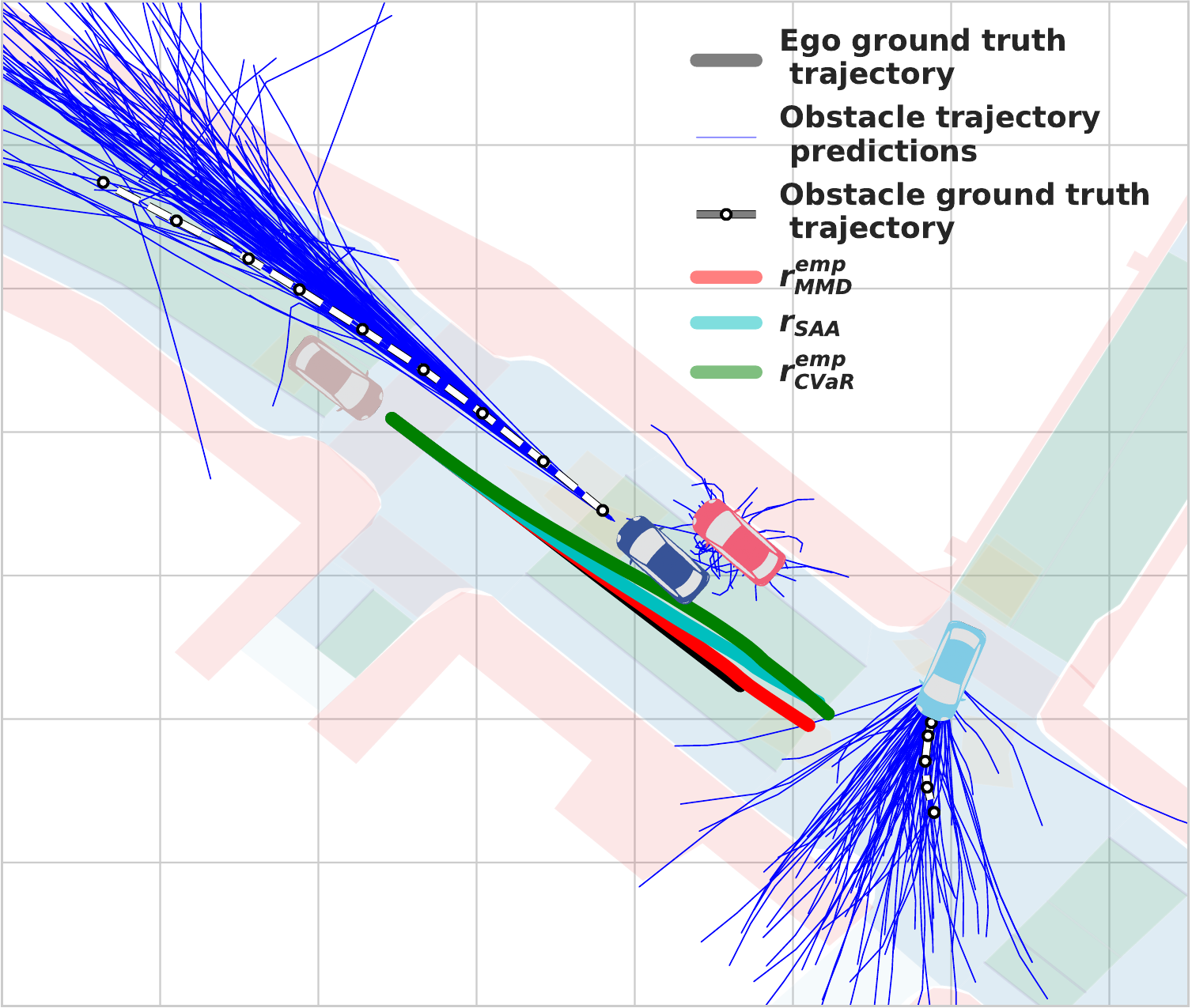}
\caption{\footnotesize{Risk-Aware trajectory planning in an unprotected intersection. The blue trajectories show the trajectory predictions of the neighboring vehicles. As can be seen, the predictions are highly multi-modal capturing different driving intents. Typical ego-vehicle trajectories resulting from different collision risk costs are also shown. It can be seen that $r_{MMD}^{emp}$ correctly understands that the cyan-vehicle is more likely to turn left. Hence, it shows least deviation from the recorded ground-truth (black), human-driven trajectory in the dataset. In contrast, $r_{CVaR}^{emp}$ and $r_{SAA}$ incorrectly puts more emphasis on the less likely scenario of cyan vehicle turning right. As a result, the resulting trajectories show a more deviation from the ground-truth trajectory. }}
\label{fig:teaser}
\vspace{-0.65cm}
\end{figure}

\begin{figure}[t!]
     \centering
     \begin{subfigure}{\columnwidth}
         \centering
         \caption{\footnotesize{\textbf{In-lane scenario}}}
         \includegraphics[scale=0.212]{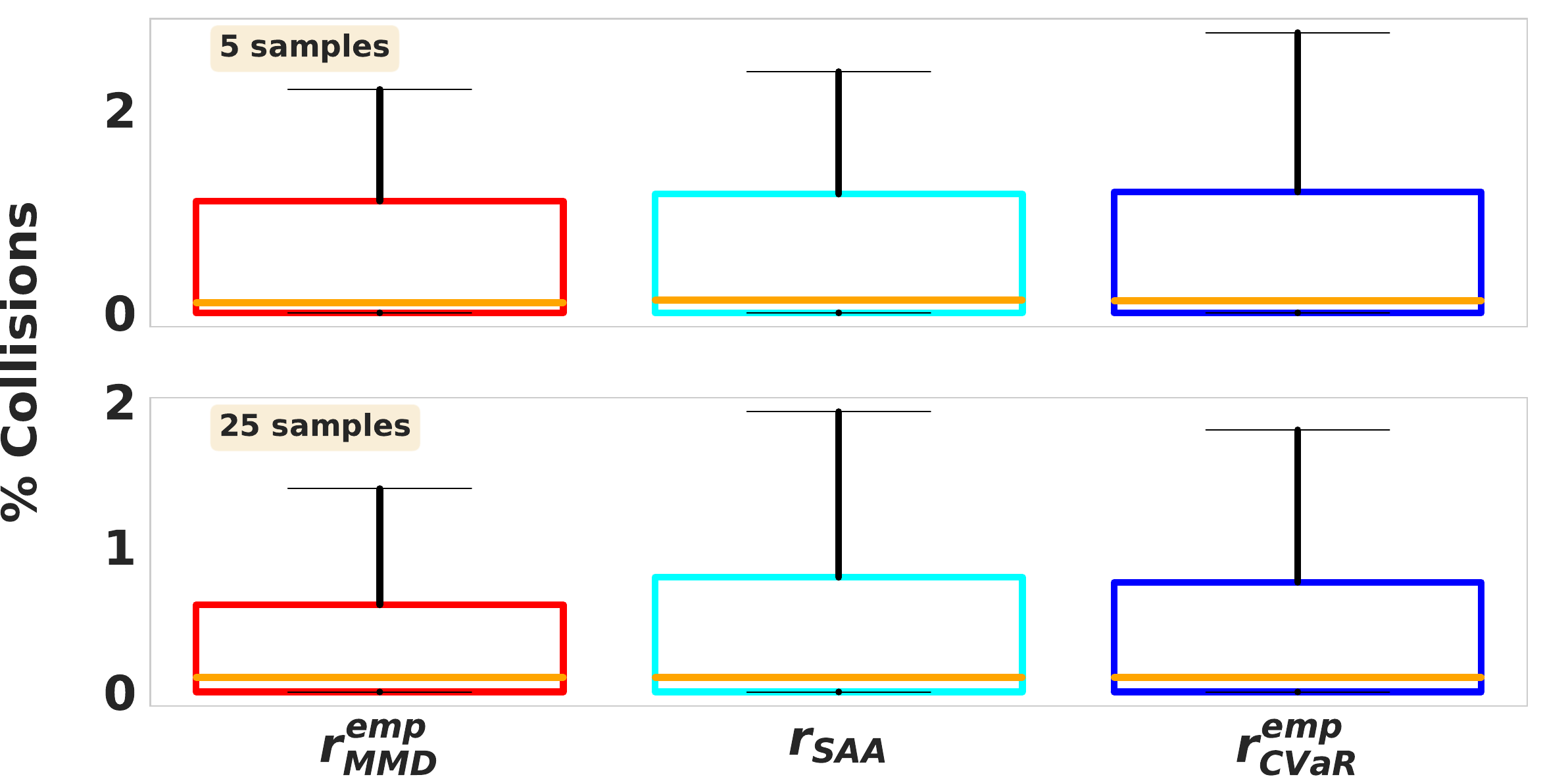}%
         \label{fig:trajectron_inlane}
     \end{subfigure}
     \begin{subfigure}{\columnwidth}
         \centering
         \caption{\footnotesize{\textbf{Lane-change scenario}}}
         \includegraphics[scale=0.215]{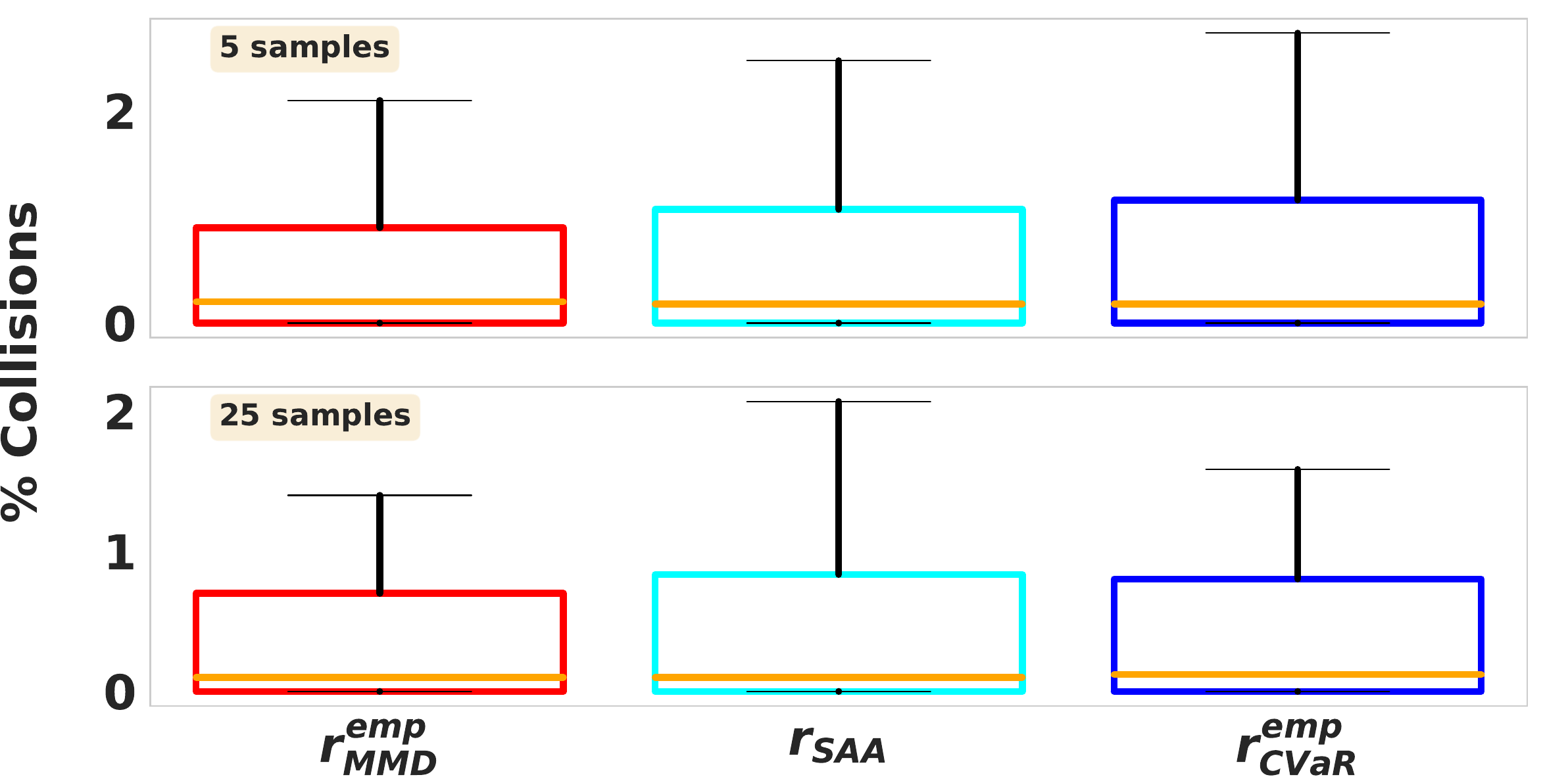}%
         \label{fig:trajectron_overtake}
     \end{subfigure}
        \caption{\footnotesize{Comparing how well different risks $r_{MMD}^{emp}$, $r_{SAA}$, $r_{CVaR}^{emp}$ perform on real-world datasets with Neural Network based trajectory predictors (Trajectron++ \cite{ivanovic2019trajectron} in this case). Typically, these predictors define a very complex distribution over the possible trajectories of the neighboring vehicles (obstacles). Hence, it is challenging to estimate the collision risk with a few predicted samples drawn from the distribution. As can be seen, our surrogate $r_{MMD}^{emp}$ leads to optimal trajectories with lowest collision-rate in both in-lane driving and lane-change scenarios.  }}
        \label{fig:trajectron_box_plots}
        \vspace{-0.5cm}
\end{figure}

\begin{figure*}[t!]
    \centering
    \includegraphics[scale=0.3]{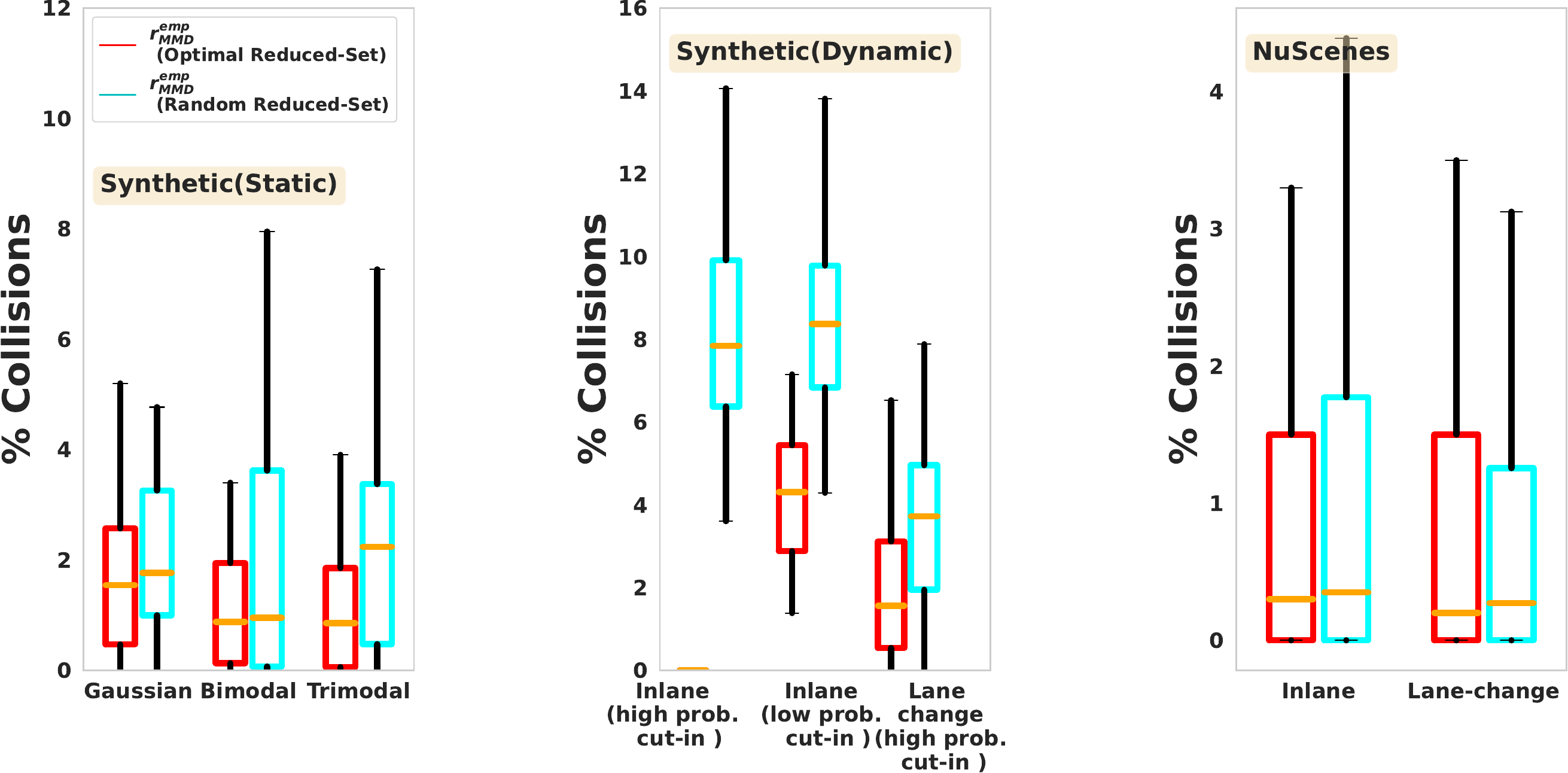}
    \caption{ \footnotesize{Comparison of collision-rate achieved by minimizing $r_{MMD}^{emp}$ constructed with the optimal and random \textit{reduced-set} of size $N^' = 5$. The former set-up leads to safer trajectories with a significantly better median and worst-case statistics. The difference is particularly stark when the obstacle trajectory distribution departs significantly from a Gaussian. } }
    \label{fig:ablation_mmd}
    \vspace{-0.3cm}
\end{figure*}
\vspace{-0.3cm}
\subsection{Benchmarking On nuScenes Dataset Using Trajectron++ Predictor}\label{trajectron}
\noindent We evaluated a total of 160 scenes in this dataset. Each scene comes with a reference center-line and a designated ego vehicle. Any vehicle other than the ego was treated as an obstacle. We considered two settings namely, in-lane driving and lane-change. In the former, the ego-vehicle needs to avoid collisions by just changing the speed, while the latter allows for the full set of maneuvers to the ego-vehicle. We report the collision statistics in Fig. \ref{fig:trajectron_box_plots}. We used Trajectron++ \cite{ivanovic2019trajectron} as a deep neural network based trajectory predictor. We observed that for both scenarios (in-lane and lane-change), our approach based on $r_{MMD}^{emp}$ has the lowest worst-case collision-rate among all the approaches. The improvement is approximately two-times in the lane-change scenario when using $N^' = 25$ samples from obstacle trajectory distribution. The median collision-rate for our approach and the baselines were zero.

A typical example of trajectories obtained in this benchmark is shown in Fig.\ref{fig:teaser} which demonstrates an unprotected intersection. In this setting, the ego-vehicle must safely navigate through cross-traffic while avoiding dynamic obstacles. Unlike standard intersection maneuvers with dedicated traffic signals or stop signs, an unprotected intersection requires the ego-vehicle to make real-time decisions based on surrounding vehicle behavior. The primary challenge in this setting is accurately estimating the risk of collision while ensuring smooth and efficient trajectory execution. As can be seen in Fig.\ref{fig:teaser} the ego-vehicle approaches the intersection while multiple obstacles follow stochastic trajectories across its path. The predicted trajectories cover multi-modal intents (e.g turn left or right). The trajectories resulting from minimizing $r_{MMD}^{emp}$ is closest to the ground-truth (human-driven) recorded trajectory in the same scenario. This in turn can be attributed to the ability to $r_{MMD}^{emp}$ to correctly identify (through its \textit{reduced-set}, recall Fig.\ref{fig:red_set_select}) the more likely maneuvers. For example, consider the cyan obstacle/vehicle. The predicted trajectories cover both left and right turns. But the ground-truth motion for the obstacle is the left turn maneuver and  $r_{MMD}^{emp}$ is able to correctly give more importance to left-turn predictions. In contrast, the $r_{SAA}$, $r_{CVaR}^{emp}$ show overly conservative maneuver driven by the trajectory samples that predict right-turn for the cyan obstacle.   

\vspace{-0.3cm}
\subsection{Ablation Results}\label{ablation}

\noindent Fig.\ref{fig:ablation_mmd} performs an ablation when instead of using Alg.\ref{algo_1}, we form the \textit{reduced-set} by random sub-selection (recall Section \ref{fixed_red_set}). It can be seen that constructing $r_{MMD}^{emp}$ on the optimal \textit{reduced-set} leads to lower collision-rate across all scenes. The difference between the random and the optimal variant is particularly stark when the underlying trajectory distribution is much more complex than a Gaussian. Moreover, the variance in the collision-rate across the benchmarking scenes is lower when using the optimal \textit{reduced-set}.

\vspace{-0.4cm}

\subsection{Comparison with a Scenario Approach}\label{det_section}
\begin{figure*}[ht!]
     \centering
     \captionsetup{justification=centering}
     \begin{subfigure}{0.325\textwidth}
         \centering
         \caption{\footnotesize{\textbf{In-lane driving scenario with low probability obstacle cut-in }}}
         \includegraphics[scale=0.14]{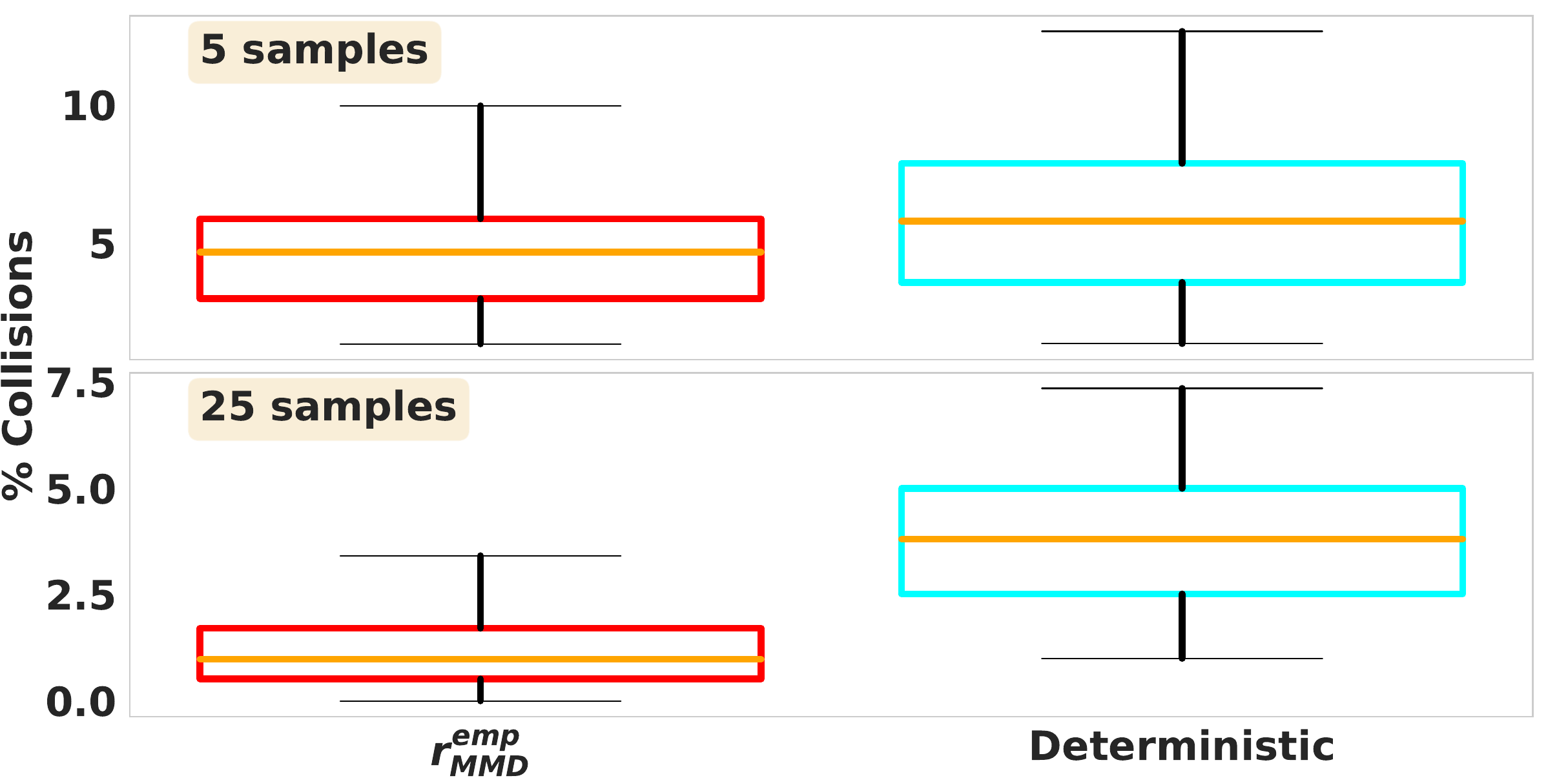}%
         \label{fig:det_inlane_cut_in_low}
     \end{subfigure}
     \begin{subfigure}{0.325\textwidth}
         \centering
         \captionsetup{justification=centering}
         \caption{\footnotesize{\textbf{In-lane driving scenario with high probability obstacle cut-in }}}
         \includegraphics[scale=0.14]{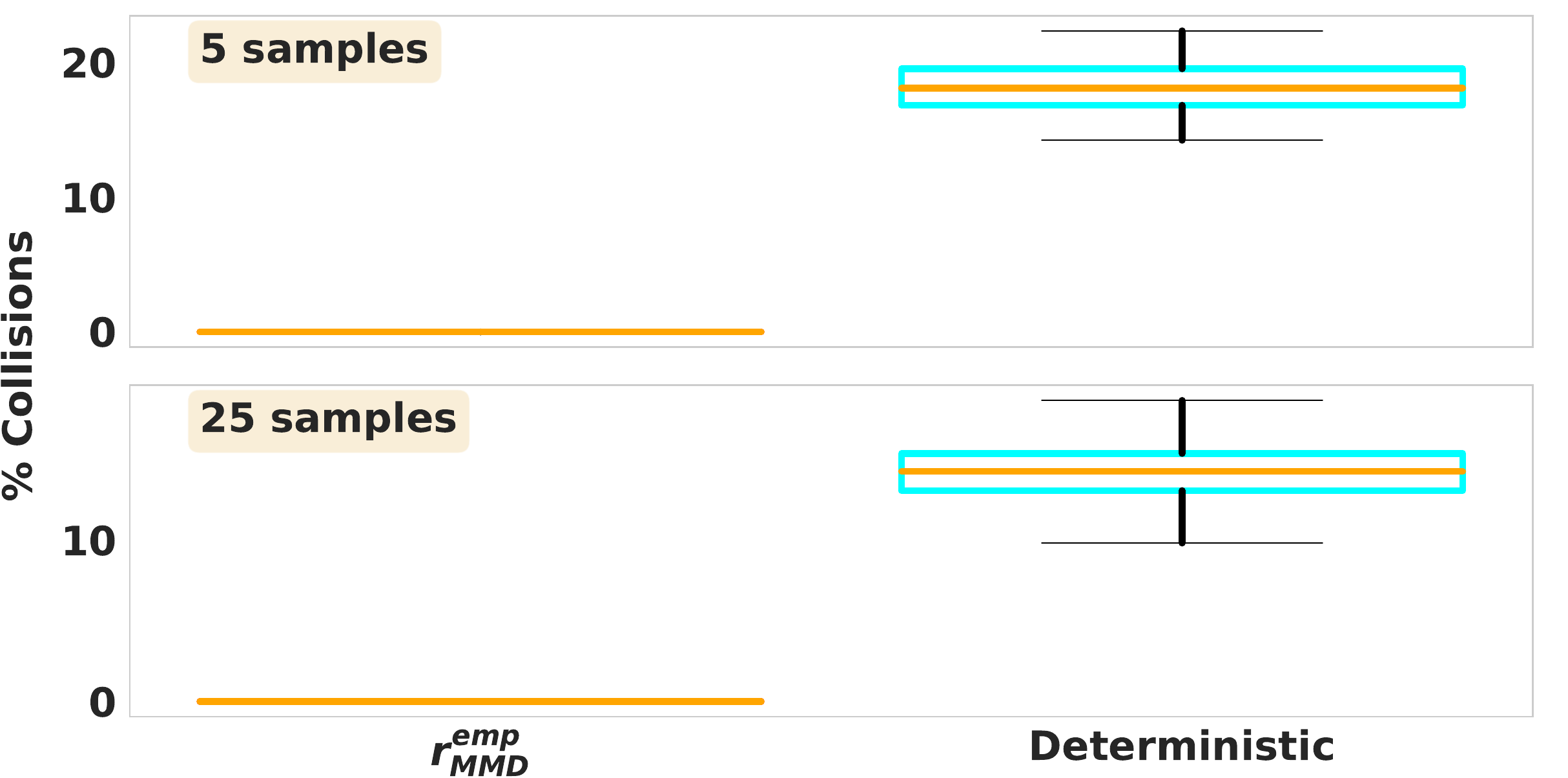}%
         \label{fig:det_inlane_cut_in_high}
     \end{subfigure}
     \begin{subfigure}{0.325\textwidth}
         \centering
         \captionsetup{justification=centering}
         \caption{\footnotesize{\textbf{Lane-change driving scenario with high probability obstacle cut-in}}}
         \includegraphics[scale=0.14]{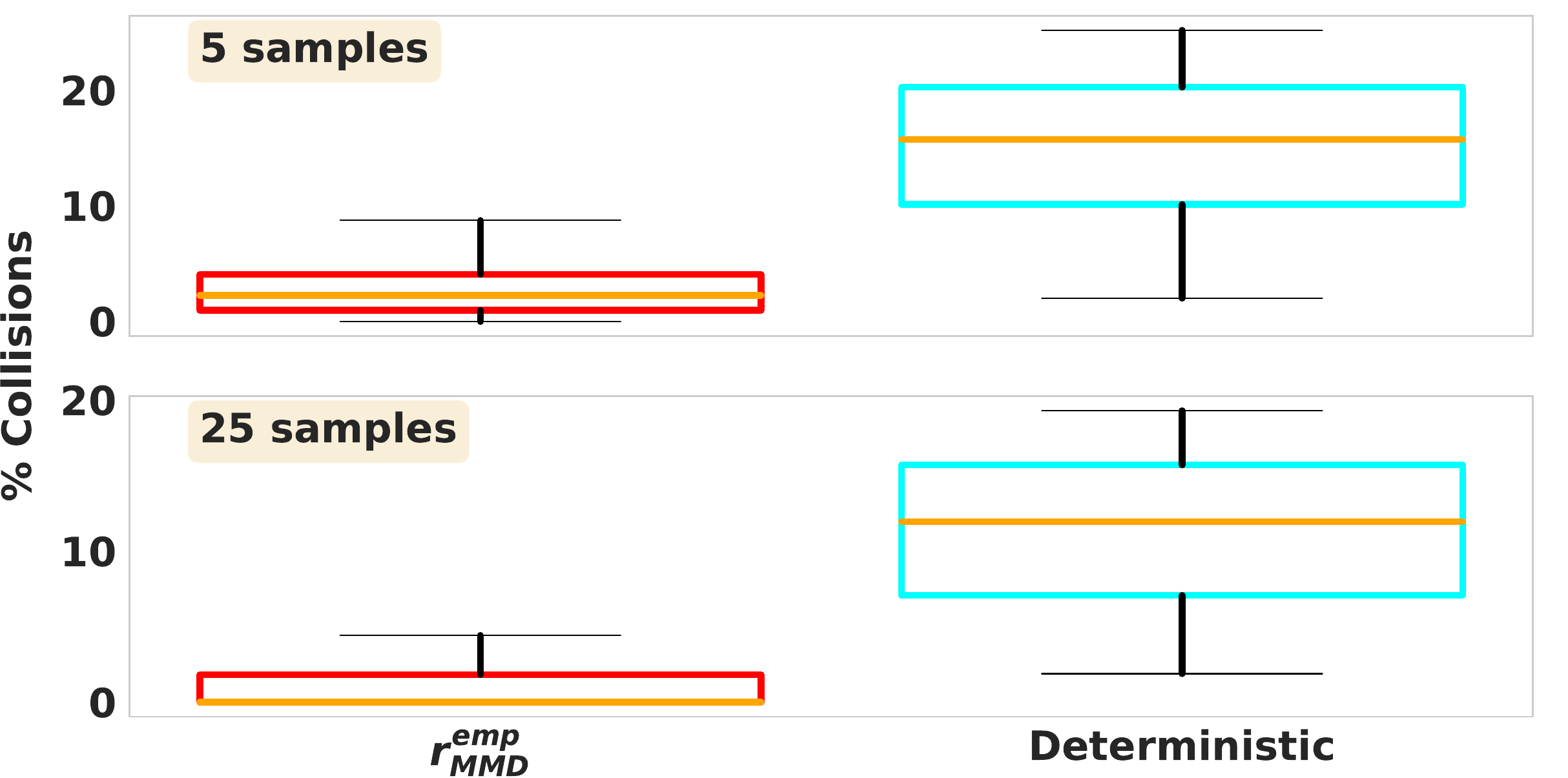}%
         \label{fig:det_overtake}
     \end{subfigure}
        \caption{\footnotesize{The box plots show the lower collision-rate achieved by minimizing $r_{MMD}^{emp}$ over the noise-ignorant deterministic baseline for the synthetic dynamic obstacle avoidance setting, involving multi-modal obstacle trajectories.  }}
        \label{fig:det_box_plots}
    \vspace{-0.3cm}
\end{figure*}

\noindent Fig.~\ref{fig:det_box_plots} presents an ablation study in a dynamic obstacle avoidance setting, comparing our approach to the classical scenario approach for risk-aware planning. This approach boils down to solving the following deterministic problem


\small
\begin{align}
    \min_{\mathbf{s}, \mathbf{d}} c(\mathbf{s}^{(q)}, \mathbf{d}^{(q)}) \label{cost_det} \\
    \mathbf{h}(\mathbf{s}^{(q)}, \mathbf{d}^{(q)}) = \mathbf{0} \label{eq_constraints_det}\\
    \mathbf{g}(\mathbf{s}^{(q)}, \mathbf{d}^{(q)}) \leq 0 \label{ineq_constraints_det}\\
    f_k({s}_{k}, d_{k}, {^j}\boldsymbol{\tau}_{k}) \leq 0,\forall k, j\label{scenario_constraints}
\end{align}
\normalsize

The inequality constraints \eqref{scenario_constraints} enforces a deterministic collision avoidance condition for the $j^{th}$ sampled obstacle motions scenario ${^j}\boldsymbol{\tau}_{k}$ (recall Section \ref{exact_collision_risk}). We solve the scenario approach with CasADi \cite{Andersson2019}. For a fair comparison, we impose as many obstacle avoidance constraints as the \textit{reduced-set} samples used in minimizing $r_{MMD}^{emp}$. The exact mathematical form of the cost and constraints is provided in Section \ref{validation}. Fig.~\ref{fig:det_box_plots} demonstrates that minimizing $r_{MMD}^{emp}$ achieves significantly lower collision rates than the scenario approach, underscoring the effectiveness of our approach in handling multi-modal uncertainty, improving collision avoidance, and enhancing planning robustness.

\vspace{-0.5cm}
\subsection{Impact of Using an Optimal Reduced-Set for Baselines}\label{opt_red_section}
\begin{figure}[t!]
    \centering
    \captionsetup{justification=centering}
    
    \begin{subfigure}[t]{0.48\columnwidth}
        \centering
        \includegraphics[width=\linewidth]{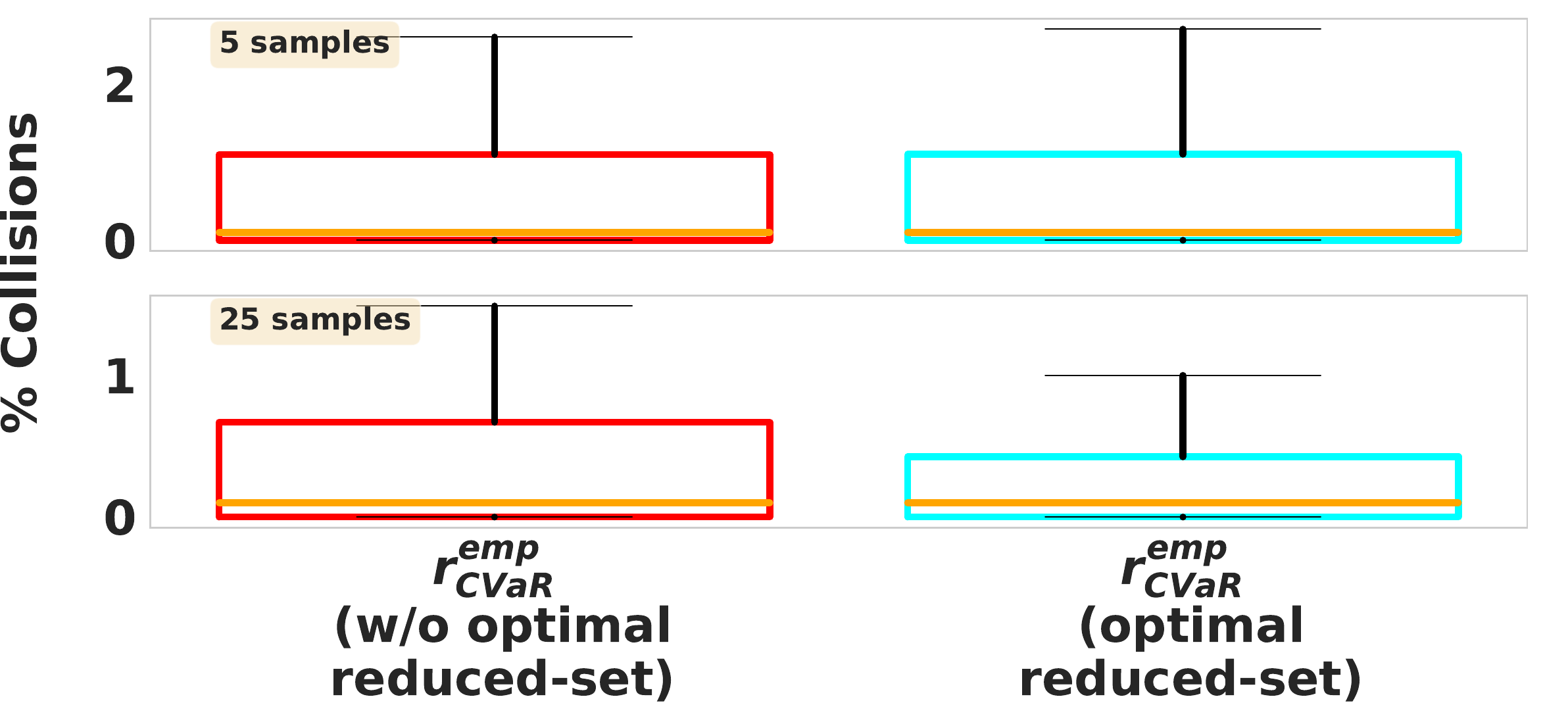}
        \caption{\footnotesize{\textbf{In-lane scenario}}}
        \label{fig:cvar_inlane_opt_red_set}
    \end{subfigure}
    \hfill
    \begin{subfigure}[t]{0.48\columnwidth}
        \centering
        \includegraphics[width=\linewidth]{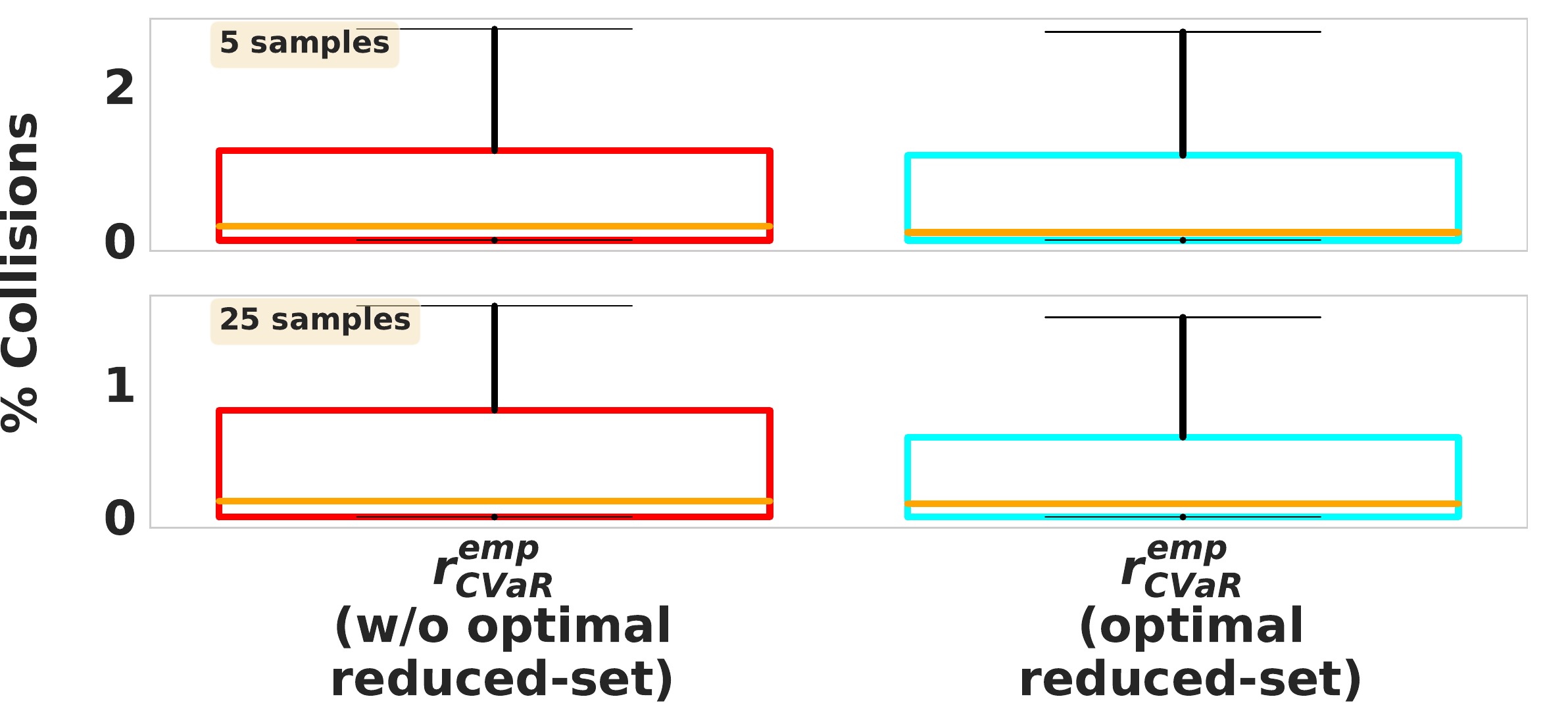}
        \caption{\footnotesize{\textbf{Lane change scenario}}}
        \label{fig:cvar_overtake_opt_red_set}
    \end{subfigure}

    \vspace{0.2cm}

    \begin{subfigure}[t]{0.48\columnwidth}
        \centering
        \includegraphics[width=\linewidth]{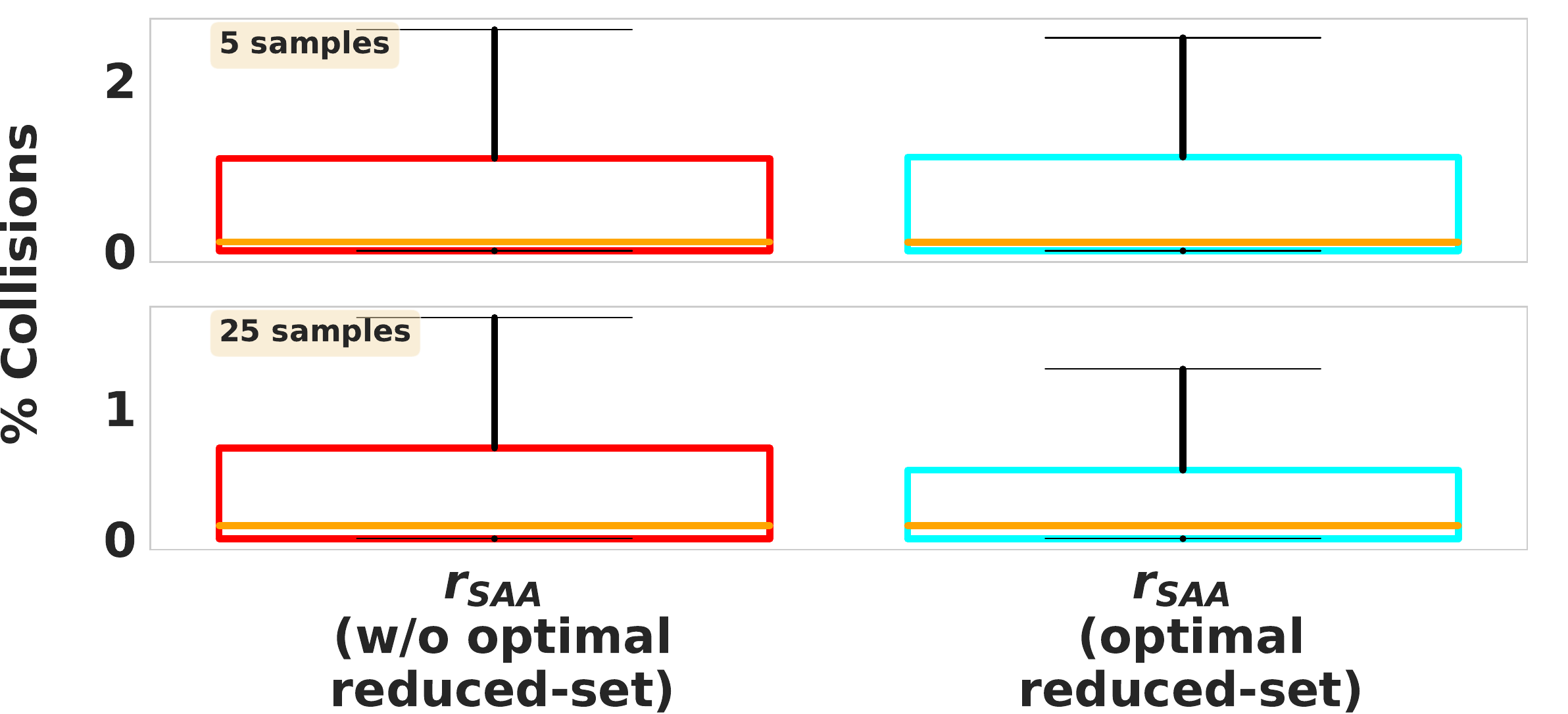}
        \caption{\footnotesize{\textbf{In-lane scenario}}}
        \label{fig:saa_inlane_opt_red_set}
    \end{subfigure}
    \hfill
    \begin{subfigure}[t]{0.48\columnwidth}
        \centering
        \includegraphics[width=\linewidth]{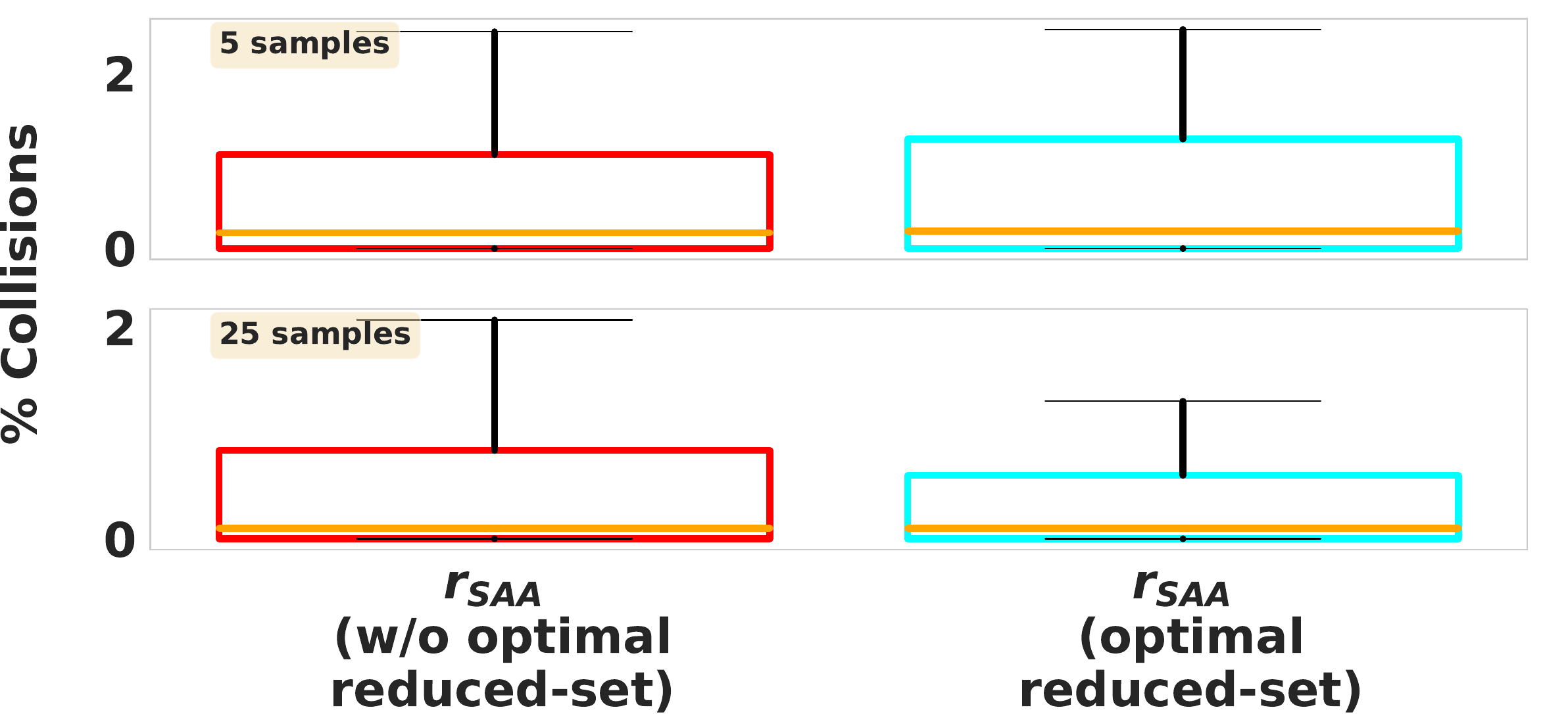}
        \caption{\footnotesize{\textbf{Lane change scenario}}}
        \label{fig:saa_overtake_opt_red_set}
    \end{subfigure}

    \caption{\footnotesize{Ablation study on the nuScenes dataset: Comparison of performance of $r_{CVaR}^{emp}$ and $r_{SAA}$ with and without the use of optimal reduced-set of obstacle trajectories. The results show that the collision-avoidance performance of the baselines $r_{CVaR}^{emp}$ and $r_{SAA}$ remains largely unaffected with the introduction of the optimal reduced-set, reinforcing the theoretical distinction between MMD-based risk estimation and conventional risk metrics.}}
    \label{fig:baseline_opt_red_set_box_plots}
    \vspace{-0.3cm}
\end{figure}

\noindent The baselines $r_{CVaR}^{emp}$ and $r_{SAA}$ do not incorporate the concept of a reduced-set, which is mathematically well-founded only for $r_{MMD}^{emp}$. This distinction arises due to the kernel-based distributional properties of MMD, as discussed in \cite{Muandet_2017}. To the best of our knowledge, no equivalent reduced-set formulation exists for $r_{CVaR}^{emp}$ or $r_{SAA}$. However, for the purpose of fairness in comparison, we conducted an additional ablation study on the nuScenes dataset. In this experiment, we applied the optimal reduced-set of obstacle trajectories—originally derived for $r_{MMD}^{emp}$—to the computation of $r_{CVaR}^{emp}$ and $r_{SAA}$. The results, presented in Figure \ref{fig:baseline_opt_red_set_box_plots}, confirm that the introduction of the reduced-set does not provide a significant advantage to these baselines. This outcome is expected, as $r_{CVaR}^{emp}$ and $r_{SAA}$ rely on independent random sampling from the underlying obstacle trajectory distribution, rather than optimizing over a subset of the samples in a manner mathematically consistent with $r_{MMD}^{emp}$. Consequently, while $r_{MMD}^{emp}$ is bound to benefit from the reduced-set formulation, the baselines remain largely unaffected by this modification. This experiment further highlights the theoretical distinction between MMD-based risk estimation and conventional risk metrics such as CVaR and SAA.

\vspace{-0.27cm}

\subsection{Risk-Aware Planning with Dynamics Model}
\noindent Our proposed risk surrogate $r_{MMD}^{emp}$ just requires the information about the ego-vehicle and the obstacle motions. Thus, it is agnostic to the actual motion model of the ego-vehicle. The results shown in the previous subsection assumed kinematic bi-cycle model. However, we can get similar results even if we use a more complex dynamics for the ego-vehicle that includes tire forces and side-slips. To further validate this assertion, we adapted our Alg. \ref{algo_2} to work with the dynamics model derived from \cite{Becker_2023},\cite{9341731}. For comparison, we also minimized $r_{CVaR}^{emp}$ and $r_{SAA}$ under the same dynamics. 

\begin{figure}[t!]
     \centering
     \begin{subfigure}{\columnwidth}
         \centering
         \caption{\footnotesize{\textbf{In-lane scenario}}}
         \includegraphics[scale=0.212]{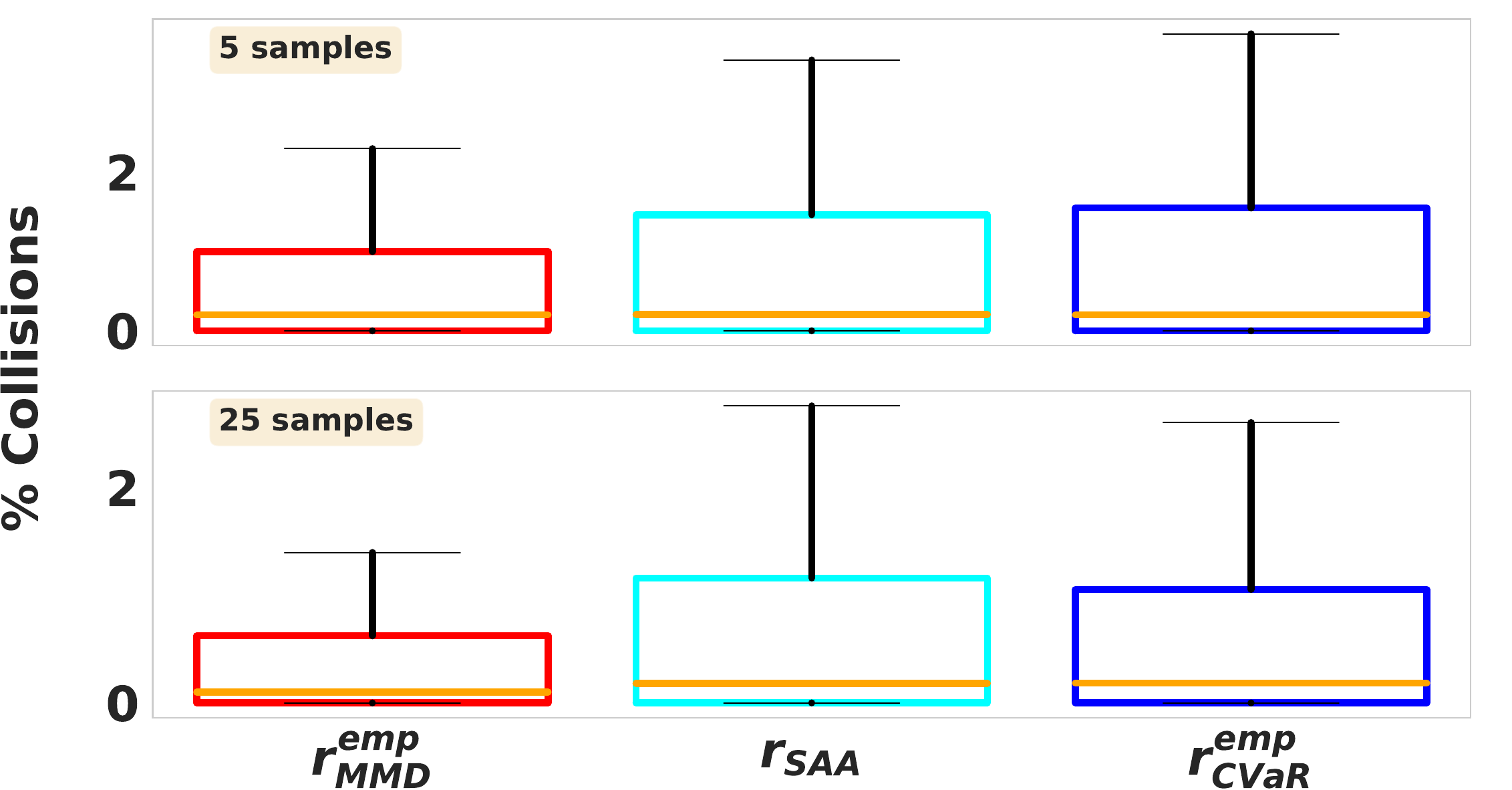}%
         \label{fig:trajectron_inlane_dynamics}
     \end{subfigure}
     \begin{subfigure}{\columnwidth}
         \centering
         \caption{\footnotesize{\textbf{Lane-change scenario}}}
         \includegraphics[scale=0.2]{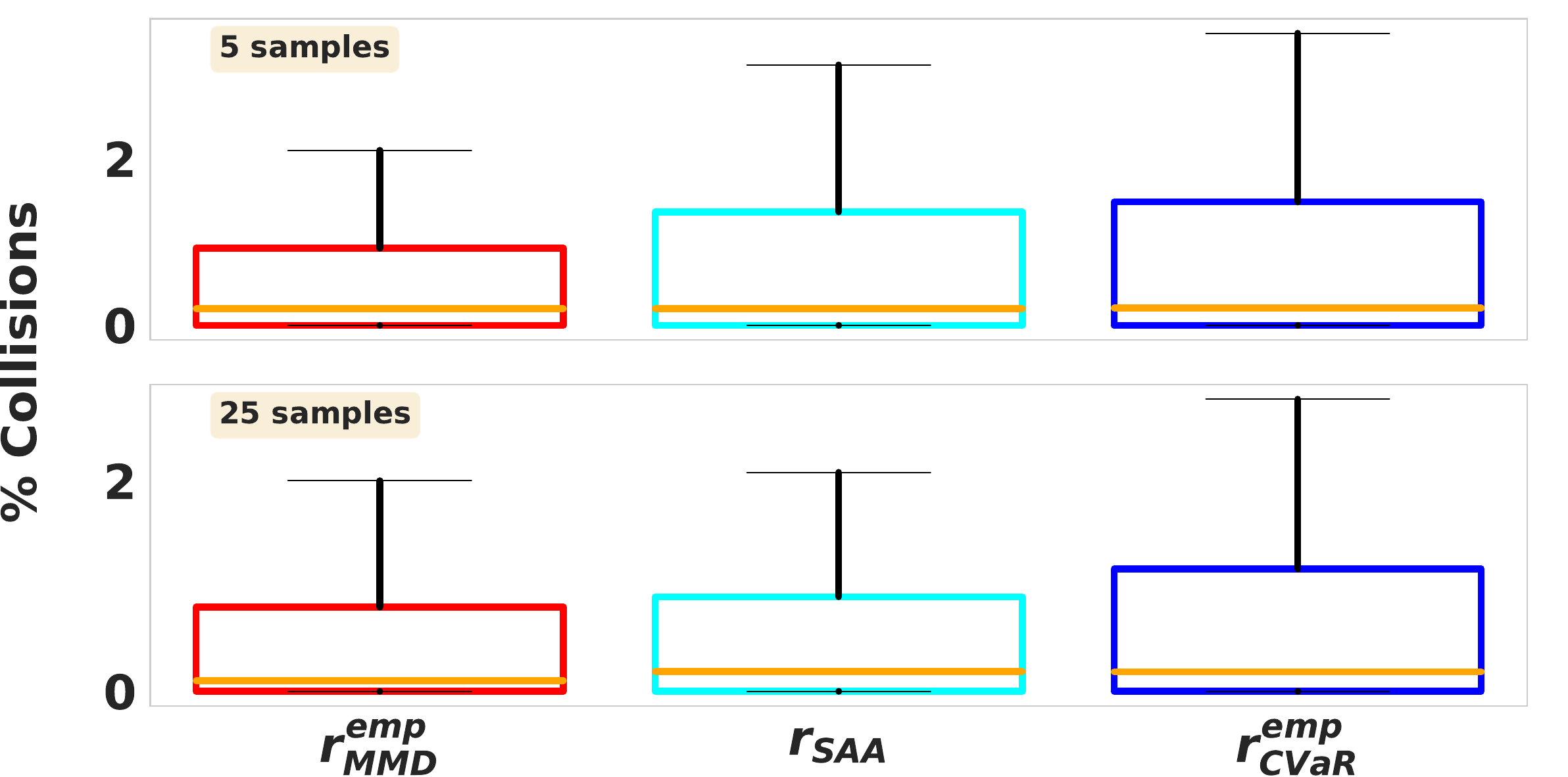}%
         \label{fig:trajectron_overtake_dynamics}
     \end{subfigure}
        \caption{\footnotesize{Dynamics bicycle model}}
        \label{fig:trajectron_box_plots_dynamics_model}
        \vspace{-0.5cm}
\end{figure}

The results are summarized in Fig.\ref{fig:trajectron_box_plots_dynamics_model} for the nuScenes dataset with Trajectron++ as the trajectory predictor. As can be seen, $r_{MMD}^{emp}$ still outperforms $r_{CVaR}^{emp}$ and $r_{SAA}$, even when the underlying motion model is more complex. The pattern is very similar to what we observed in Fig.\ref{fig:trajectron_box_plots} for the bi-cycle kinematic model.

\vspace{-0.3cm}
\subsection{Computation Time}\label{computation_time_section}
\noindent Table \ref{cem_timing} shows the computation time required for our reduced-set optimization Alg.\ref{algo_1} and Alg.\ref{algo_2} for different sample sizes on an RTX 3090 Desktop with 16 GB RAM. We report the timing separately for Alg.\ref{algo_1} and \ref{algo_2}. The former depends on the number of obstacles and also on the size of the optimal \textit{reduced-set} ($N^'$) that we want to construct. At any given $N^'$, Alg.\ref{algo_1} show almost linear growth in the computation-time with respect to the number of obstacles. However, once Alg.\ref{algo_1} is done, the subsequent computation of estimating $r_{MMD}^{emp}$ for a given ego-vehicle trajectory can be suitably parallelized. Hence the increase in computation-time for Alg.\ref{algo_2} with respect to the number of obstacles is very minimal. The run-time indicates that our approach can be used for fast re-planning in a receding horizon fashion.

\section{Conclusions and Future Work}
\indent Estimating and minimizing collision risk based on predicted obstacle trajectories is one of the fundamental problems in autonomous driving. Existing works have used techniques like SAA ($r_{SAA}$, Eqn. \ref{saa_estimate}) and CVaR ($r_{CVaR}^{emp}$, Eqn. \ref{cvar_risk}) to estimate the collision constraint residual function (recall \eqref{constraint_viol_exact}) and define a risk surrogate conditioned on the ego-vehicle and obstacle trajectories. Although effective, a key limitation of these techniques is the lack of a systematic approach to maximize their effectiveness at a given finite sample size. To address this gap, we proposed using the MMD between the distribution of collision-constraint residuals and the Dirac-Delta distribution to model collision risk ($r_{MMD}^{emp}$, Eqn. \ref{risk_mmd}).

\indent The choice of our risk surrogate allows us to use the notion of \textit{reduced-set} to maximize its effectiveness at a given sample size. In other words, the \textit{reduced-set} enables selecting only a subset of predicted obstacle trajectories to formulate our risk surrogate while minimizing accuracy loss. However, fully leveraging its potential requires moving beyond the conventional approach of constructing \textit{reduced-sets} through random sub-selection \cite{simon2016consistent}. Motivated by this, we proposed a novel bi-level optimization method to construct \textit{reduced-sets} while ensuring minimal loss in downstream risk estimation. As a byproduct, the bi-level optimizer also provides an optimal kernel parameter.

\indent We compared the effectiveness of $r_{MMD}^{emp}$, $r_{SAA}$, and $r_{CVaR}^{emp}$ when used as a risk-cost in trajectory optimization. The results show that $r_{MMD}^{emp}$ significantly improves performance, particularly at lower sample sizes. This improvement is especially evident when the underlying prediction distribution deviates substantially from a Gaussian. Our findings demonstrate the suitability of $r_{MMD}^{emp}$ for use alongside large neural network-based predictors, which induce unknown or intractable distributions over obstacle trajectories. We also provided preliminary results on driving scenes from the nuScenes dataset \cite{caesar2020nuscenes} using Trajectron++ as our trajectory predictor.

\indent MMD-OPT is designed for real-time applicability while improving sample efficiency for risk-aware planning.  The \textit{reduced-set} selection mechanism keeps computational costs manageable, and our GPU-accelerated implementation ensures practical feasibility. Validations on nuScenes with state-of-the-art trajectory prediction models further demonstrate MMD-OPT’s effectiveness in handling dynamic, multi-obstacle interactions. Though our current formulation just covers an open-loop optimal planning, it can be trivially extended to a Model Predictive Control (MPC) setting by invoking the planner (Alg.\ref{algo_2}) at each control cycle. The fast computation-times shown in Table \ref{cem_timing} indicate that the resulting MPC can run at around 50 Hz.

\indent Our approach is not limited to autonomous driving, and future efforts will focus on extending it to 3D drone navigation and manipulation in cluttered environments. We also aim to extend our MMD-based risk surrogate to model collision risk under uncertain vehicle dynamics. Future work will also focus on optimizing MMD-OPT for embedded systems and extending its applicability to dense urban traffic environments.

\begin{table}[!t]
\centering
\caption{Computation Time (in seconds) vs Number of Reduced-set samples}
\label{cem_timing}
\scriptsize
\begin{tabular}{|c|c|c|c|}
\cline{2-4}
\multicolumn{1}{c|}{} &  \multicolumn{3}{c|}{\textbf{Number of obstacles}} \\
\hline
\makecell{\textbf{Number of} \\ \textbf{Reduced-set samples}} & \makecell{\textbf{1 obs.} \\(Alg. \ref{algo_1}/Alg.\ref{algo_2})} & \makecell{\textbf{2 obs.} \\(Alg. \ref{algo_1}/Alg.\ref{algo_2})} & \makecell{\textbf{3 obs.} \\(Alg. \ref{algo_1}/Alg.\ref{algo_2})} \\
\hline
5 samples & 0.005/0.01 & 0.01/0.01  & 0.016/0.01 \\ \hline
10 samples & 0.006/0.01 & 0.012/0.01  & 0.019/0.01 \\ \hline
20 samples & 0.006/0.01 & 0.012/0.01  & 0.017/0.01 \\ \hline
30 samples & 0.007/0.01 & 0.014/0.01  & 0.02/0.01 \\ \hline
40 samples & 0.007/0.01 & 0.014/0.01  & 0.02/0.01 \\ \hline
50 samples & 0.008/0.01 & 0.015/0.01  & 0.023/0.01 \\ \hline
\end{tabular}
\normalsize
\vspace{-0.3cm}
\end{table}

\vspace{-0.3cm}
\section{Appendix}\label{appendix}
\subsection{Frenet Planner}\label{frenet_appendix}
\noindent Let $\mathbf{b}= (b_d, b_v)$ be the lateral offset and desired velocity setpoints. Our Frenet planner inspired from \cite{wei2014behavioral} boils down to solving the following trajectory optimization.

\vspace{-0.3cm}
\small
\begin{subequations}
\begin{align}
  \min  \sum_k c_{s} +c_{l}+c_v\label{cost_frenet} \\
     \mathbf{h}(\mathbf{s}^{(q)}, \mathbf{d}^{(q)}) = \mathbf{0}  \label{boundary_cond_frenet}
\end{align}
\end{subequations}
\normalsize
\vspace{-0.5cm}
\small
\begin{subequations}
\begin{align}
    c_{s} (\ddot{s}_{k}, \ddot{d}_{k}) = \ddot{s}^2_{k}+\ddot{d}^2_{k}\\
    c_{l}(d_{k},\dot{d}_{k},\ddot{d}_{k}) = (\ddot{d}_{k}-\kappa_p(d_{k}-b_d)-\kappa_v\dot{d}_{k})^2\\
    c_v(\dot{s}_{k}, \ddot{s}_{k}) = (\ddot{s}_{k}-\kappa_p(\dot{s}_{k}-b_v ))^2
\end{align}
\end{subequations}
\normalsize
The first term $c_s(.)$ in the cost function \eqref{cost_frenet} ensures smoothness in the planned trajectory by penalizing high accelerations at discrete time instants. The last two terms ($c_l(.), c_v(.)$) model the tracking of lateral offset ($b_d$) and forward velocity $(b_v)$ set-points respectively with gain $(\kappa_p, \kappa_v)$. The above optimization is an equality constrained QP that have a closed-form solution. 

\subsection{\textbf{Simplification of $r_{MMD}^{emp}$ Based On Kernel Trick}} 
\noindent We expand equation \eqref{risk_mmd} in the following manner
\small
\begin{align}
    \left\Vert \widehat{\mu}[\overline{f^'}]-\widehat{\mu}[\delta]\right\Vert_{\mathcal{H}}^2 = \langle \widehat{\mu}[\overline{f^'}],\widehat{\mu}[\overline{f^'}]\rangle 
    - 2\langle \widehat{\mu}[\overline{f^'}],\widehat{\mu}[\delta]\rangle
    + \langle \widehat{\mu}[\delta],\widehat{\mu}[\delta]\rangle
    \label{mmd_expand}
\end{align}
\normalsize
\noindent Substituting RKHS embeddings of $\overline{f^'}$ \eqref{mu_fbar_red_set} and $\delta \sim p_{\delta}$ in \eqref{mmd_expand}
\small
\begin{subequations}
    \begin{align}
        \langle \widehat{\mu}[\overline{f^'}],\widehat{\mu}[\overline{f^'}]\rangle 
        &= \langle \sum_{i=1}^{N^'} {^{i}}\beta K(^{i}\overline{f^'},\cdot), \sum_{j=1}^{N^'} {^{j}}\beta K(^j\overline{f^'},\cdot) \rangle \\
        \langle \widehat{\mu}[\overline{f^'}],\widehat{\mu}[\delta]\rangle
         &= \langle \sum_{i=1}^{N^'} {^{i}}\beta K(^{i}\overline{f^'},\cdot), \frac{1}{N^'}\sum_{j=1}^{N^{'}} K(^{j}\delta,\cdot) \rangle \\
    \langle \widehat{\mu}[\delta],\widehat{\mu}[\delta]\rangle
        &= \langle \frac{1}{N^'}\sum_{i=1}^{N^{'}} K(^{i}\delta,\cdot), \frac{1}{N^'}\sum_{j=1}^{N^'} K(^{j}\delta,\cdot) \rangle 
    \end{align}
    \label{mmd_expand_group}
\end{subequations}
\normalsize

\noindent where the computations are done over the reduced-set introduced in Section \ref{reduced_set}. Substituting \eqref{mmd_expand_group} in \eqref{mmd_expand} and using the kernel trick introduced in Section \ref{kernel_section}, we can now derive an optimization friendly expression for the r.h.s of \eqref{risk_mmd} as follows
\small
\begin{align}
    \left\Vert \widehat{\mu}[\overline{f^'}]-\widehat{\mu}[\delta]\right\Vert_2^2 = 
    \mathbf{C}_{\beta}\mathbf{K}_{\overline{f^'},\overline{f^'}}\mathbf{C}_{\beta}^\intercal
    -2\mathbf{C}_{\beta}\mathbf{K}_{\overline{f^'}, \delta}\mathbf{C}_{\overline{\beta}}^\intercal
    + \mathbf{C}_{\overline{\beta}}\mathbf{K}_{\delta,\delta}\mathbf{C}_{\overline{\beta}}^\intercal
\end{align}
\normalsize
\noindent where $\mathbf{C}_{\beta}$ and $\mathbf{C}_{\overline{\beta}}$ are weight vectors given by
\small
\begin{align}
    \mathbf{C}_{\beta} = \left[{^1}\beta,{^2}\beta,\dots,{^{N^'}}\beta \right], \mathbf{C}_{\overline{\beta}} = \left[\frac{1}{N^'},\frac{1}{N^'},\dots,\frac{1}{N^'}\right] \nonumber
\end{align}
\normalsize
\noindent and $\mathbf{K}_{\overline{f^'},\overline{f^'}}, \mathbf{K}_{\overline{f^'}, \delta}$ and $\mathbf{K}_{\delta,\delta}$ are \textit{kernel} matrices defined as
\small
\begin{align}
    \mathbf{K}_{\overline{f^'},\overline{f^'}} &= \begin{bmatrix}
    K(^{1}\overline{f^'},^{1}\overline{f^'}) & \dots  & K(^{1}\overline{f^'},^{N^'}\overline{f^'}) \\
    K(^{2}\overline{f^'},^{1}\overline{f^'}) & \dots  & K(^{2}\overline{f^'},^{N^'}\overline{f^'}) \\
    \vdots & \ddots & \vdots \\
    K(^{N^'}\overline{f^'},^{1}\overline{f^'}) & \dots  & K(^{N^'}\overline{f^'},^{N^'}\overline{f^'})
\end{bmatrix} \nonumber \\
    \mathbf{K}_{\overline{f^'},\delta} &=  \begin{bmatrix}
    K(^{1}\overline{f^'},0) & \dots  & K(^{1}\overline{f^'},0) \\
    K(^{2}\overline{f^'},0) & \dots  & K(^{2}\overline{f^'},0) \\
    \vdots & \ddots & \vdots \\
    K(^{N^'}\overline{f^'},0) & \dots  & K(^{N^'}\overline{f^'},0)
\end{bmatrix} , \mathbf{K}_{\delta,\delta} &= \mathbf{1}_{N^'\times N^'} \nonumber
\end{align}
\normalsize
\balance
\bibliography{references}
\bibliographystyle{IEEEtran}

\end{document}